\documentclass{article}
\usepackage{jfrExamplee}
\usepackage{graphicx}
\usepackage{apalike}
\usepackage{setspace}

\usepackage{caption}
\usepackage{subcaption}
\usepackage{url}

\usepackage{natbib}
\usepackage{amsmath}
\usepackage{booktabs}

\usepackage{moreverb,url}

\title{Multi-Modal Obstacle Detection in Unstructured Environments with Conditional Random Fields}

\author{
Mikkel Kragh \\
Department of Engineering \\
Aarhus University \\
Denmark \\
\texttt{mkha@eng.au.dk} \\
\And
James Underwood \\
Australian Centre for Field Robotics \\
The University of Sydney \\
Australia \\
\texttt{james.underwood@sydney.edu.au}
}

\begin{document}

\maketitle

\begin{abstract}
Reliable obstacle detection and classification in rough and unstructured terrain such as agricultural fields or orchards remains a challenging problem. 
These environments involve large variations in both geometry and appearance, challenging perception systems that rely on only a single sensor modality.
Geometrically, tall grass, fallen leaves, or terrain roughness can mistakenly be perceived as non-traversable or might even obscure actual obstacles. Likewise, traversable grass or dirt roads and obstacles such as trees and bushes might be visually ambiguous.

In this paper, we combine appearance- and geometry-based detection methods by probabilistically fusing lidar and camera sensing with semantic segmentation using a conditional random field. 
We apply a state-of-the-art multi-modal fusion algorithm from the scene analysis domain and adjust it for obstacle detection in agriculture with moving ground vehicles. 
This involves explicitly handling sparse point cloud data and exploiting both spatial, temporal, and multi-modal links between corresponding 2D and 3D regions.

The proposed method was evaluated on a diverse dataset, comprising a dairy paddock and different orchards gathered with a perception research robot in Australia.
Results showed that for a two-class classification problem (ground and non-ground), only the camera leveraged from information provided by the other modality with an increase in the mean classification score of $0.5\%$.
However, as more classes were introduced (\textit{ground}, \textit{sky}, \textit{vegetation}, and \textit{object}), both modalities complemented each other with improvements of $1.4\%$ in 2D and $7.9\%$ in 3D.
Finally, introducing temporal links between successive frames resulted in improvements of $0.2\%$ in 2D and $1.5\%$ in 3D.
\end{abstract}

\section{Introduction}

\label{introduction}
In recent years, automation in the automotive industry has expanded rapidly with products ranging from assisted-driving features to semi-autonomous cars that are fully self-driven in certain restricted circumstances.
Currently, the technology is limited to handle only very structured environments in clear conditions.
However, frontiers are constantly pushed, and in the near future, fully autonomous cars will emerge that both detect and differentiate between objects and structures in their surroundings at all times.

In agriculture, automated steering systems have existed for around two decades~\citep{Abidine2004}.
Farmland is an explicitly constructed environment, which permits recurring driving patterns.
Therefore, exact route plans can be generated and followed to centimeter precision using accurate global navigation systems.
In order to fully eliminate the need for a human driver, however, the vehicles need to perceive the environment and automatically detect and avoid obstacles under all operating conditions.
Unlike self-driving cars, farming vehicles further need to handle unknown and unstructured terrain and need to distinguish traversable vegetation such as crops and high grass from actual obstacles, although both protrude from the ground.
These strict requirements are often addressed by introducing multiple sensing modalities and sensor fusion, thus increasing detection performance, solving ambiguities, and adding redundancy.
Typical sensors are monocular and stereo color cameras, thermal camera, radar, and lidar.
Due to the difference in their physical sensing, the detection capabilities of these modalities both complement and overlap each other ~\citep{Peynot2010,Brunner2013}.

A number of approaches have been made to combine multiple modalities for obstacle detection in agriculture.
Self-supervised systems have been proposed for stereo-radar~\citep{Reina2016}, rgb-radar~\citep{Milella2014,Milella2014_2}, and rgb-lidar~\citep{Zhou2012}.
Here, one modality is used to continuously supervise and improve the detection results of the other.
In contrast, actual sensor fusion provides reduced uncertainty when combining multiple sensors as opposed to applying each sensor individually.
A distinction is often made between low-level (early) fusion, combining raw data from different sensors, and high-level (late) fusion, integrating information at decision level.
At low-level, lidar has been fused with other range-based sensors (lidar and radar) using a joint calibration procedure~\citep{Underwood2010}.
Additionally, lidar has been fused with cameras (monocular, stereo, and thermal) by projecting 3D lidar points onto corresponding images and concatenating either their raw outputs~\citep{Dima2004,Wellington2005} or pre-calculated features~\citep{Haselich2013}.
This approach potentially leverages the full potential of all sensors, but suffers from the fact that only regions covered by all modalities are defined.
Furthermore, it assumes perfect extrinsic calibration between the sensors involved.
At high-level, lidar and camera have been fused for ground/non-ground classification, where the idea is to simply weight the a posteriori outputs of individual classifiers by their prior classification performances~\citep{Reina2015}.
Another approach combines lidar and camera in grid-based fusion for terrain classification into four classes, where again a weighting factor is used for calculating a combined probability for each cell~\citep{Laible2013}.
A similar approach uses occupancy grid mapping to combine lidar, radar, and camera by probabilistically fusing their equally weighted classifier outputs~\citep{Kragh2016}.
However, weighting classifier outputs by a common weighting factor does not leverage the potentially complex connections between sensor technologies and their detection capabilities across object classes.
One sensor may recognize class A but confuse B and C, whereas another sensor may recognize C but confuse A and B.
By learning this relationship, the sensors can be fused to effectively distinguish all three classes.

Recent work on object detection for autonomous driving has fused lidar and camera at a low-level to successfully learn these relationships and improve localization and detection of cars, pedestrians, and cyclists \citep{cvpr17chen}.
The method involves a multi-view convolutional neural network performing region-based feature fusion.
The idea is to apply a region proposal network in 3D to generate bounding boxes of potential objects.
These 3D regions can then be projected to 2D such that features from both modalities can be fused for each region.
A similar method evaluated on the same dataset has been proposed for high-level fusion of lidar and camera \citep{asvadi2017multimodal}.
The detection performance is lower than the above low-level equivalent.
However, the method is considerably faster as it exploits a state-of-the-art real-time 2D network for all modalities.

Research within autonomous underwater vehicles (AUV) has fused camera images from an AUV with \textit{a priori} remote sensing data of ocean depth \citep{Rao2017}.
Here, high-level features from a deep neural network are fused across the two modalities to provide improved classification performance, even when one of the modalities is unavailable during inference.
Similarly, \cite{Eitel2015} have used a convolutional neural network to fuse color and depth images for robotic object recognition on high-level to handle imperfect or missing sensor data.

Within the domain of scene analysis, lidar and camera have recently been combined to improve classification accuracy of semantic segmentation.
In these approaches, a common setup is to acquire synchronized camera and lidar data from a side-looking ground vehicle passing by a scene. 
A camera takes images at a fixed frequency, and a single-beam vertically-scanning laser is used in a push-broom setting, allowing subsequent accumulation of points into a combined point cloud. 
By looking at an area covered by both modalities, a scene consisting of a high number of 3D points and corresponding images is then post-processed, either by directly concatenating features of both modalities at low-level~\citep{Namin2014,Posner2009,Douillard2010,Cadena2013}, or by fusing intermediate classification results provided by both modalities individually at high-level~\citep{Namin2015,Xiao2015,Zhang2015,Munoz2012}.
For this purpose, conditional random fields (CRFs) are often used, as they provide an efficient and flexible framework for including both spatial, temporal, and multi-modal relationships.

In this paper, we apply semantic segmentation on multiple modalities (lidar and camera) for obstacle detection in agriculture.
Unlike object detection (such as detecting cars, pedestrians, and cyclists), semantic segmentation can capture objects that are not easily delimited by bounding boxes (e.g. ground, vegetation, sky).
We adapt the offline fusion algorithm of~\cite{Namin2015} and adjust it for online applicable obstacle detection in agriculture with a moving ground vehicle.
\cite{Namin2015} apply a CRF to jointly infer optimal class labels for both 2D image segments and 3D point cloud segments.
The two modalities are represented with separate nodes in the CRF, allowing partly overlapping regions to be assigned different class labels.
The amount of overlap between a 2D and 3D segment adjusts the link between the modalities, which effectively accounts for inevitable misalignment errors due to calibration and synchronization inaccuracies.
\cite{Namin2015} apply an offline post-processing approach that utilizes the availability of multiple view points of the same objects.
That is, the entire scene is processed as one optimization problem, incorporating a full, dense 3D point cloud accumulated over a traversal of the scene, along with a large number of images from different view points.
For online obstacle detection, however, only the current and previous view points are available.
Point clouds are therefore sparse, and objects are typically only seen from a single view point.
In this paper, we therefore explicitly handle sparse point cloud data and add temporal links to the CRF proposed by \cite{Namin2015} in order to utilize past and present view points.
The method effectively exploits both spatial, temporal, and multi-modal links between corresponding 2D and 3D regions. 
We combine appearance- and geometry-based detection methods by probabilistically fusing lidar and camera sensing using a CRF. 
Visual information (2D) from a color camera serves to classify visually distinctive regions, whereas geometric (3D) information from a lidar serves to distinguish flat, traversable ground areas from protruding elements.
We further investigate a traditional computer vision pipeline and deep learning, comparing the influence on sensor fusion performance.
The proposed method is evaluated on a diverse dataset of agricultural orchards (mangoes, lychees, custard apples, and almonds) and a dairy paddock gathered with a perception research robot.
The dataset is made publicly available and can be downloaded from \url{https://data.acfr.usyd.edu.au/ag/2017-orchards-and-dairy-obstacles/}.

The technical novelty of the paper lies with the introduction of temporal links in the CRF.
Additionally, because the application of the framework is new within agriculture, the paper also presents a thorough evaluation in a range of different agricultural domains.
The main contributions of the paper are therefore fourfold:
\begin{itemize}
\item Adaptation of an offline sensor fusion method used for scene analysis to an online applicable method used for obstacle detection.
This involves extending the framework with temporal links between successive frames, utilizing the localization system of the robot.
\item Comparison of sensor fusion performance when using traditional computer vision and deep learning.
\item Comprehensive evaluation of multi-modal obstacle detection in various agricultural environments. This involves detailed comparisons of single- vs. multi-modality performance, binary vs. multiclass classification, and domain adaptation vs. two domain training strategies.
\item Publicly available datasets including calibrated and annotated images, point clouds, and navigation data. The datasets target multi-modal object detection in robotics and allow for testing domain adaptation across a range of different agricultural domains.
\end{itemize}
The paper is divided into 5 sections. 
Section 2 presents the proposed approach including initial classifiers for the camera and the lidar, individually, and a CRF for fusing the two modalities. 
Section 3 presents the experimental platform and datasets, followed by experimental results in section 4. 
Ultimately, section 5 presents a conclusion and future work.

\section{Approach}
\label{approach}
Our method works by jointly inferring optimal class labels of 2D segments in images and 3D segments in corresponding point clouds. 
By first training individual, initial classifiers for the two modalities, we use a CRF for combining the information using the perspective projection of 3D points onto 2D images. 
This provides pairwise edges between 2D and 3D segments, thus allowing one modality to correct the initial classification result of the other. 
Clustering of 2D pixels into 2D segments and 3D points into 3D segments is necessary in order to reduce the number of nodes in the CRF graph structure.

\begin{figure}[t]
\centering
\includegraphics[width=0.9\textwidth]{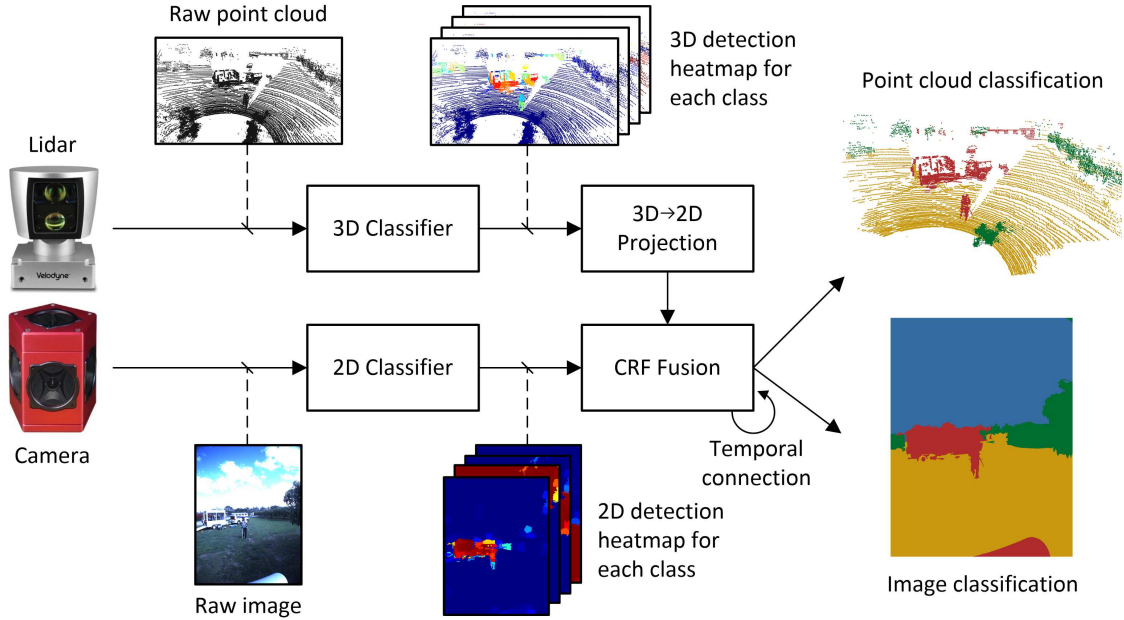}
\caption{Schematic overview of fusion algorithm. 
Initial 2D and 3D classifiers generate class-specific heat maps for a synchronized camera image and a lidar point cloud, individually.
The 3D point cloud is projected onto the 2D image, and a CRF fuses the two information spatially, temporally, and across the two modalities.}
\label{fig:approach}
\end{figure}

A schematic overview of the algorithm is shown in Figure~\ref{fig:approach}.
A synchronized image and point cloud are fed into a pipeline, where feature extraction, segmentation and an initial classification are performed for each modality.
3D segments from the point cloud are then projected onto the 2D image, and a CRF is trained to fuse the two modalities.
Finally, temporal edges are introduced to the CRF by connecting the current and previous frames, utilizing the localization system of the robot.

In the following subsections, the 2D and 3D classifiers are first described individually.
The CRF fusion algorithm is then explained in detail.

\subsection{2D Classifier}
\label{2D}
Most approaches combining lidar and camera use traditional computer vision with hand-crafted image features for the initial 2D classification~\citep{Douillard2010,Cadena2013,Namin2015,Xiao2015,Zhang2015,Munoz2012}.
However, recent advances with self-learned features using deep learning have outperformed the traditional approach for many applications.
In this paper, we therefore compare the two approaches and evaluate their influence when fusing image and lidar data.
Results are presented in section \ref{2D_image_features}.

\begin{figure}[t]
\centering
\begin{subfigure}[t]{0.32\textwidth}
\includegraphics[width=\textwidth]{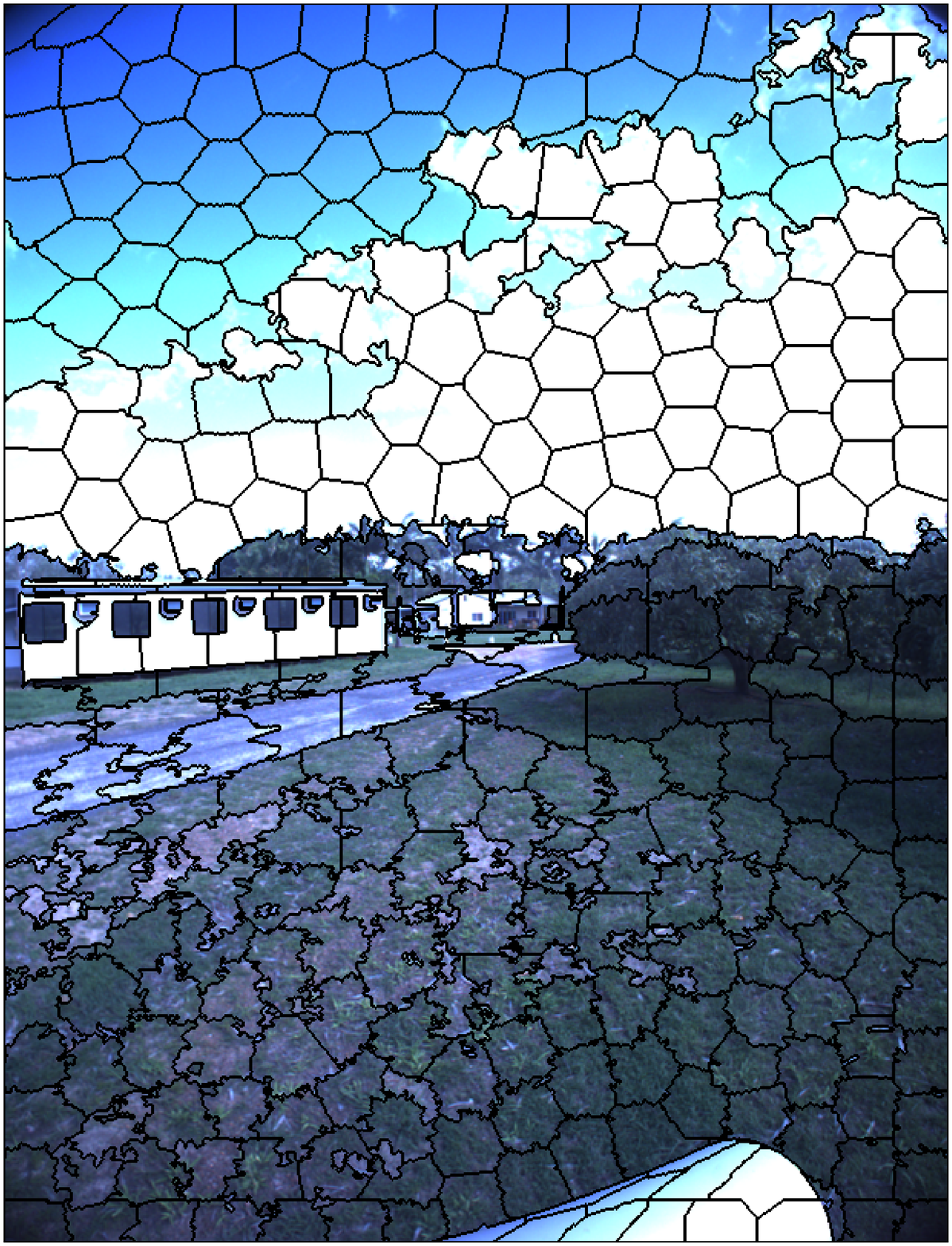} 
\caption{2D superpixels}
\label{fig:2D_example_a}
\vskip 0pt
\end{subfigure}
\begin{subfigure}[t]{0.32\textwidth}
\includegraphics[width=\textwidth]{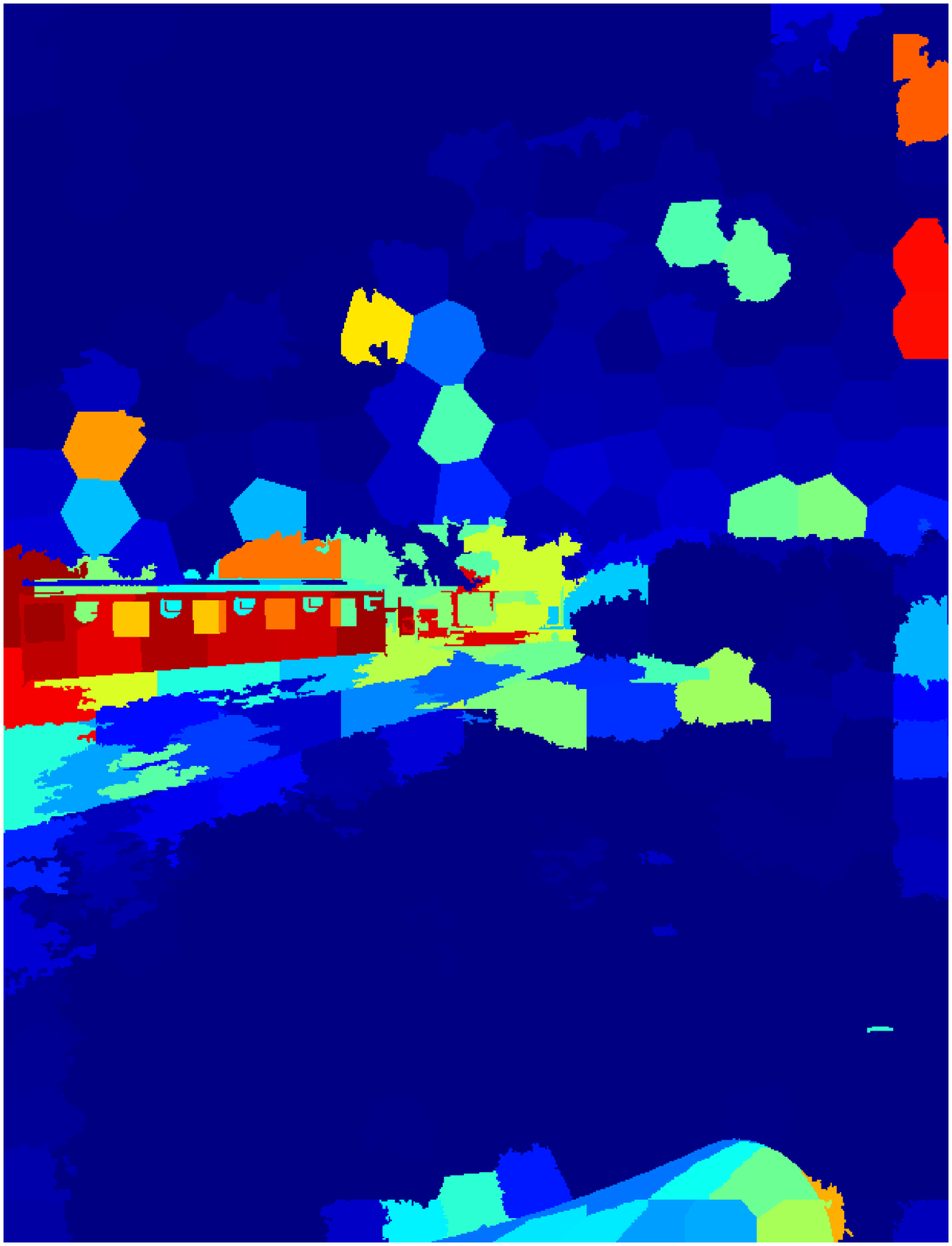} 
\caption{Traditional vision \textit{object} heatmap}
\label{fig:2D_example_b}
\end{subfigure}
\begin{subfigure}[t]{0.32\textwidth}
\includegraphics[width=\textwidth]{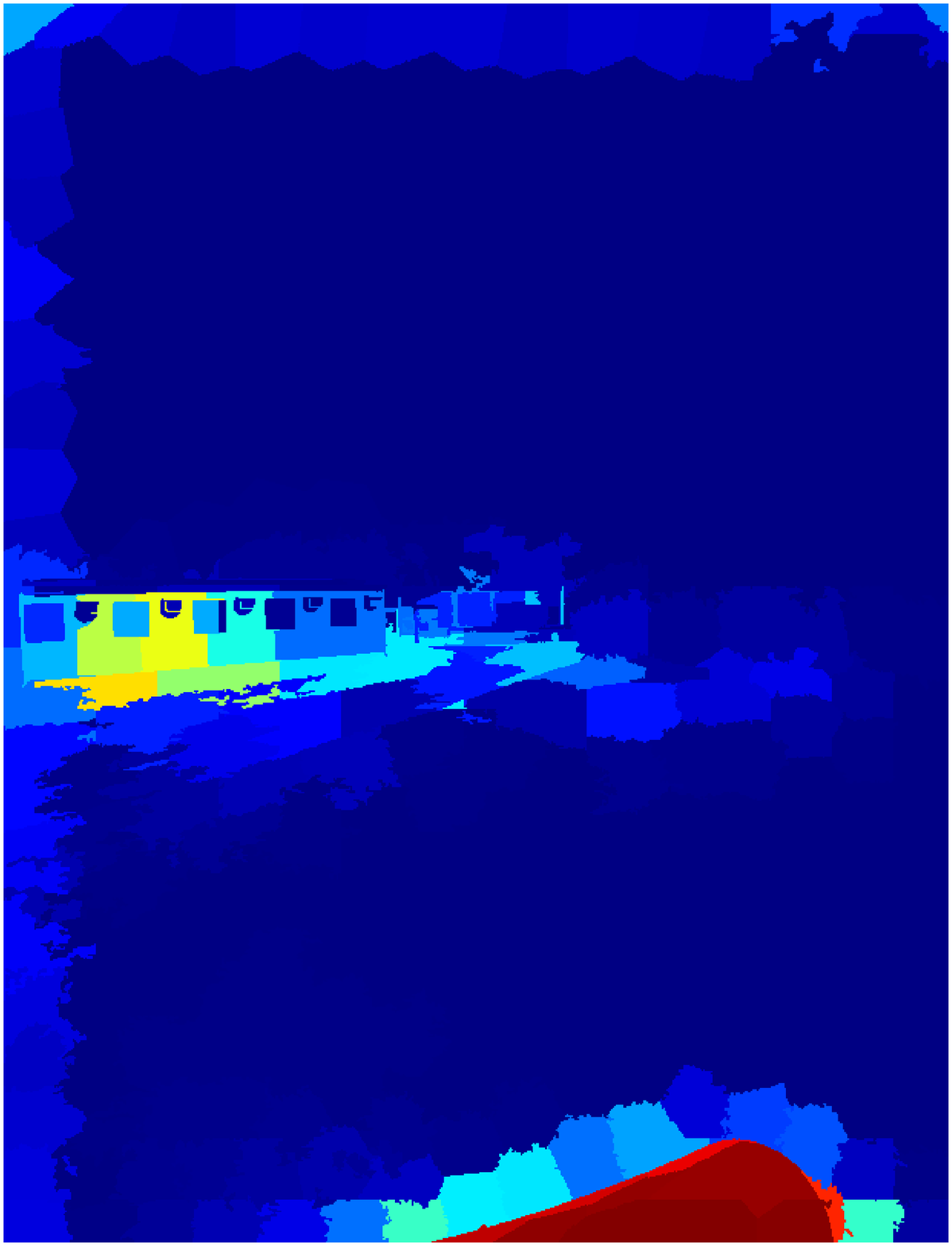} 
\caption{Deep learning \textit{object} heatmap}
\label{fig:2D_example_c}
\end{subfigure}
\caption{Example of 2D segmentation, and probability estimates for traditional vision and deep learning.
(\subref{fig:2D_example_b}) and (\subref{fig:2D_example_c}) use pseudo-coloring for visualizing low (dark blue) and high (dark red) probability estimates.}
\label{fig:2D_example}
\end{figure}

The \textbf{traditional computer vision} pipeline consists of three steps:
the image is first segmented, features are then extracted for each segment, and a classifier is finally trained to distinguish a number of classes based on the features.
In our case, we segment the image into superpixels using SLIC~\citep{Achanta2010} with parameters optimized by cross-validation and listed in Table~\ref{parameter-table}.
Figure~\ref{fig:2D_example_a} shows an example of this segmentation. 
For each superpixel, average RGB values, GLCM features (energy, homogeneity and contrast)~\citep{Haralick1973} and a histogram of SIFT features~\citep{Lowe2004} are extracted. 
The histogram of SIFT features uses a bag-of-words (BoW) representation built using all images in the training set.
Dense SIFT features are calculated over the image, and a histogram of word occurrences is generated for each superpixel. 
All features are then normalized by subtracting the mean and dividing by the standard deviation across the training set. 
Finally, they are used to train a support vector machine (SVM)~\citep{Wu2004} classifier with probability estimates using a one-against-one approach with the libsvm library~\citep{Chang2011}.
This provides probability estimates $P_{\text{initial}}\left(x_i^{\text{2D}}\mid\mathbf{z}_i^{\text{2D}}\right)$ of class label $x_i^{\text{2D}}$, given the features $\mathbf{z}_i^{\text{2D}}$ of superpixel $i$.
An example heatmap of an \textit{object} class is visualized in Figure~\ref{fig:2D_example_b}.

In recent years, \textbf{deep learning} has been used extensively for various machine learning problems.
Especially for image classification and semantic segmentation, convolutional neural networks (CNNs) have outperformed traditional image recognition methods and are today considered state-of-the-art~\citep{Krizhevsky2012,He2015,Long2014}.
In this paper, we use a CNN for semantic segmentation (per-pixel classification) proposed by~\cite{Long2014}.
As we have a very limited amount of training data available, we use a model pre-trained on the PASCAL-Context dataset~\citep{mottaghi_cvpr14}.
This includes 59 general classes, of which only a few map directly to the 9 classes present in our dataset (\textit{ground}, \textit{sky}, \textit{vegetation}, \textit{building}, \textit{vehicle}, \textit{human}, \textit{animal}, \textit{pole}, and \textit{other}).
For the remaining classes, we remap such that all objects (bottle, table, chair, computer, etc.) map to a common \textit{other} class, and all traversable surfaces (grass, ground, floor, road, etc.) map to a common \textit{ground} class.
We then maintain the 59 classes of the pre-trained model, and finetune on the overlapping class labels from our annotated dataset.
In this way, we preserve the ability of the pre-trained network to recognize general object classes (humans, buildings, vehicles, etc.), but use our own data for optimizing the weights towards the specific camera, illumination conditions, and agricultural environment used in our setup.
Table~\ref{parameter-table} lists hyperparameters used for fine-tuning the network.
From our experiments, this procedure has shown to perform better than simply retraining the last layer of the network from scratch with the agriculturally specific classes present in our dataset.

The softmax layer of the CNN provides per-pixel probability estimates for each object class.
However, in this paper, class probability estimates are needed for each superpixel.
We therefore use the same superpixel segmentation as for the traditional vision pipeline, and average and normalize per-pixel estimates within each superpixel.
An example heatmap of an \textit{object} class is visualized in Figure~\ref{fig:2D_example_c}.

\subsection{3D Classifier}
\label{3D}
\begin{figure}[t]
\centering
\begin{subfigure}[b]{0.45\textwidth}
\includegraphics[width=\textwidth]{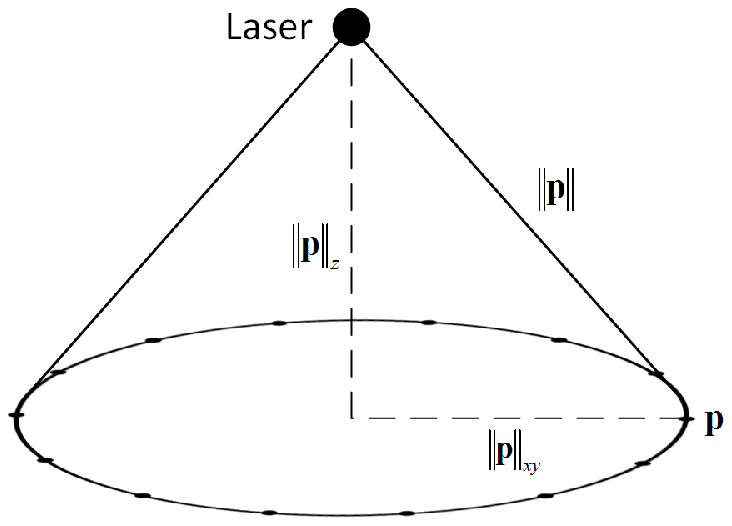} 
\caption{Circular scan pattern on flat ground surface.}
\label{fig:3D_neighborhood_radius_a}
\end{subfigure}
\begin{subfigure}[b]{0.45\textwidth}
\includegraphics[width=\textwidth]{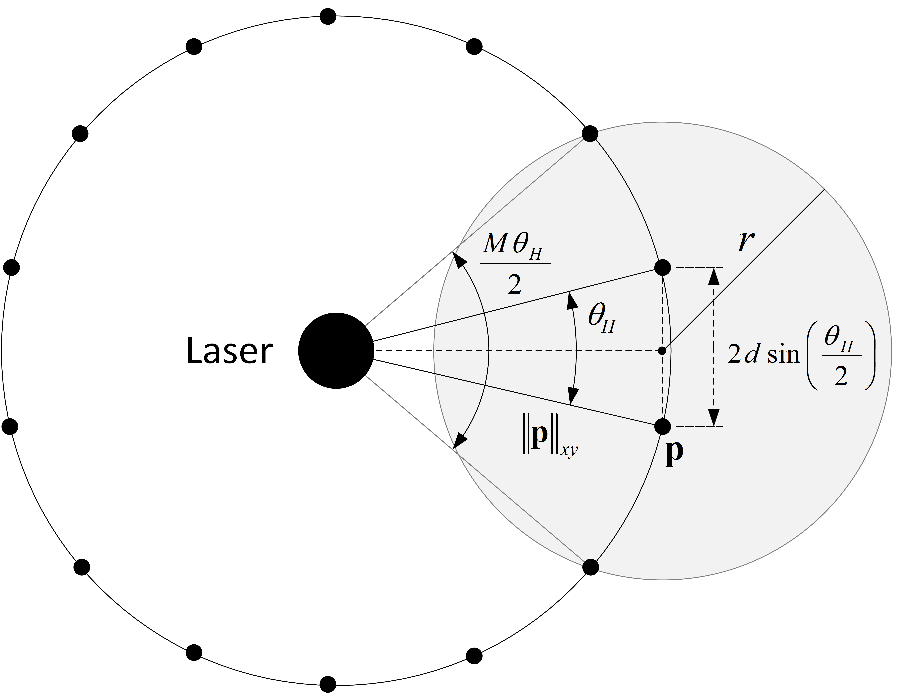} 
\caption{Overlaid adaptive neighborhood radius.}
\label{fig:3D_neighborhood_radius_b}
\end{subfigure}
\caption{Example of adaptive neighborhood radius for single-beam lidar with $M=4$.}
\label{fig:3D_neighborhood_radius}
\end{figure}

\begin{figure}[t]
\centering
\begin{subfigure}[b]{0.32\textwidth}
\includegraphics[width=\textwidth]{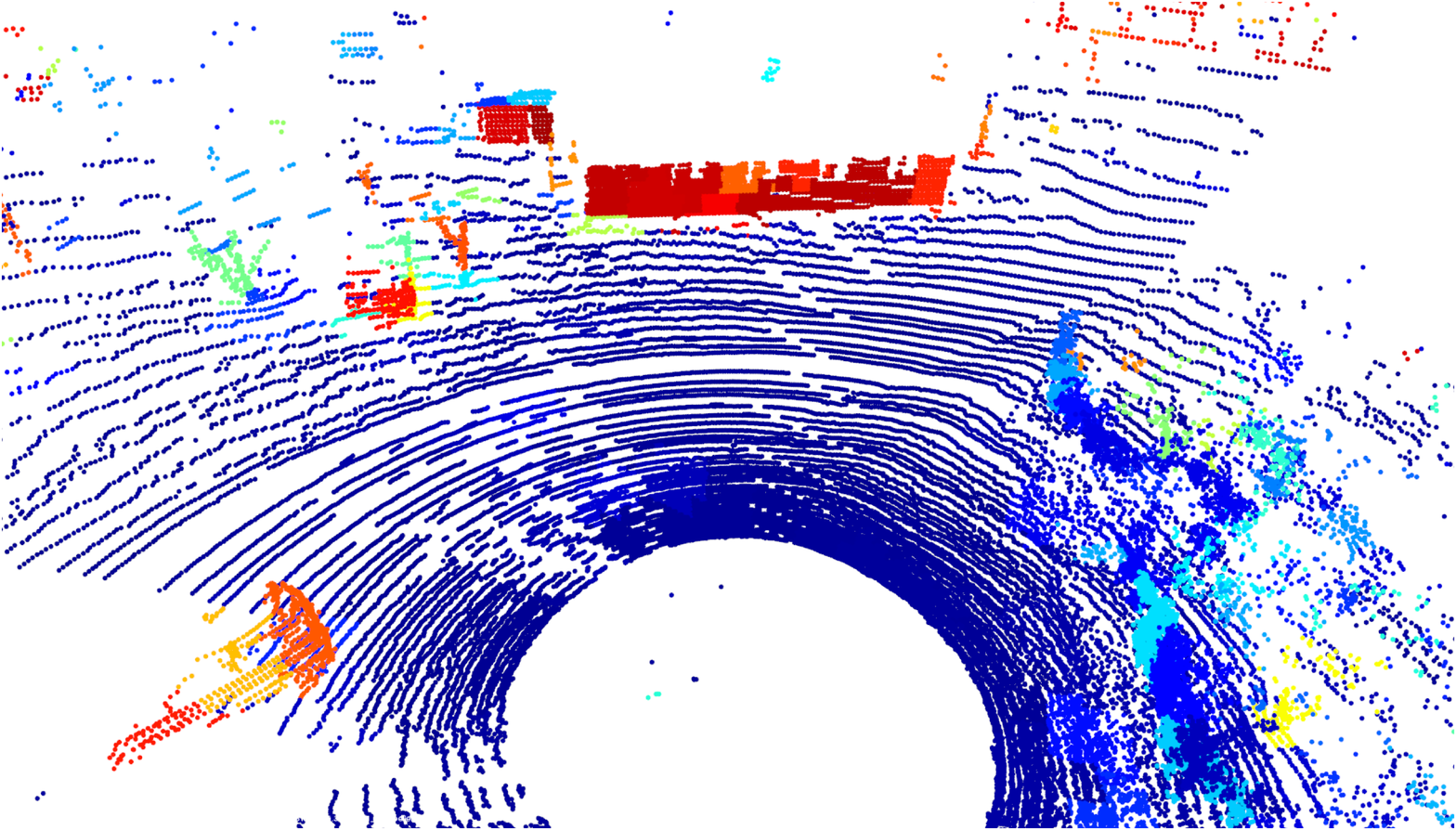} 
\caption{Probability output of \textit{object} class}
\label{fig:3D_example_probabilities}
\end{subfigure}
\begin{subfigure}[b]{0.32\textwidth}
\includegraphics[width=\textwidth]{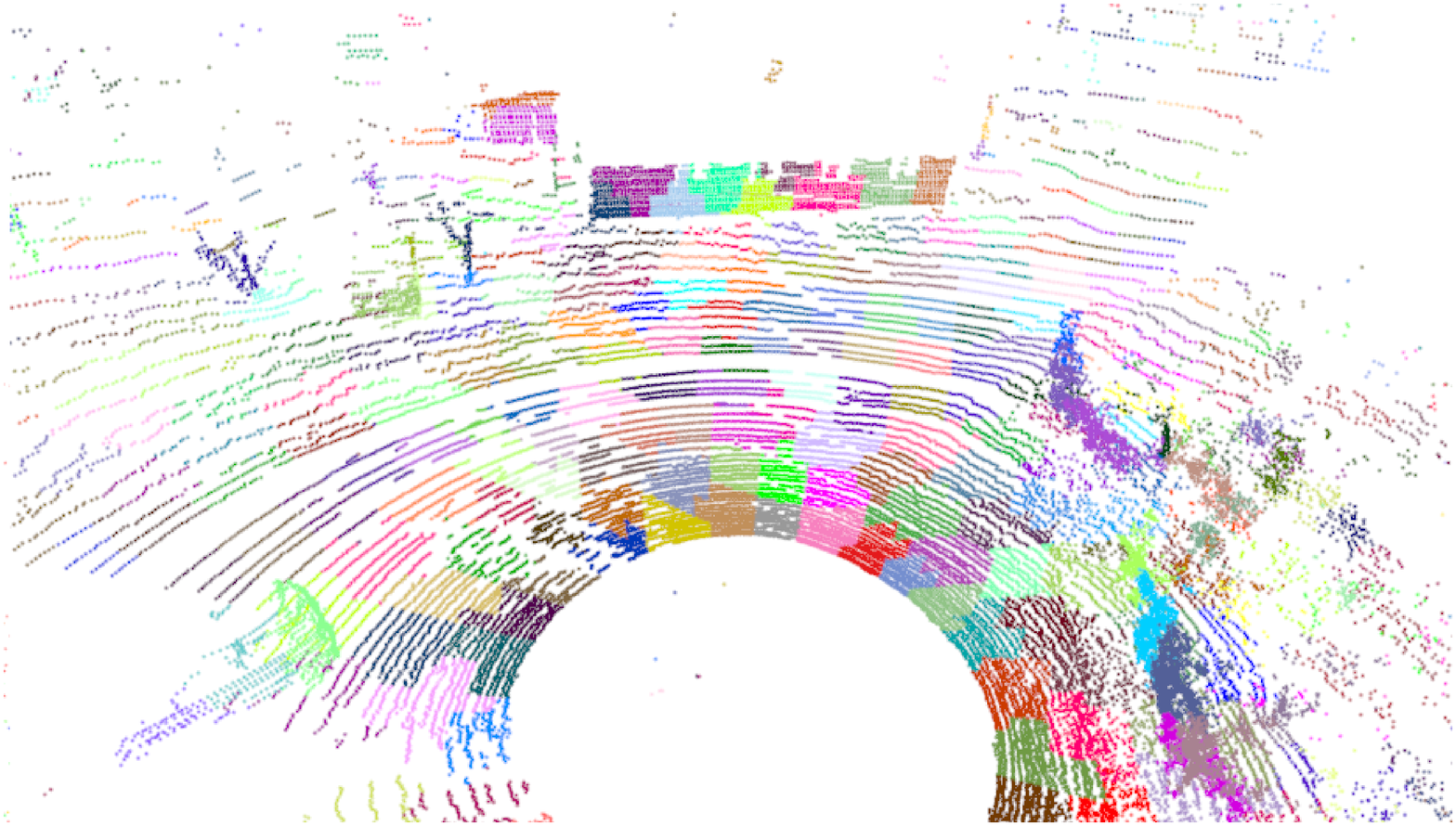} 
\caption{Segmented point cloud}
\label{fig:3D_example_segmented}
\end{subfigure}
\begin{subfigure}[b]{0.32\textwidth}
\includegraphics[width=\textwidth]{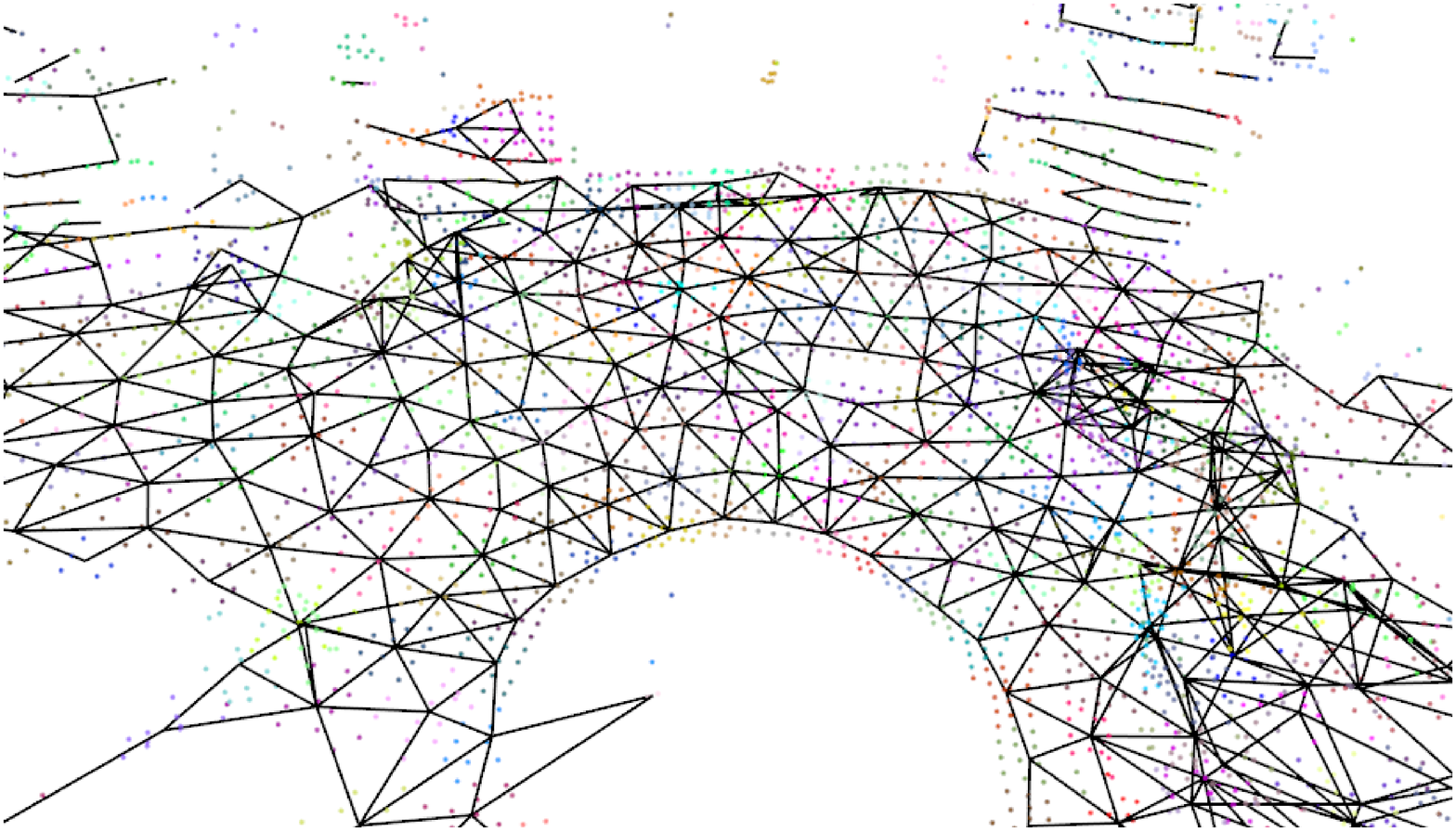} 
\caption{Supervoxel edges}
\end{subfigure}
\caption{Example of 3D classification, segmentation, and edge construction.
(\subref{fig:3D_example_probabilities}) uses pseudo-coloring for visualizing low (dark blue) and high (dark red) probability estimates.}
\label{fig:3D_example}
\end{figure}

When classifying individual points in a point cloud, the point density and distribution influence the attainable classification accuracy, but also the method of choice for feature extraction.
Point features are calculated using a local neighborhood around each point.
Traditionally, this is accomplished with a constant neighborhood size~\citep{Wellington2005,Hebert2003,Lalonde2006,Quadros2012}.
For a single-beam laser accumulating points in a push-broom setting, this procedure works fine, as the point distribution is roughly constant, resulting in a dense point cloud.
For a rotating, multi-beam lidar generating a single scan, however, the point density varies with distance, resulting in a sparse point cloud.
Using a constant neighborhood size in this case, results in either a low resolution close to the sensor or noisy features at far distance.
Therefore, in this paper, we use an adaptive neighborhood size depending on the distance between each point and the sensor.
This ensures high resolution at short distance and prevents noisy features at far distance.
We use the method from~\cite{Kragh2015} which in~\cite{kragh2018} has shown to outperform the generalized 3D feature descriptor FPFH~\citep{rusu2009fast} for sparse, lidar-acquired point clouds.
The method scales the neighborhood size linearly with the sensor distance.
The intuition behind this relationship assumes a flat ground surface beneath the sensor, such that points from a single, rotating beam pointing towards the ground are distributed equally along a circle.
Figure~\ref{fig:3D_neighborhood_radius_a} illustrates this circle along with a top-down view in Figure~\ref{fig:3D_neighborhood_radius_b}.
The radius $\lVert \mathbf{p} \rVert_{xy}$ corresponds to the distance in the ground plane between the sensor and a point $\mathbf{p}$.
The distance between any two neighboring points on the circle is thus $2 \lVert \mathbf{p} \rVert_{xy} \sin{\frac{\theta_H}{2}}$ where $\theta_H$ is the horizontal angle difference (angular resolution).
In order to achieve a neighborhood (gray area) with $M$ points on a single beam, the neighborhood radius must be:
\begin{equation}
\label{eq:neighborhood_radius}
r = 2 \lVert \mathbf{p} \rVert_{xy} \sin{\frac{M \theta_H}{4}}
\end{equation}
which scales linearly with $\lVert \mathbf{p} \rVert_{xy}$.
This relationship holds only for single laser beams.
However, since the angular resolution for a multi-beam lidar is normally much higher horizontally than vertically, the relationship still serves as a good approximation.

The point cloud is first preprocessed by aligning the $xy$-plane with a globally estimated plane using the RANSAC algorithm~\citep{Fischler1981}.
This transformation makes the resulting point cloud have an approximately vertically oriented $z$-axis.
Using the adaptive neighborhood, 9 features related to height, shape, and orientation are then calculated for each point~\citep{Kragh2015}.
$f_1$-$f_4$ are height features.
$f_1$ is simply the $z$-coordinate of the evaluated point, whereas $f_2$, $f_3$, and $f_4$ denote the minimum, mean and variance of all $z$-coordinates within the neighborhood, respectively.
$f_5$-$f_7$ are shape features calculated with principal component analysis. As eigenvalues of the $3\times 3$ covariance matrix, they describe the distribution of the neighborhood points~\citep{Lalonde2006}.
$f_8$ is the orientation of the eigenvector corresponding to the largest eigenvalue.
It serves to distinguish horizontal and vertical structures (e.g. a ground plane and building).
Finally, $f_9$ denotes the reflectance intensity of the evaluated point, provided directly by the lidar sensor utilized in the experiments.
Since the size of the neighborhood varies with distance, all features are made scale-invariant.

\begin{figure}[t]
\centering
\includegraphics[width=0.5\textwidth]{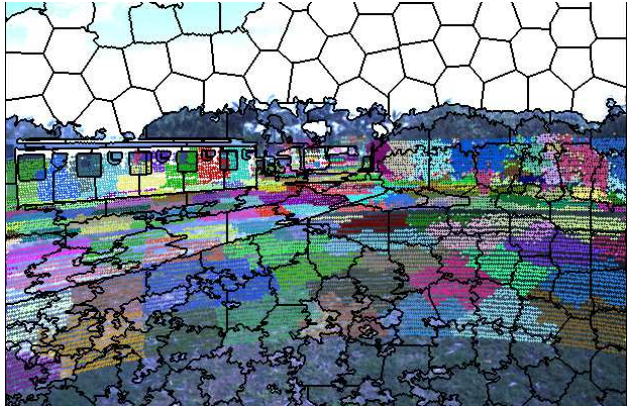} 
\caption{Projection of 3D segments onto 2D superpixels.
Black edges denote 2D superpixel boundaries as in Figure~\ref{fig:2D_example_a}.
Colored crosses denote individual 3D supervoxels (as in Figure~\ref{fig:3D_example_segmented}) projected onto the image.}
\label{fig:3D_projection}
\end{figure}

As for the 2D features, an SVM classifier with probability estimates is trained to provide per-point class probabilities.
A segmentation procedure then clusters points into supervoxels by minimizing both spatial distance and class probability difference between segments. 
Our method uses the approach by \cite{Papon2013} where voxels are clustered iteratively.
However, we modify the feature distance measure $D$ between neighboring segments:
\begin{equation}
\label{eq:supervoxel_equation}
D = \lambda D_s+\chi^2
\end{equation}
where $D_s$ is the spatial Euclidean distance between two segments, $\chi^2$ is the Chi-Squared histogram distance~\citep{Pele2010} between their mean histograms of probability estimates, and $\lambda > 0$ is a weighting factor.
By minimizing this measure during the clustering procedure, points are grouped together based on their spatial distance and initial probability estimates.
Each segment $i$ is then given a probability estimate $P_{\text{initial}}\left(x_i^{\text{3D}}\mid\mathbf{z}_i^{\text{3D}}\right)$ by averaging the class probabilities of all points within the segment. 
Finally, edges between adjacent segments are stored. 
Figure~\ref{fig:3D_example} shows a probability output example of a single class (\textit{object}), the segmented point cloud and its supervoxel edges connecting the segment centers.

Using the extrinsic parameters defining the pose of the lidar and the camera, the point cloud can be projected onto the image using a perspective projection. 
The extrinsic parameters are given by the solid CAD model of the platform including sensors and refined using an unsupervised calibration method for cameras and lasers~\citep{Levinson2013}. 
For computational purposes, the projected points are distorted according to the intrinsics of the camera instead of undistorting the image. 
Figure~\ref{fig:3D_projection} illustrates the projected point cloud, pseudo-coloring points by their associated 3D segments. 
Edges between 2D and 3D segments are then defined by their overlap, such that a large overlap between two segments results in a strong connection, whereas a small overlap results in a weak connection.
Single 2D segments can map to multiple 3D segments and vice versa.
See section~\ref{pairwise_potentials} for further details.

\subsection{Conditional Random Field}
\label{crf}
\begin{figure}[t]
\centering
\includegraphics[width=0.7\textwidth]{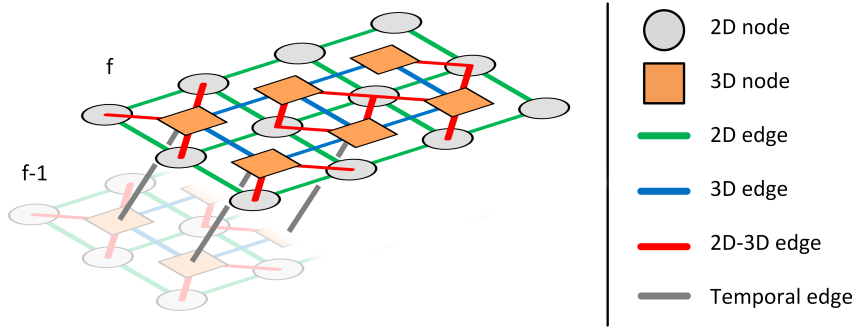} 
\caption{CRF graph with 2D nodes (superpixels), 3D nodes (supervoxels), and edges between them both spatially and temporally.}
\label{fig:crf_graph}
\end{figure}

Once initial probability estimates of all 2D and 3D segments have been found and their edges defined, an undirected graphical model similar to the one visualized in Figure~\ref{fig:crf_graph} can be constructed. 
Each 2D and 3D segment (superpixel and supervoxel) is assigned a node in the graph, and edges between the nodes are defined as described in the sections above.
In Figure~\ref{fig:crf_graph}, additional temporal edges are shown between frame $f$ and $f-1$.
These serve as temporal links between 3D nodes in subsequent frames.

A CRF directly models the conditional probability distribution $p\left(\mathbf{x}\mid \mathbf{z}\right)$, where the hidden variables $\mathbf{x}$ represent the class labels of nodes and $\mathbf{z}$ represent the observations/features. 
The conditional distribution can be written as:
\begin{equation}
\label{eq:cond_prob_dist}
p\left(\mathbf{x}\mid\mathbf{z}\right) = \frac{1}{Z\left(\mathbf{z}\right)} exp\left(-E\left(\mathbf{x}\mid\mathbf{z}\right)\right)
\end{equation}
where $Z\left(\mathbf{z}\right)$ is the partition (normalization) function and $E\left(\mathbf{x}\mid\mathbf{z}\right)$ is the Gibbs energy.
Considering a pairwise CRF for the above graph structure, this energy can be written as:
\begin{equation}
\label{eq:crf_energy}	
E\left(\mathbf{x}\mid\mathbf{z}\right) = \sum_{i=1}^{N^{\text{2D}}}{\phi_i^{\text{2D}}} 
+ \sum_{i=1}^{N^{\text{3D}}}{\phi_i^{\text{3D}}} 
+ \sum_{i,j \in E^{\text{2D}}}^{}{\psi_{ij}^{\text{2D}}}
+ \sum_{i,j \in E^{\text{3D}}}^{}{\psi_{ij}^{\text{3D}}} 
+ \sum_{i,j \in E^{\text{2D-3D}}}^{}{\psi_{ij}^{\text{2D-3D}}} 
+ \sum_{i,j \in E^{\text{Time}}}^{}{\psi_{ij}^{\text{Time}}}
\end{equation}
where $\phi_i^{\text{2D}}$ and $\phi_i^{\text{3D}}$ are unary potentials, $N^{\text{2D}}$ and $N^{\text{3D}}$ are the number of 2D and 3D nodes, $\psi_{ij}^{\text{2D}}$, $\psi_{ij}^{\text{3D}}$, $\psi_{ij}^{\text{2D-3D}}$ and $\psi_{ij}^{\text{Time}}$ are pairwise potentials, and $E^{\text{2D}}$, $E^{\text{3D}}$, $E^{\text{2D-3D}}$ and $E^{\text{Time}}$ are edges.
For simplicity, function variables and weights for the unary and pairwise potentials are left out but explained in more detail in the following sections.

\subsubsection{Unary Potentials}
The unary potentials for 2D and 3D segments are defined by the negative logarithm of their initial class probabilities.
This ensures that the conditional probability distribution in equation~\ref{eq:cond_prob_dist} will correspond exactly to the probability distribution of the initial classifiers if no pairwise potentials are present:
\begin{align}
\phi_i^{\text{2D}}\left(x_i^{\text{2D}},\mathbf{z}_i^{\text{2D}}\right) = -log\left(P_{\text{initial}}\left(x_i^{\text{2D}} \mid \mathbf{z}_i^{\text{2D}}\right)\right) \\
\phi_i^{\text{3D}}\left(x_i^{\text{3D}},\mathbf{z}_i^{\text{3D}}\right) = -log\left(P_{\text{initial}}\left(x_i^{\text{3D}} \mid \mathbf{z}_i^{\text{3D}}\right)\right)
\end{align}
where $\mathbf{z}_i^{\text{2D}}$ and $\mathbf{z}_i^{\text{3D}}$ are the 2D and 3D features described above, and $x_i^{\text{2D}}$ and $x_i^{\text{3D}}$ are the class labels.
The potentials describe the cost of assigning label $x$ to the $i$'th 2D or 3D segment. 
If the probability estimate of the initial classifier is close to $1$, the cost is low, whereas if the probability is close to $0$, the cost is high.

For unary potentials, no CRF weights are included, since we assume class imbalance to be handled by the initial classifiers. 

\subsubsection{Pairwise Potentials}
\label{pairwise_potentials}
In equation~\ref{eq:crf_energy}, three different types of pairwise potentials and edges appear.
These are 2D edges between neighboring 2D superpixel nodes, 3D edges between neighboring 3D supervoxel nodes, 2D-3D edges connecting 2D and 3D nodes through the perspective projection, and temporal edges connecting subsequent frames.
\\

\paragraph{2D and 3D edges}
The pairwise potentials for neighboring 2D or 3D segments act as smoothing terms by introducing costs for assigning different labels.
As is common for 2D segmentation and classification, the cost depends on the exponentiated distance between the two neighbors, such that a small distance will incur a high cost and vice versa~\citep{Boykov2001,Krahenbuhl2011}.
In 2D, the distance is in RGB-space:
\begin{equation}
\psi_{ij}^{\text{2D}}\left(x_i^{\text{2D}},x_j^{\text{2D}},\mathbf{z}_i^{\text{2D}},\mathbf{z}_j^{\text{2D}}\right) = w_p^{\text{2D}}\left(x_i^{\text{2D}},x_j^{\text{2D}}\right)
\cdot \delta\left(x_i^{\text{2D}} \neq x_j^{\text{2D}}\right)
\exp{\left(-\frac{\left\lvert I_i-I_j \right\rvert^2}{2\sigma_{\text{2D}}^2}\right)}
\end{equation}
where $I_i$ is the RGB-vector for superpixel $i$ and $\sigma_{\text{2D}}$ is a weighting factor trained with cross-validation.
$\mathbf{w_p^{\text{2D}}}$ is a weight matrix.
It is learned during training and represents the importance of the pairwise potentials.
The matrix is symmetric and class-dependent, such that interactions between classes are taken into account.
As is common for pairwise potentials, an indicator function (delta function) ensures that the potential is zero for neighboring segments that are assigned the same label.

In 3D, the cost depends on the difference between plane normals~\citep{Hermans2014,Namin2015}:
\begin{equation}
\psi_{ij}^{\text{3D}}\left(x_i^{\text{3D}},x_j^{\text{3D}},\mathbf{z}_i^{\text{3D}},\mathbf{z}_j^{\text{3D}}\right) = w_p^{\text{3D}}\left(x_i^{\text{3D}},x_j^{\text{3D}}\right)
\cdot \delta\left(x_i^{\text{3D}} \neq x_j^{\text{3D}}\right) 
\exp{\left(-\frac{\left\lvert \theta_i - \theta_j\right\rvert^2}{2\sigma_{\text{3D}}^2}\right)}
\end{equation}
where $\theta_i$ is the angle between the vertical z-axis and the locally estimated plane normal for supervoxel $i$ and $\sigma_{\text{3D}}$ is a weighting factor trained with cross-validation. 
The angle is calculated as $\theta = cos^{-1}\left(f_8\right)$ (see section~\ref{3D}).
Similar to 2D, the weight matrix $\mathbf{w_p^{\text{3D}}}$ is symmetric and class-dependent.

\paragraph{2D-3D edges}
\label{2D-3D_edges}
The pairwise potential for 2D and 3D segments connected through the perspective projection is defined by their area of overlap as in~\cite{Namin2015}. 
Let $S_i^{\text{2D}}$ denote the set of pixels in 2D segment $i$, and let $S_j^{\text{3D} \rightarrow \text{2D}}$ denote the set of pixels intersected by the projection of 3D segment $j$ onto the image.
Then, we first define a weight $\omega\left(S_i^{\text{2D}},S_j^{\text{3D}}\right)$ as the cardinality (number of elements) of the intersection of the two sets:
\begin{equation}
\omega\left(S_i^{\text{2D}},S_j^{\text{3D}}\right) = \left\vert{S_i^{\text{2D}} \cap S_j^{\text{3D} \rightarrow \text{2D}}}\right\vert 
\end{equation}
Effectively, this describes the area of overlap between a 2D segment $i$ and a projected 3D segment $j$.
The pairwise potential is then calculated by normalizing this weight by the maximum weight across all 2D segments that are overlapped by the projected 3D segment $j$:
\begin{equation}
\label{eq:weight2D3D}
\psi_{ij}^{\text{2D-3D}}\left(x_i^{\text{2D}},x_j^{\text{3D}},\mathbf{z}_i^{\text{2D}},\mathbf{z}_j^{\text{3D}}\right) = w_p^{\text{2D-3D}}\left(x_i^{\text{2D}},x_j^{\text{3D}}\right)
\cdot \delta\left(x_i^{\text{2D}} \neq x_j^{\text{3D}}\right) \frac{\omega\left(S_i^{\text{2D}},S_j^{\text{3D}}\right)}{\max\limits_{k \in E_j^{\text{2D-3D}}} \omega\left(S_k^{\text{2D}},S_j^{\text{3D}}\right)} 
\end{equation}
where $k$ denotes a 2D segment in the set of all edges $E_j^{\text{2D-3D}}$ generated during the projection of 3D segment $j$ onto the image.
Using this definition of the pairwise potential between 2D and 3D segments, we introduce a cost of assigning corresponding 2D and 3D nodes with different class labels. 
The cost depends on the overlap between the segments, such that a large overlap will result in a high cost, and vice versa.
The normalization in equation~\ref{eq:weight2D3D} ensures that the weights for associating a 3D node to multiple 2D nodes sums to 1.
However, it does not guarantee the opposite.
The sum of weights for associating a 2D node to multiple 3D nodes can thus in theory take any positive value.

Similar to 2D and 3D edges, the weight matrix for 2D-3D edges $\mathbf{w_p^{\text{2D-3D}}}$ is class-dependent.
However, since the potential concerns different domains (2D and 3D), the weights are made asymmetric as in~\cite{Winn2006}.
That is, the cost of assigning $x_i^{\text{2D}}$ to class A and $x_i^{\text{3D}}$ to class B might not be the same as the other way around. 
This allows for interactions that depend on both class label and sensor technology.

\paragraph{Temporal edges}
In order to fuse information temporally across multiple view points, temporal links are added between the current frame and a previous frame.
By utilizing the localization system of the robot, the location of 3D nodes in a previous frame $f_p$ are transformed from the sensor frame into the world frame.
From here, they are then transformed into the current frame $f_c$ where they will likely overlap with the same observed structures.
Effectively, this adds another view point to the sensors and can thus help solve potential ambiguities.
The extrinsic parameters defining the transformation from the navigation frame (localization system) to the sensor frame (lidar) are given by the CAD model of the platform and refined using an extrinsic calibration method for range-based sensors~\citep{Underwood2010}.
In the CRF, temporal edges introduce another pairwise potential:
\newcommand*\mean[1]{\overline{#1}}
\begin{multline}
\psi_{ij}^{\text{Time}}\left(x_{i,f_c}^{\text{3D}},x_{j,f_p}^{\text{3D}},\mathbf{p}_{i,f_c}^{\text{3D}},\mathbf{p}_{j,f_p}^{\text{3D}}\right) = w_p^{\text{Time}}\left(x_{i,f_c}^{\text{3D}},x_{j,f_p}^{\text{3D}}\right) \delta\left(x_{i,f_c}^{\text{3D}} \neq x_{j,f_p}^{\text{3D}}\right) \\
\cdot \exp{\left(-\frac{\mean{\text{diag}\left(\Sigma_{\text{Nav}}\right)}}{2\sigma_\text{Nav}^2}\right)}
\cdot \exp{\left(-\frac{\left\lvert \mathbf{p}_{i,f_c}^{\text{3D}}-T_{f_p}^{f_c}\left(\mathbf{p}_{j,f_p}^{\text{3D}}\right) \right\rvert^2}{2\sigma_\text{Time}^2}\right)}
\end{multline}
Here, $x_{i,f_c}^{\text{3D}}$ is the label of 3D node $i$ in the current frame $f_c$ and
$x_{j,f_p}^{\text{3D}}$ is the label of 3D node $j$ in a previous frame $f_p$.
$\mean{\text{diag}\left(\Sigma_{\text{Nav}}\right)}$ is the mean localization variance, calculated as the mean along the diagonal of the localization covariance matrix averaged from frame $f_p$ to $f_c$.
It incorporates the position and orientation variances and is therefore a measure of the localization accuracy.
$\sigma_\text{Nav}$ is a corresponding weighting factor.
$T_{f_p}^{f_c}$ is the transformation from frame $f_p$ to $f_c$, and $\sigma_\text{Time}$ is an associated weighting factor.
Both weighting factors are trained with cross-validation.

The transformation is provided by the localization system of the robot.
The potential thus depends on the Euclidean distance between a 3D node in the current frame and a transformed 3D node in a previous frame, such that a cost is introduced for assigning different labels at the same 3D location.
By also incorporating localization variance $\mean{\text{diag}\left(\Sigma_{\text{Nav}}\right)}$, the cost is only introduced when localization can be trusted.
That is, a large variance indicates bad localization accuracy which reduces the cost, whereas a small variance indicates good localization accuracy which increases the cost.
Only 3D nodes can be transformed, as 2D nodes do not have a 3D position.
However, since 3D nodes in a previous frame are connected with corresponding 2D nodes, 2D information is indirectly carried on to subsequent frames as well.
Similar to 2D and 3D edges, the weight matrix for temporal edges $\mathbf{w_p^{\text{Time}}}$ is symmetric and class-dependent.

The obtainable improvement with temporal edges depends on a number of factors.
First, the navigation system must be accurate enough to allow reliable transforms of 3D nodes from one frame to another.
Second, the time span between frame $f_p$ and $f_c$ must be large enough to actually add another view point to the sensors.
If $f_p$ and $f_c$ are too close, the robot will not have moved, and no new information is introduced.
However, localization errors can accumulate with distance and time, and therefore $f_p$ and $f_c$ should not be too far apart.
Even further, the temporal connection assumes that the world is static between frame $f_p$ and $f_c$.
If an object (e.g. human) is moving, errors will accumulate over time.
As listed in Table~\ref{parameter-table}, a reasonable compromise of $f_c-f_p=2$ seconds was found to provide the best results.

For training the weight matrix $\mathbf{w_p^{\text{Time}}}$, annotations in 2D and 3D should ideally be available for both frame $f_c$ and $f_p$.
However, this would effectively double the required size of the training set, compared to the other pairwise potentials.
As we are only interested in decoding nodes from $f_c$ (and not $f_p$) during inference, a training procedure utilizing only annotations from the current frame $f_c$ is proposed.
All nodes (2D and 3D) from the previous frame $f_p$ are thus unobserved and have unknown labels.
In order to allow the likelihood of annotated nodes to be maximized, we marginalize out all unobserved nodes.
That is, we sum over all possible classes for each unobserved node, such that the accumulated log likelihood over the entire graph is independent of class labels for unobserved nodes.
In practice, this procedure therefore only optimizes nodes in frame $f_c$, using any information from frame $f_p$ that can increase performance.

\subsubsection{Training and Inference}
\label{crf_training_and_inference}
During training, the CRF weights $\mathbf{w} = \left[\mathbf{w}_p^{\text{2D}},\mathbf{w}_p^{\text{3D}},\mathbf{w}_p^{\text{2D-3D}},\mathbf{w}_p^{\text{Time}}\right]$ are estimated with maximum likelihood estimation.
Additionally, bias weights are introduced for all pairwise terms to account for tendencies independent of the features.
To avoid overfitting, we use $L_2$-regularization for all non-bias weights.
Since the graph is cyclic, exact inference is intractable and loopy belief propagation is therefore used for approximate inference.
The same applies at test time for decoding.
The decoding procedure seeks to determine the most likely configuration of class labels by minimizing the energy $E\left(\mathbf{x}\mid\mathbf{z}\right)$.
The energy can thus be seen as a cost for choosing the label sequence $\mathbf{x}$ given all measurements $\mathbf{z}$.

\section{Experimental Platform and Datasets}
\label{experimental_platform_and_datasets}

\subsection{Platform}
\label{shrimp}
\begin{figure}[t]
\centering
\includegraphics[width=0.25\textwidth]{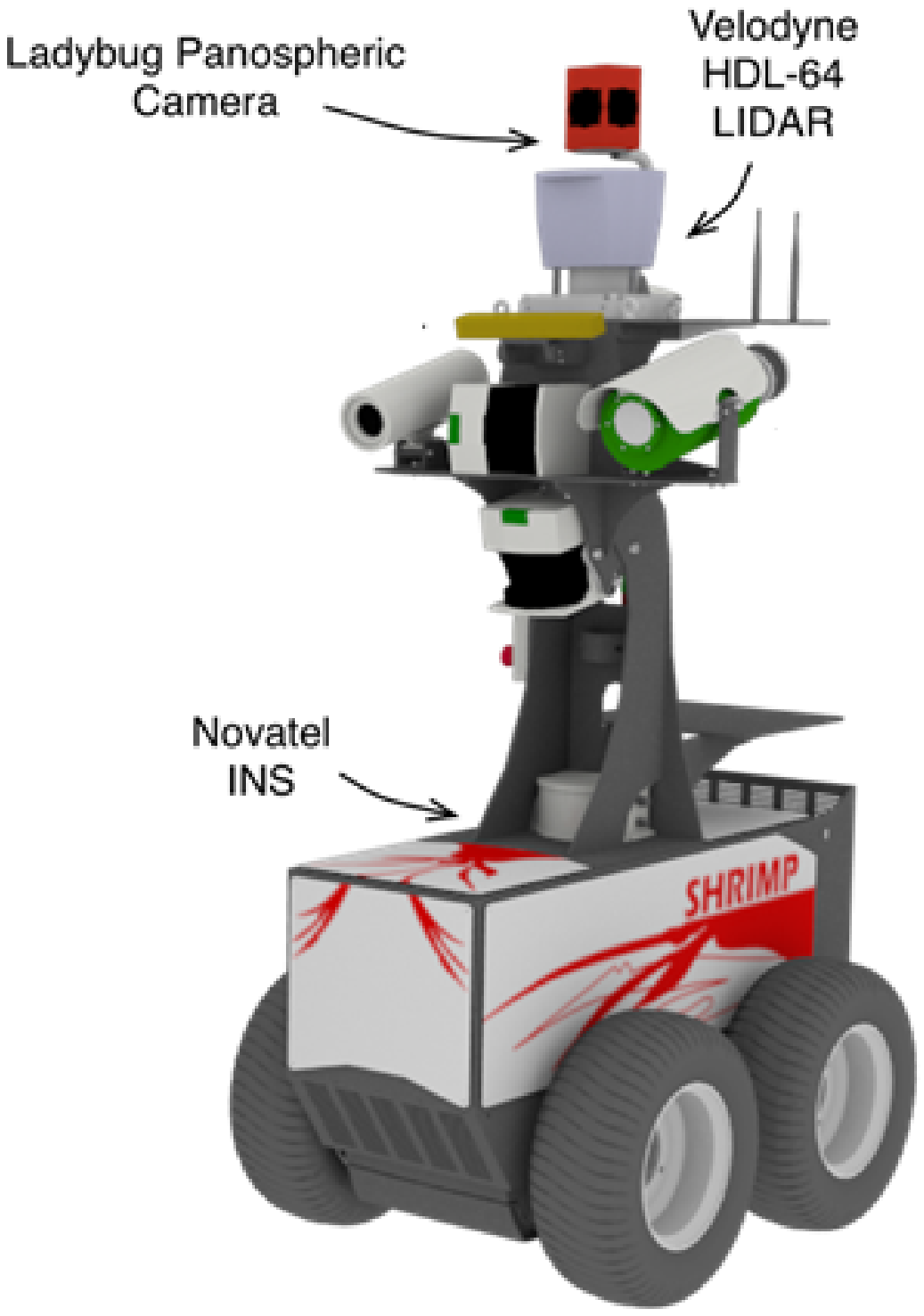} 
\hspace{1cm}
\includegraphics[width=0.6\textwidth]{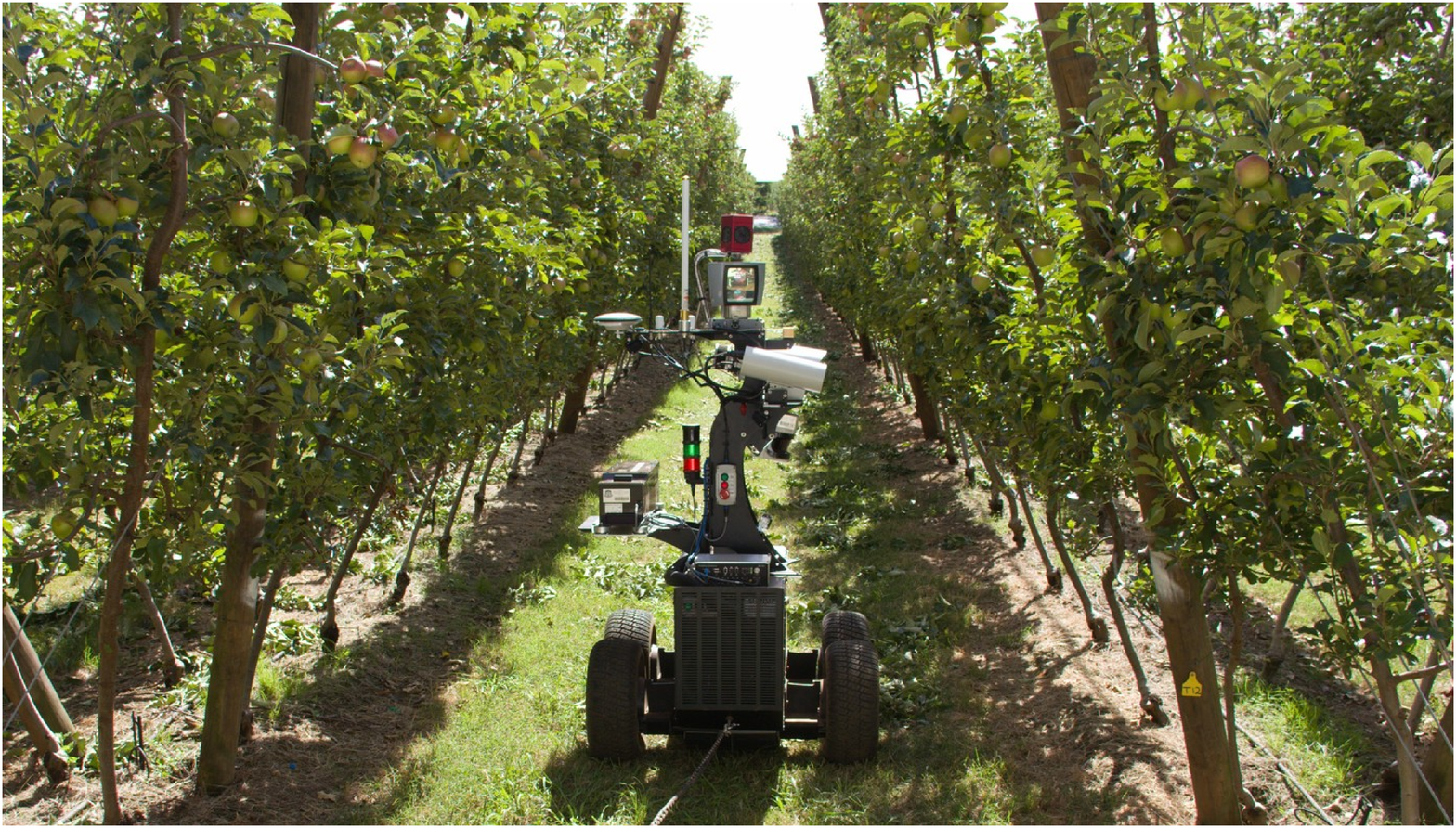} 
\caption{Robotic platform ``Shrimp'' with lidar, panospheric camera, and navigation system.}
\label{fig:shrimp}
\end{figure}

The experimental research platform in Figure~\ref{fig:shrimp} has been used to collect data from various locations in Australia.
The robotic platform is based on a Segway RMP 400 module and has a localization system consisting of a Novatel SPAN OEM3 RTK-GPS/INS with a Honeywell HG1700 IMU, providing accurate 6-DOF position and orientation estimates.
A Point Grey Ladybug 3 panospheric camera system with 6 cameras and a Velodyne HDL-64E lidar both cover a $360^{\circ}$ horizontal view around the vehicle recording synchronized images and point clouds.

Since this paper focuses on obstacle detection, only the forward-facing camera and the corresponding overlapping part of the point clouds are used for the evaluation.

\subsection{Datasets}
\label{datasets}
\begin{figure}[t]
\centering
\begin{subfigure}[b]{0.1949\textwidth}
\includegraphics[width=\textwidth]{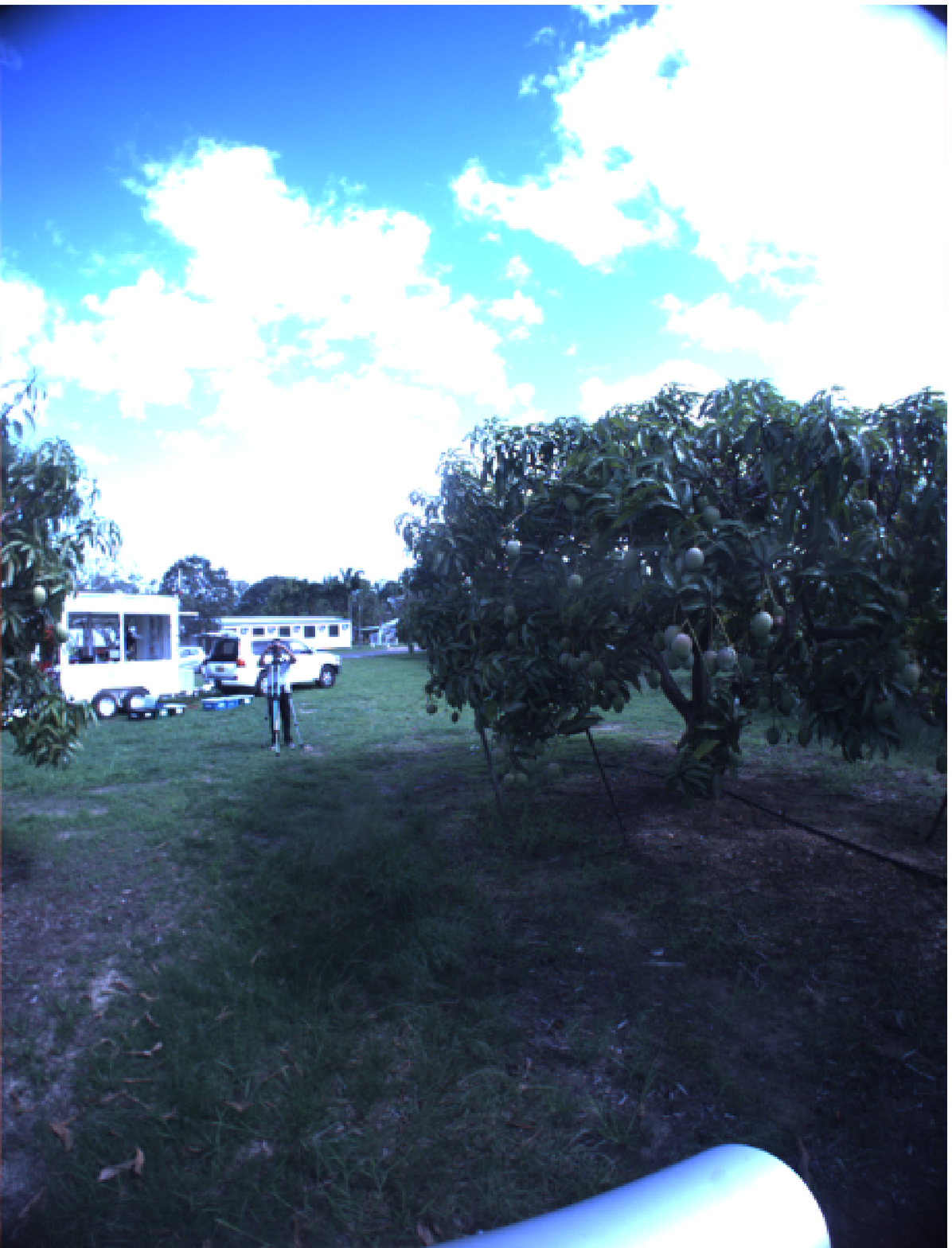} 
\caption{Mangoes}
\label{fig:datasets-mangoes}
\end{subfigure}
\begin{subfigure}[b]{0.1949\textwidth}
\includegraphics[width=\textwidth]{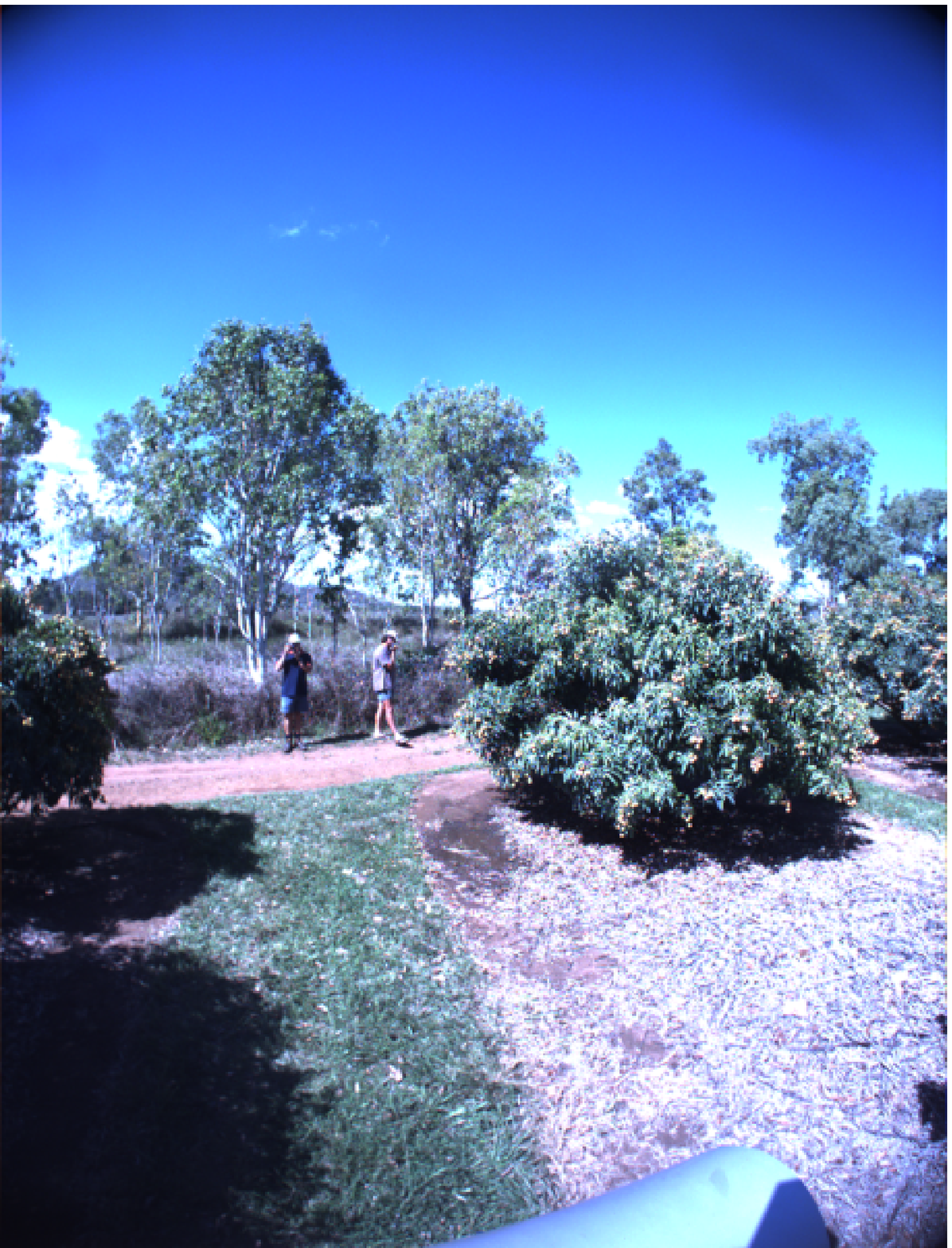} 
\caption{Lychees}
\label{fig:datasets-lychees}
\end{subfigure}
\begin{subfigure}[b]{0.1949\textwidth}
\includegraphics[width=\textwidth]{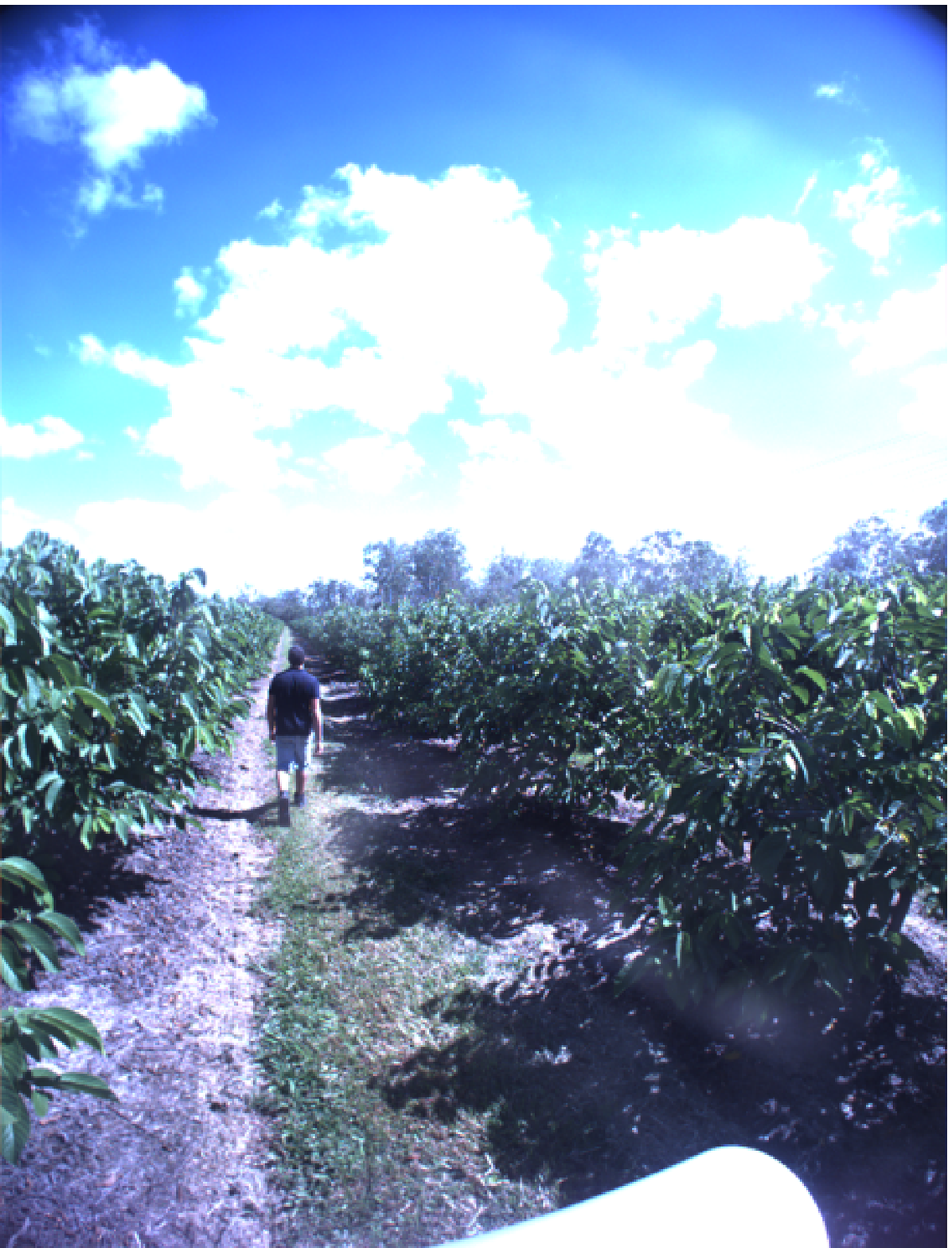} 
\caption{Custard apples}
\label{fig:datasets-apples}
\end{subfigure}
\begin{subfigure}[b]{0.1949\textwidth}
\includegraphics[width=\textwidth]{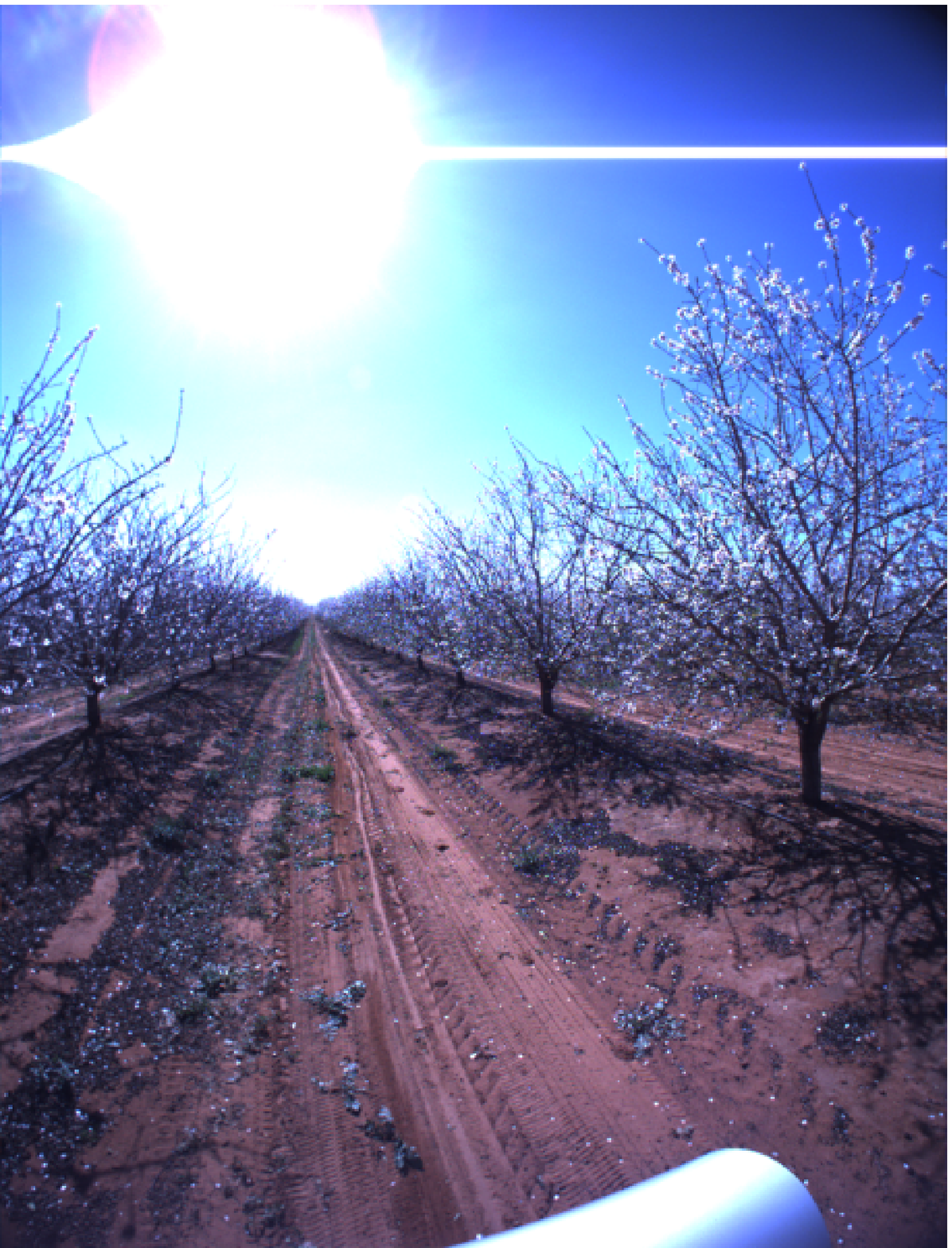} 
\caption{Almonds}
\label{fig:datasets-almonds}
\end{subfigure}
\begin{subfigure}[b]{0.1949\textwidth}
\includegraphics[width=\textwidth]{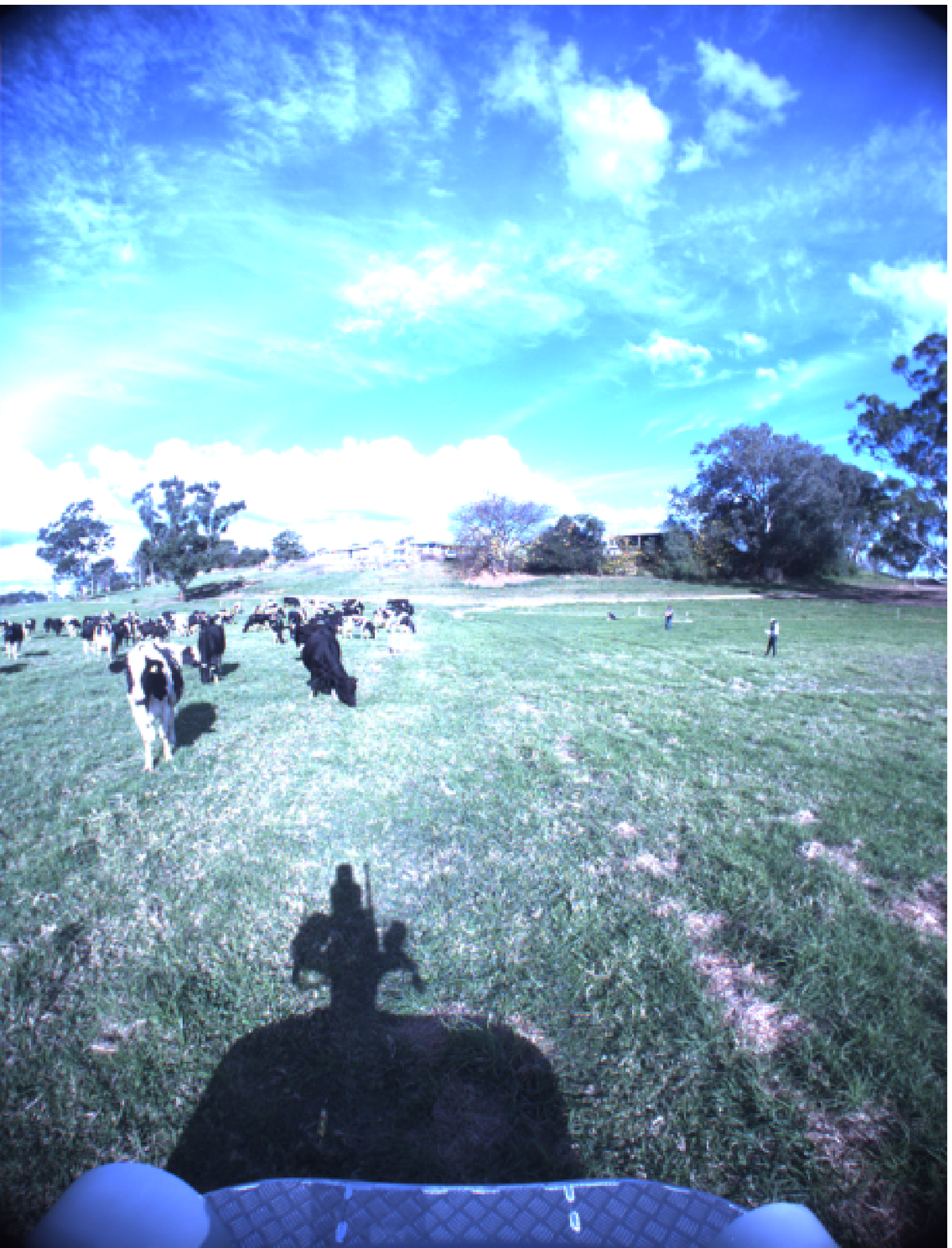} 
\caption{Dairy}
\label{fig:datasets-dairy}
\end{subfigure}
\caption{Example images from datasets.}
\label{fig:datasets}
\end{figure}

\newcommand{\multiline}[1]{\begin{tabular}[t]{@{}l@{}}#1\end{tabular}}
\begin{table}[t]
\caption{Dataset overview.} \label{dataset-table}
\resizebox{\textwidth}{!}{
\begin{tabular}{llllll@{\hskip 0.1cm}lll}
\toprule
Dataset & Environment & Season & Length & \multiline{Annotated\\frames} & \multicolumn{2}{l}{\multiline{Annotated\\2D/3D segments}} & Obstacles*                                       \\
\midrule
Mangoes & Orchard     & Summer & 408 m (359 s) & 36 & 12096 & / 28001    & \multiline{Buildings, trailer, cars, \\tractor, boxes, humans} \\
Lychees & Orchard     & Summer & 122 m (121 s) & 15 & 5040 & / 7400      & \multiline{Buildings, trailers, cars, \\humans, iron bars}     \\
Apples  & Orchard     & Summer & 159 m (128 s) & 23 & 7728 & / 9708      & \multiline{Trailer, car, humans, \\poles}                      \\
Almonds & Orchard     & Spring & 258 m (212 s) & 31 & 10416 & / 33260    & \multiline{Buildings, cars, humans, \\dirt pile, plate}        \\
Dairy   & Field       & Winter & 91 m (106 s)  & 15 & 5040 & / 18511     & \multiline{Humans, hills, poles, \\cows} \\
\bottomrule
\end{tabular}
}
* All frames contain ground and vegetation (trees).
\end{table}

From May to December 2013, data were collected across different locations in Australia. 
The diverse datasets include recordings from both a dairy paddock and orchards with mangoes, lychees, custard apples, and almonds. 
Figure~\ref{fig:datasets} illustrates a few examples from the forward-facing Ladybug camera during the recordings. 
Various objects/obstacles such as humans, cows, buildings, vehicles, trees, and hills are present in the datasets.
A total of $120$ frames have been manually annotated per-pixel in 2D images and per-point in 3D point clouds.
By annotating both modalities separately, we can evaluate non-overlapping regions and get reliable ground truth data even if there is a slight calibration error between the two modalities.
$9$ categories are defined (\textit{ground}, \textit{sky}, \textit{vegetation}, \textit{building}, \textit{vehicle}, \textit{human}, \textit{animal}, \textit{pole}, and \textit{other}). 
Due to the physics of the lidar, \textit{sky} is only present in the images.
Table~\ref{dataset-table} presents an overview of the datasets.
The dataset along with all annotations is made publicly available and can be downloaded from \url{https://data.acfr.usyd.edu.au/ag/2017-orchards-and-dairy-obstacles/}.

\section{Experimental Results}
\label{results}
To evaluate the proposed algorithm, a number of experiments were carried out on the datasets presented in Table~\ref{dataset-table}.
First, the overall results are presented by evaluating the improvement in classification when introducing the fusion algorithm.
Then, we specifically address binary and multiclass scenarios, compare traditional vision with deep learning, and evaluate the transferability of features and classifiers across domains (\textit{mangoes}, \textit{lychees}, \textit{apples}, \textit{almonds}, and \textit{dairy}) with domain adaptation.
Finally, we compare the performance of domain adaptation and domain training.

To obtain sufficient training examples for each class, the categories \textit{building}, \textit{vehicle}, \textit{human}, \textit{animal}, \textit{pole} and \textit{other} were all mapped to a common \textit{object} class.
A total of four classes were thus used for the following experiments, $x_i = \left\{\textit{ground},\textit{sky},\textit{vegetation},\textit{object}\right\}$.
For all experiments, $5$-fold cross-validation was used corresponding to the $5$ different datasets in Table~~\ref{dataset-table}.
That is, for each dataset, data from the remaining four datasets were used for training initial classifiers and CRF weights.
This was done to test the system in the more challenging but realistic scenario, where training data is not available for the identical conditions as where the system would be deployed.

For image classification and CRF training and decoding, we used MATLAB along with the computer vision library VLFeat~\citep{vedaldi08vlfeat}, and the undirected graphical models toolbox UGM \citep{Schmidt2007}.
For point cloud classification, we used C\texttt{++} and Point Cloud Library (PCL) \citep{Rusu2011}.
A list of parameter settings for all algorithms is available in Appendix A.

\subsection{Results Overview}
\label{results_overview}
Table~\ref{crf-table} presents the results for applying the CRF with the three different types of pairwise potentials enabled.
\textit{Initial}, \textit{CRF$_{\text{2D}}$}, and \textit{CRF$_{\text{3D}}$} thus refer to single-modality results obtained with the direct output of the initial 2D or 3D classifier and the ``smoothed'' version of the CRF, respectively.
\textit{CRF$_{\text{2D-3D}}$} additionally introduces sensor fusion by adding edges across the two modalities, while \textit{CRF$_{\text{2D-3D,Time}}$} further adds temporal links across subsequent frames.
The results are presented in terms of intersection over union (IoU) and accuracy.
Both measures were evaluated per-pixel in 2D and per-point in 3D, thus disregarding the superpixel and supervoxel clusters.
Results were obtained with the traditional vision classifier (instead of the deep learning variant) for 2D as it provided the better fusion results.
A detailed comparison of traditional vision and deep learning is described in section~\ref{2D_image_features}.

\begin{table}[t]
\caption{Classification results for 2D and 3D.} \label{crf-table}
\begin{center}
\begin{tabular}{lccccc|c}
\toprule
& \multicolumn{5}{c|}{IoU}																& accuracy 		\\
& ground			& sky			& vegetation		& object		& mean			& 				\\
\midrule
2D, Initial	                    & 0.847 & 0.933 & 0.729 & 0.233 & 0.685 & 0.900			\\ 
2D, CRF$_{\text{2D}}$	        & 0.893	& \textbf{0.971}	& 0.763	& 0.342	& 0.742 & 0.937			\\
2D, CRF$_{\text{2D-3D}}$	    & \textbf{0.907} & \textbf{0.971} & 0.774 & 0.372 & 0.756 & \textbf{0.943}			\\
2D, CRF$_{\text{2D-3D,Time}}$   & \textbf{0.907} & \textbf{0.971} & \textbf{0.775}	& \textbf{0.379} & \textbf{0.758} & \textbf{0.943}         \\
\midrule
3D, Initial						& \textbf{0.936} & - & 0.735 & 0.365 & 0.678 & 0.881			\\
3D, CRF$_{\text{3D}}$			& 0.933 & - & 0.846 & 0.466 & 0.748 & 0.923			\\
3D, CRF$_{\text{2D-3D}}$		& 0.929 & -	& 0.886 & 0.667 & 0.827 & 0.943			\\
3D, CRF$_{\text{2D-3D,Time}}$	& 0.933	& - & \textbf{0.897} & \textbf{0.697} & \textbf{0.842} & \textbf{0.948}         \\
\bottomrule
\end{tabular}
\end{center}
\end{table}

From Table~\ref{crf-table}, we see a gradual improvement in classification performance when introducing more terms in the CRF.
First, the initial classifiers for 2D and 3D were improved separately by adding spatial links between neighboring segments.
This caused an increase in mean IoU of $5.7\%$ in 2D and $7.0\%$ in 3D.
Then, by introducing multi-modal links between 2D and 3D, the performance was further increased.
In 2D, the increase in mean IoU was only $1.4\%$, whereas in 3D it amounted to $7.9\%$.
The most prominent increases belonged to the \textit{object} class, where appearance or geometric clues from one modality significantly helped recognize the class in the other modality.
Ultimately, adding temporal edges provided the best overall performance. 
In 2D, a subtle increase in mean IoU of $0.2\%$ was achieved, whereas in 3D, temporal edges caused an increase of $1.5\%$.
The most significant increase was for the \textit{object} class in 3D with an increase in IoU of $3.0\%$.
As temporal edges link 3D nodes between frames, it makes intuitive sense that 3D performance was improved more than 2D.

\begin{figure}[t]
\centering
\begin{subfigure}[b]{\textwidth}
\hfill
\includegraphics[width=0.48\textwidth]{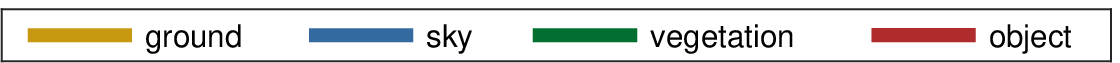} 
\vspace{0.05cm}
\end{subfigure}
\begin{subfigure}[b]{0.194\textwidth}
\includegraphics[clip,trim={0 4cm 0 0},width=\textwidth]{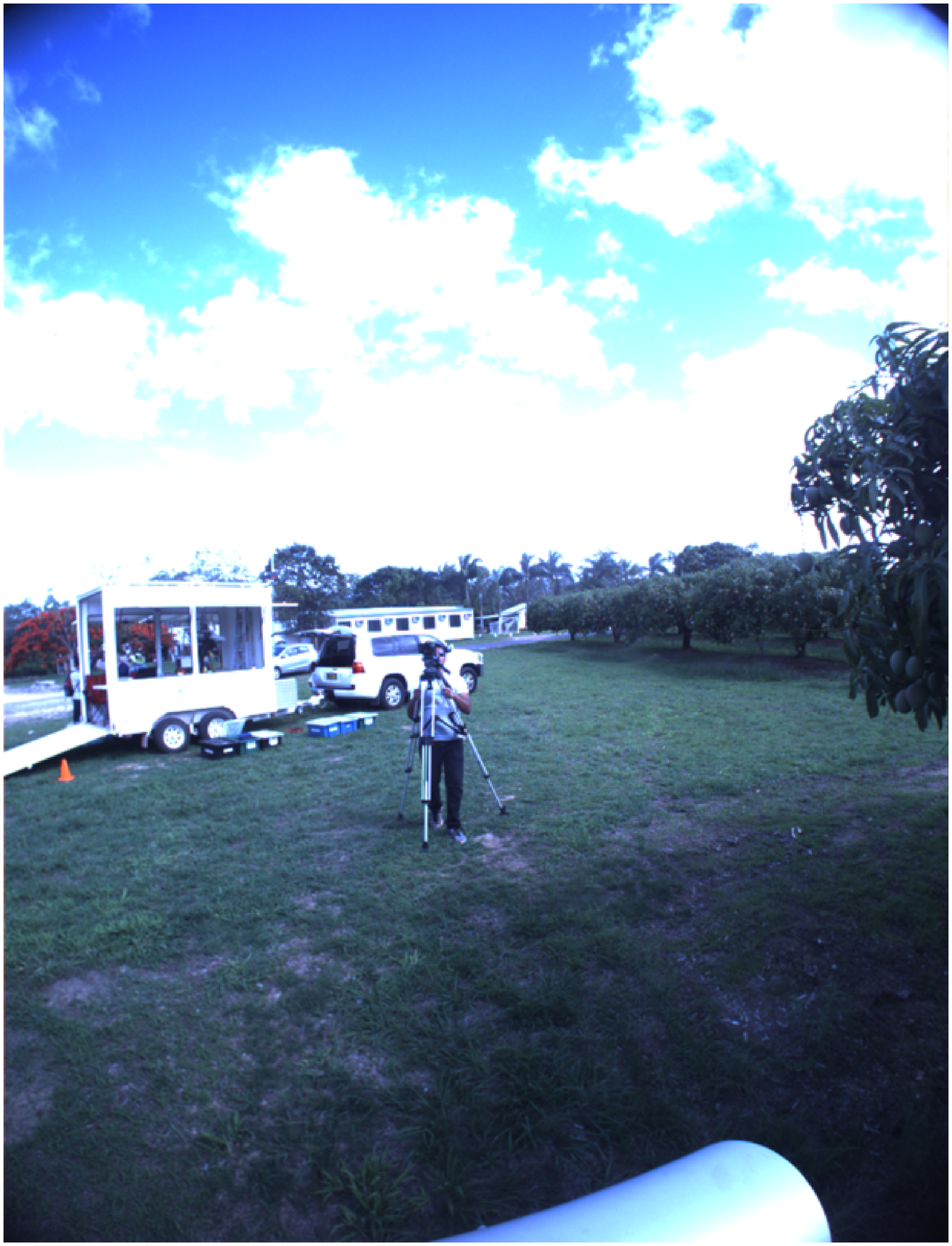} 
\caption{2D image}
\end{subfigure}
\begin{subfigure}[b]{0.194\textwidth}
\includegraphics[clip,trim={0 4cm 0 0},width=\textwidth]{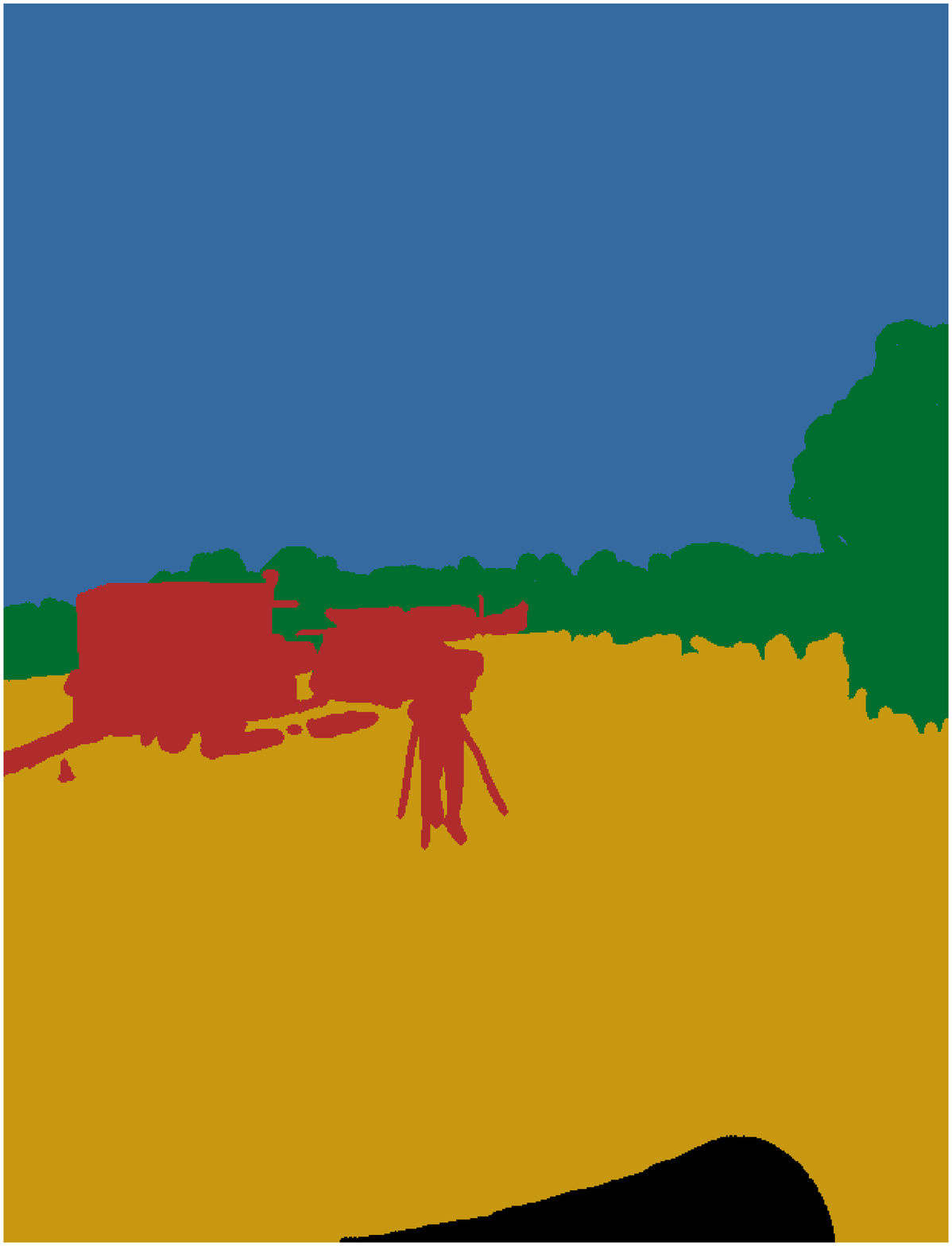} 
\caption{Ground truth}
\end{subfigure}   
\begin{subfigure}[b]{0.194\textwidth}
\includegraphics[clip,trim={0 4cm 0 0},width=\textwidth]{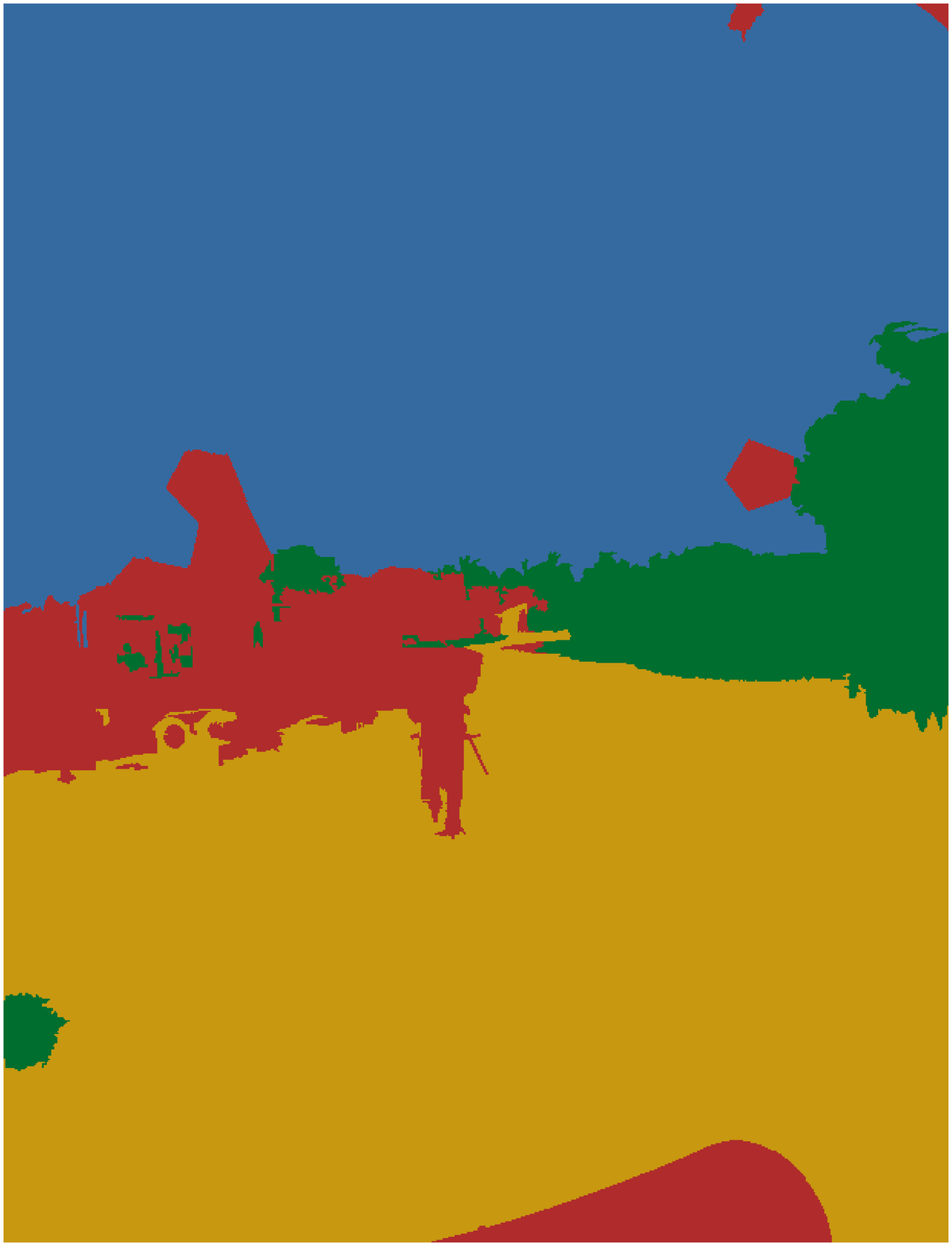} 
\caption{Initial classifier}
\end{subfigure}  
\begin{subfigure}[b]{0.194\textwidth}
\includegraphics[clip,trim={0 4cm 0 0},width=\textwidth]{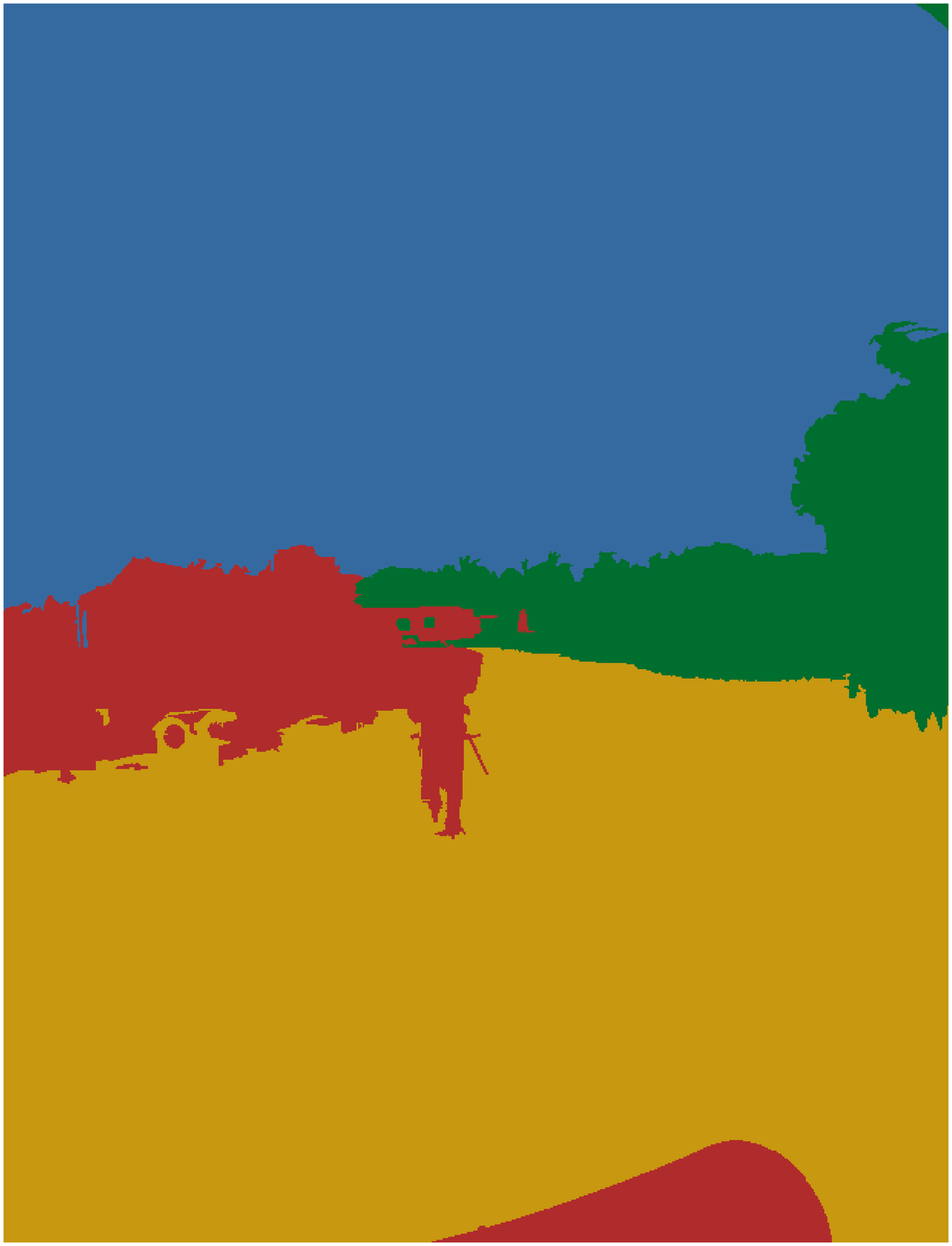} 
\caption{CRF$_{\text{2D}}$}
\end{subfigure} 
\begin{subfigure}[b]{0.194\textwidth}
\includegraphics[clip,trim={0 4cm 0 0},width=\textwidth]{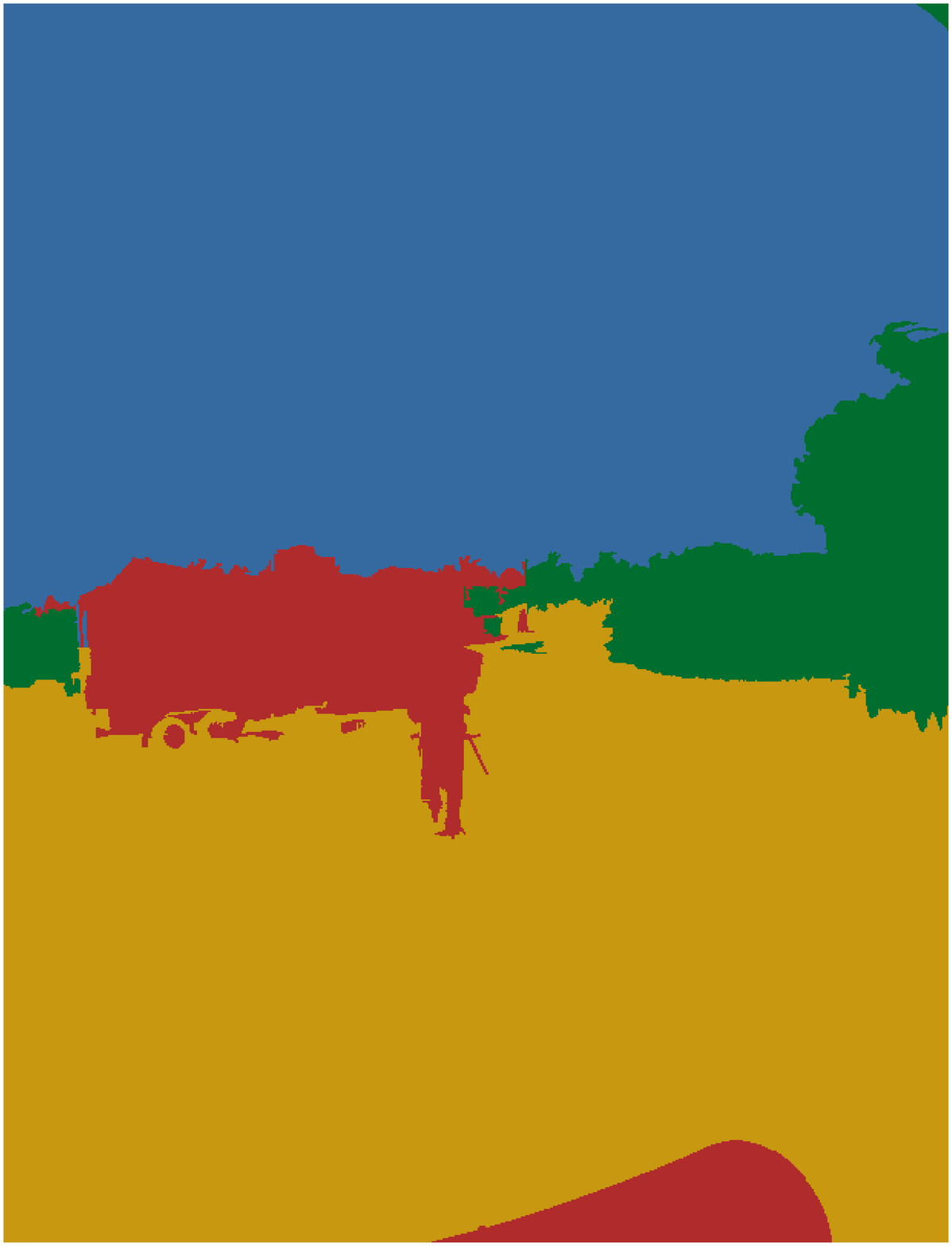} 
\caption{CRF$_{\text{2D-3D,Time}}$}
\end{subfigure}  
\begin{subfigure}[b]{0.194\textwidth}
\includegraphics[clip,trim={6cm 0 0 0},width=\textwidth]{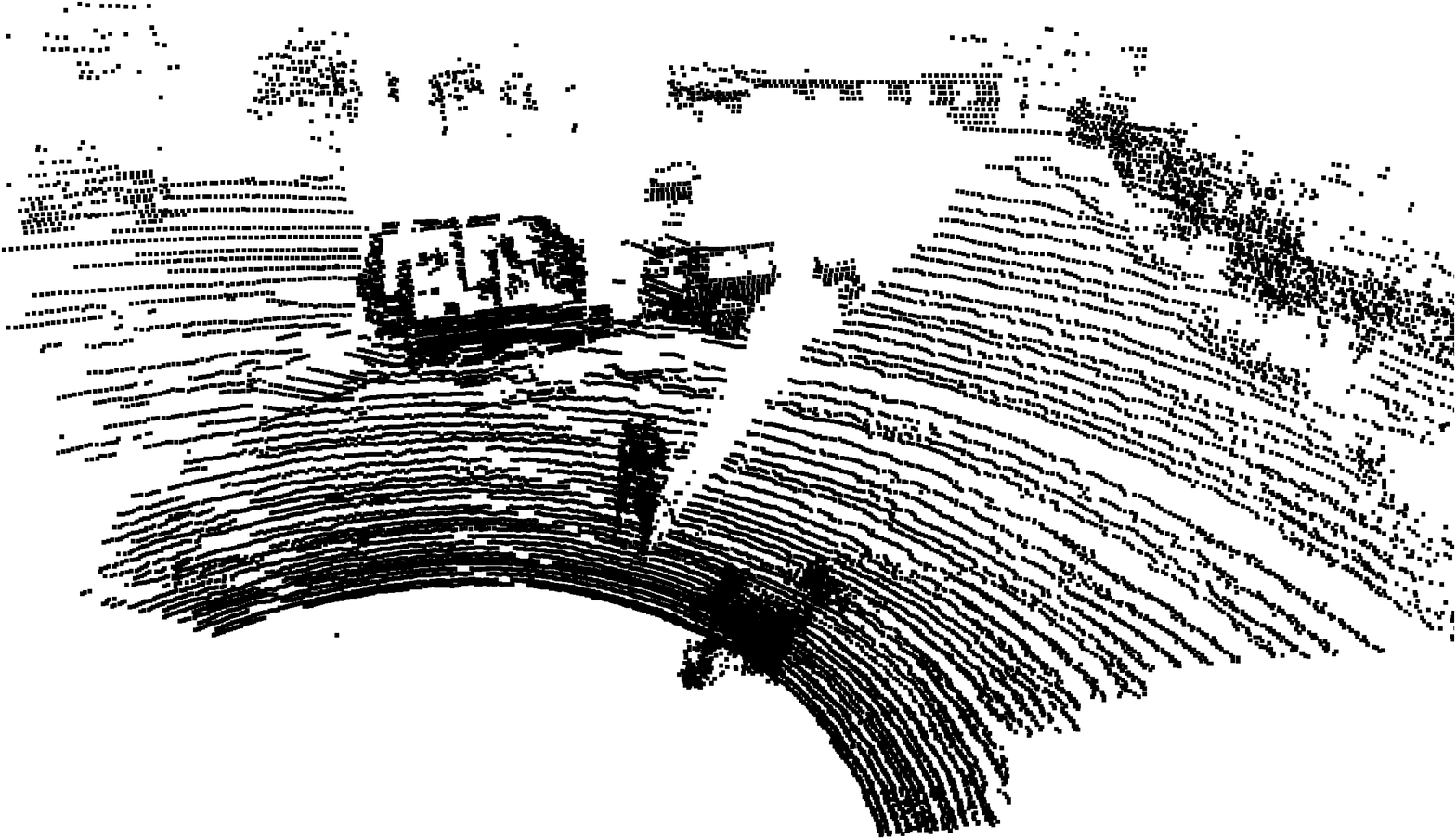} 
\caption{3D point cloud}
\end{subfigure}  
\begin{subfigure}[b]{0.194\textwidth}
\includegraphics[clip,trim={6cm 0 0 0},width=\textwidth]{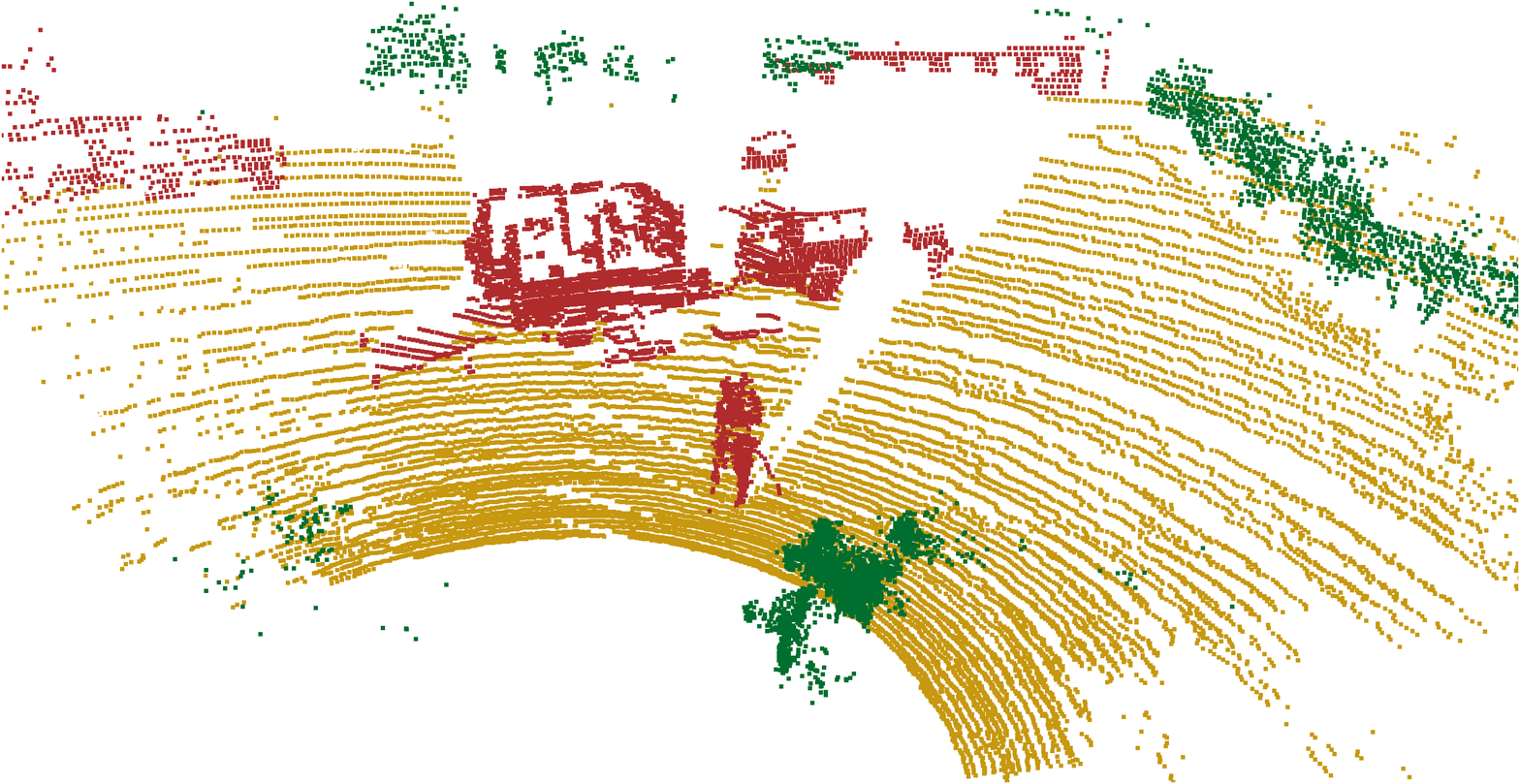} 
\caption{Ground truth}
\end{subfigure}  
\begin{subfigure}[b]{0.194\textwidth}
\includegraphics[clip,trim={6cm 0 0 0},width=\textwidth]{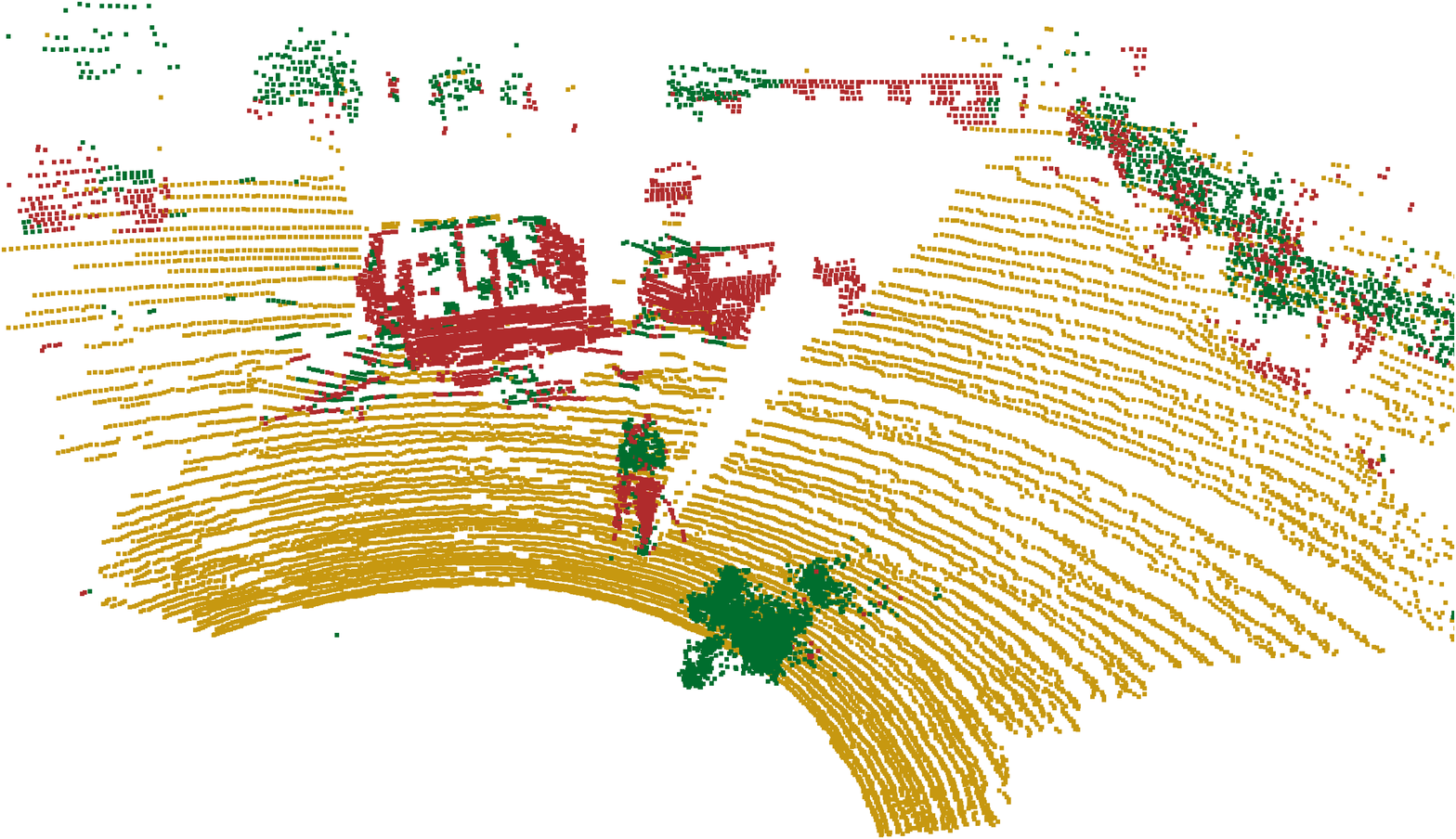} 
\caption{Initial classifier}
\end{subfigure}  
\begin{subfigure}[b]{0.194\textwidth}
\includegraphics[clip,trim={6cm 0 0 0},width=\textwidth]{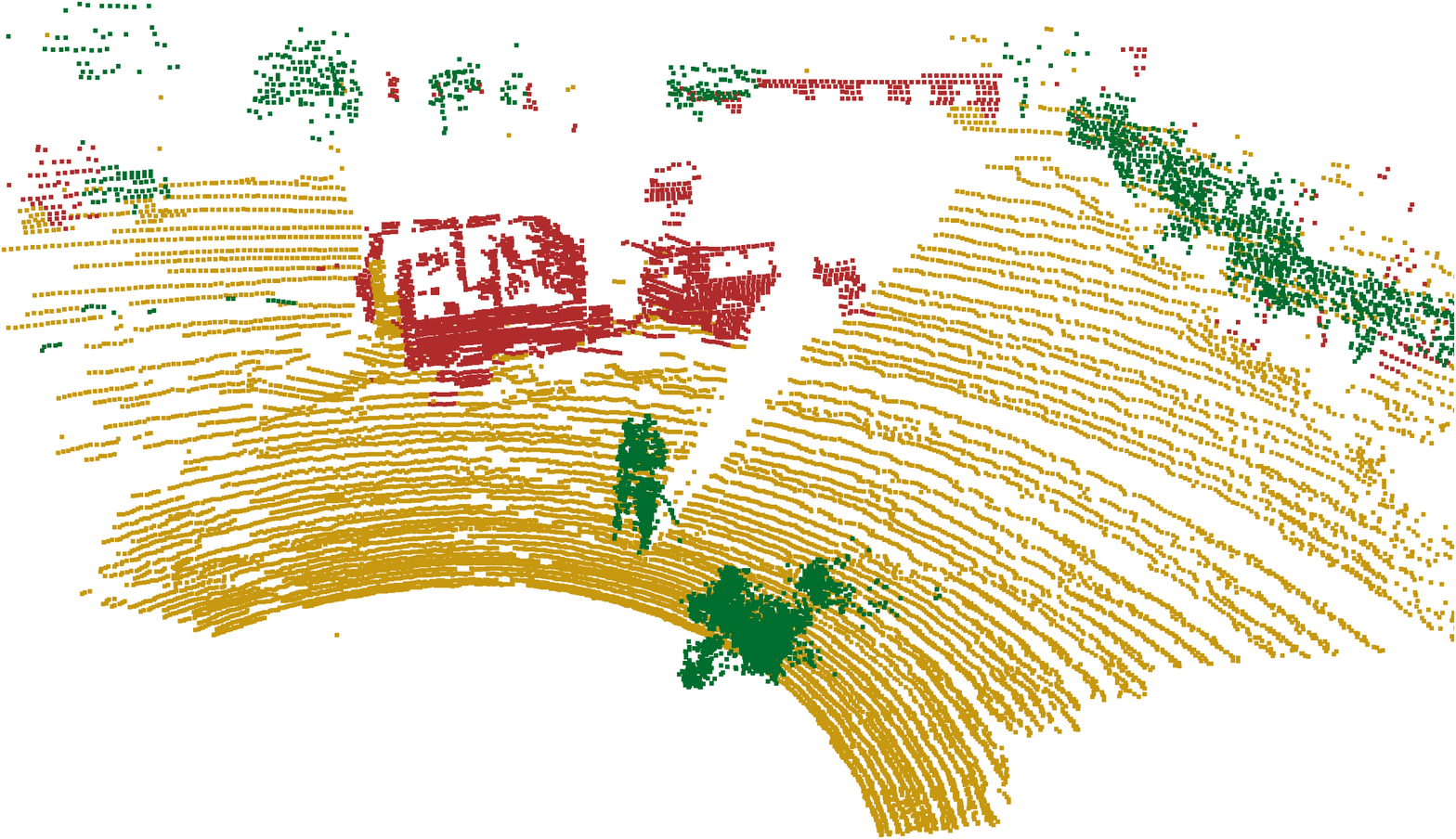} 
\caption{CRF$_{\text{3D}}$}
\end{subfigure} 
\begin{subfigure}[b]{0.194\textwidth}
\includegraphics[clip,trim={6cm 0 0 0},width=\textwidth]{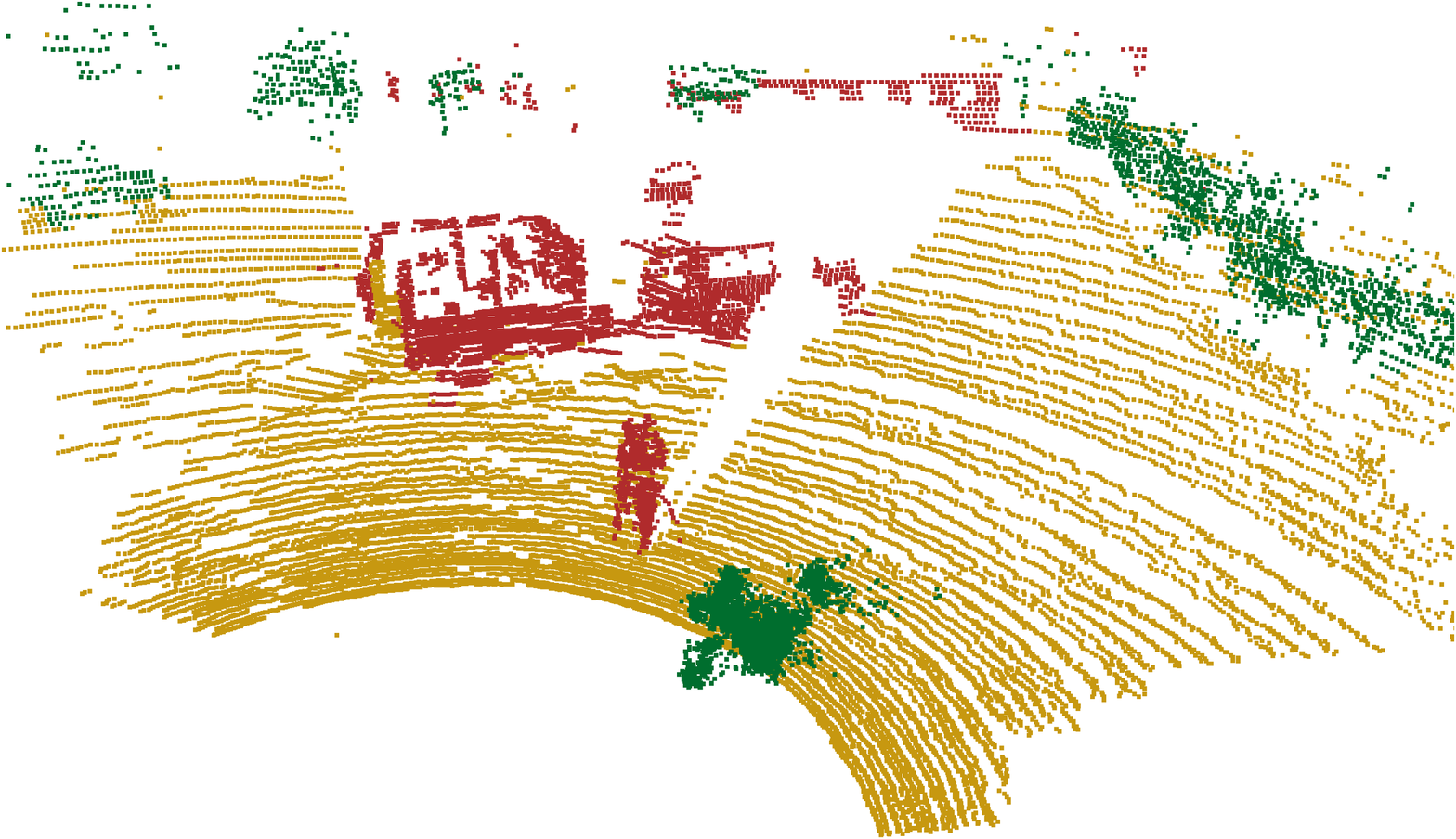} 
\caption{CRF$_{\text{2D-3D,Time}}$}
\end{subfigure}  
\caption{Example results for qualitative evaluation. 
The upper row shows 2D results, and the lower row shows 3D results.
In the first column, the raw image and point cloud are shown for reference, whereas the second column shows the ground truth annotations.
The third row shows initial classifier predictions for the two modalities.
The fourth row shows single-modality results after adding spatial CRF links between neighboring segments.
The fifth row shows the final results with temporal edges and CRF fusion between modalities.
}
\label{fig:4class-example}
\end{figure}

\begin{figure}[t]
\centering
\includegraphics[width=0.9\textwidth]{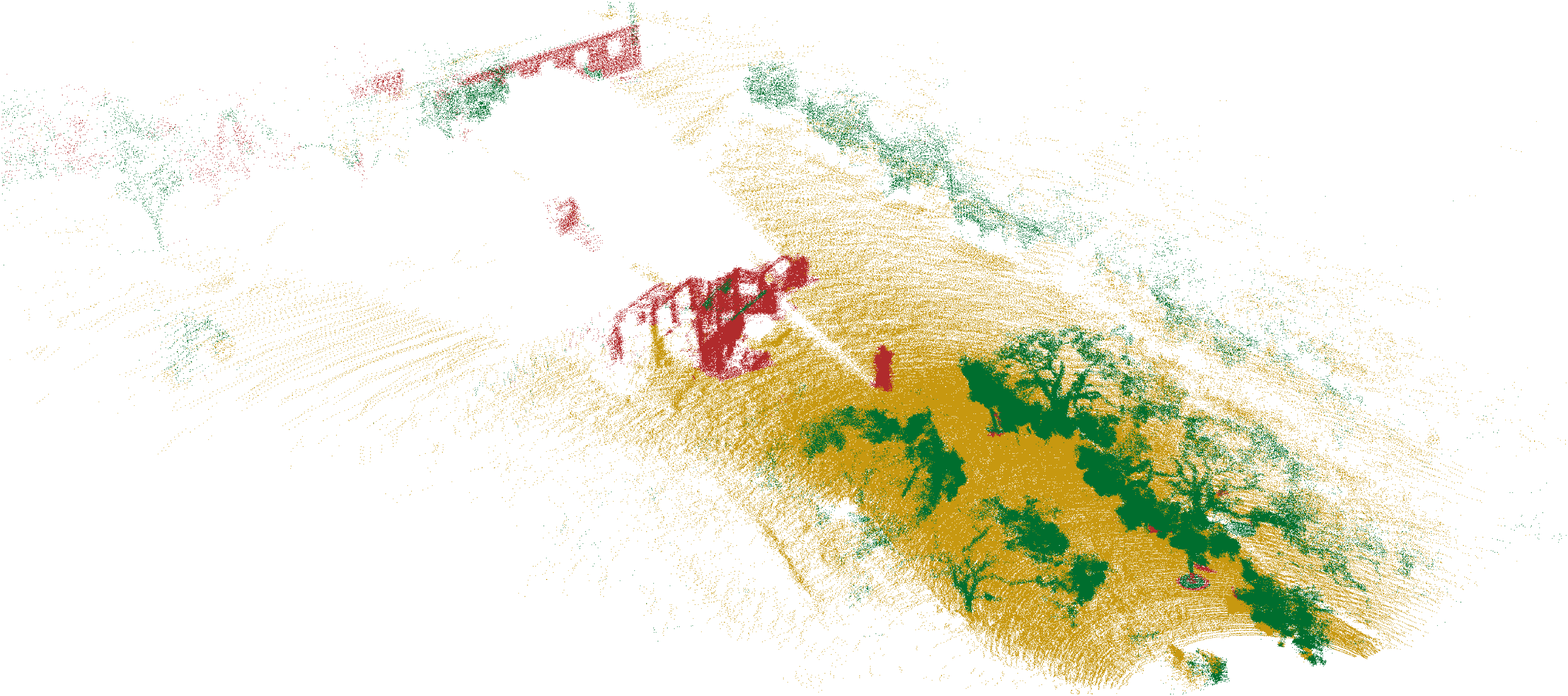} 
\caption{Example of accumulated classification results in 3D, as the robot traversed the end of a row. 
Only the most recent predictions for all 3D points are shown, as \textit{CRF$_{\text{2D-3D,Time}}$} fusion was applied to each frame along the trajectory.}
\label{fig:trajectory}
\end{figure}

Figure~\ref{fig:4class-example} illustrates an example of a corresponding image and point cloud classified with the initial classifiers and with the CRF. 
From (c), it is clear that the initial classification of the image was noisy and affected by saturation problems in the raw image.
When introducing 2D edges in the CRF (d), most of these mistakes were corrected.
Finally, when combined with information from 3D, the CRF was able to correct \textit{vegetation} and \textit{ground} pixels around the trailer (e).
For 3D, some confusion between \textit{vegetation} and \textit{object} occurred in the initial 3D estimate (h), but was mostly solved by introducing 3D edges in the CRF (i).
The person in the front of the scene was mistakenly classified as vegetation when using 3D edges, but this was corrected after fusing with information from 2D (j). 
In some cases, misclassifications in one domain also affected the other. 
In 2D, sensor fusion introduced a misclassification of the trailer ramp (e), which was seen as \textit{ground} by the initial 3D classifier.
Most likely, this happened because the ramp was flat and essentially served the purpose of connecting the ground and the trailer.

For the same example section of the dataset presented in Figure~\ref{fig:4class-example}, Figure~\ref{fig:trajectory} illustrates the accumulated classification results in 3D, of a trajectory along the end of a row, driving from the bottom right towards the center of the image.
This section was chosen as a compact area with many examples of the different classes.
The accumulated point cloud was generated by applying the \textit{CRF$_{\text{2D-3D,Time}}$} fusion method to each frame and then transforming all 3D points from the sensor frame into the world frame.
To generate the figure, the most recent class prediction within any $0.5m$ radius is chosen to represent the region.
That is, if a point $\mathbf{p}_1$ was given class label $c_1$ at time $t_1$, then this inherited class label $c_2$ of point $\mathbf{p}_2$ at time $t_2$ if $\lvert \mathbf{p}_2-\mathbf{p}_1 \rvert \leq 0.5m$ and $t_2>t_1$.
Effectively, this corresponds to always trusting the most recent prediction of the algorithm.
The figure illustrates how the algorithm was able to correctly classify most of the environment in 3D, as the robot traversed a row.
However, a few classification mistakes were made between \textit{vegetation} and \textit{object}.
In the lower right corner, some parts of the mango trees were mistaken for \textit{object}, and in the center of the image the edges of trailer roof door was mistaken for \textit{vegetation}.

\begin{figure}[t]
\centering
\begin{subfigure}[b]{0.245\textwidth}
\includegraphics[width=\textwidth]{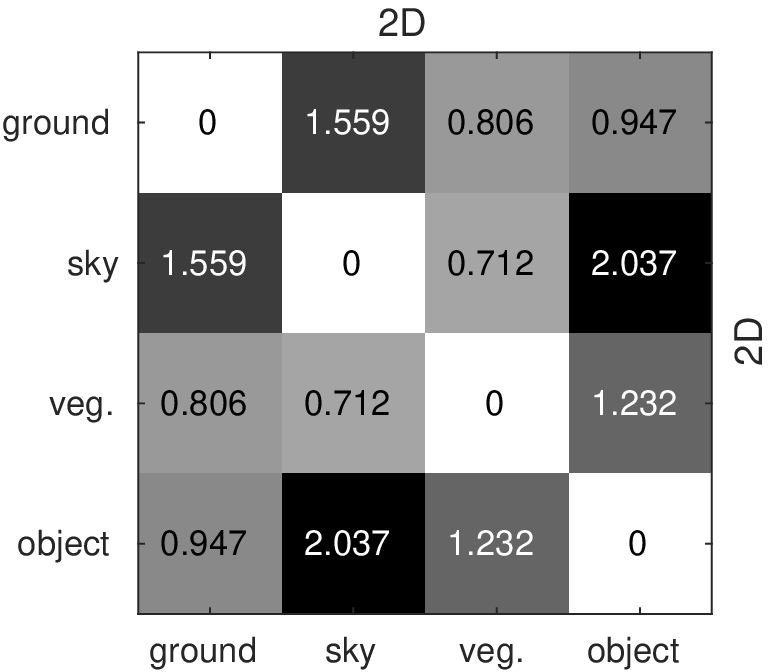} 
\caption{$\mathbf{w}_p^{\text{2D}}$}
\label{crf-weights-2D}
\end{subfigure}
\begin{subfigure}[b]{0.245\textwidth}
\includegraphics[width=\textwidth]{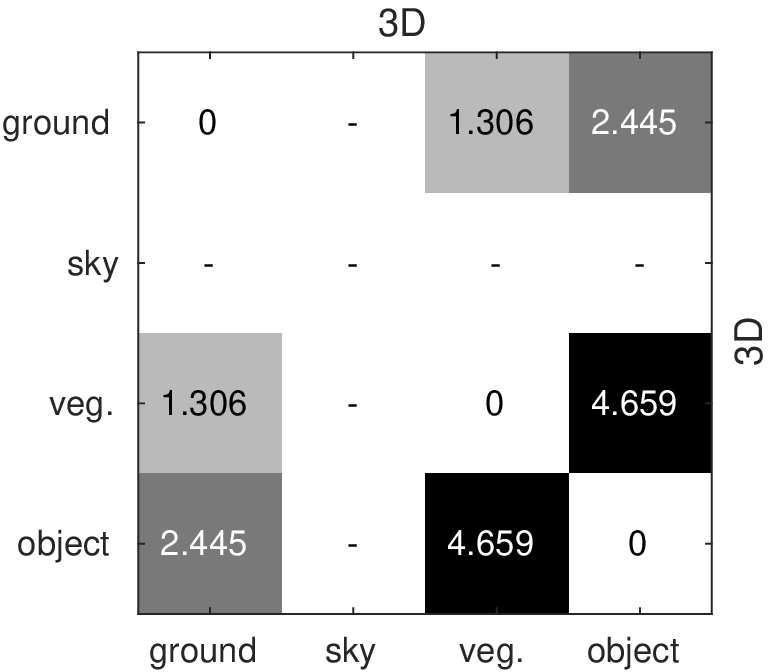} 
\caption{$\mathbf{w}_p^{\text{3D}}$}
\label{crf-weights-3D}
\end{subfigure}   
\begin{subfigure}[b]{0.245\textwidth}
\includegraphics[width=\textwidth]{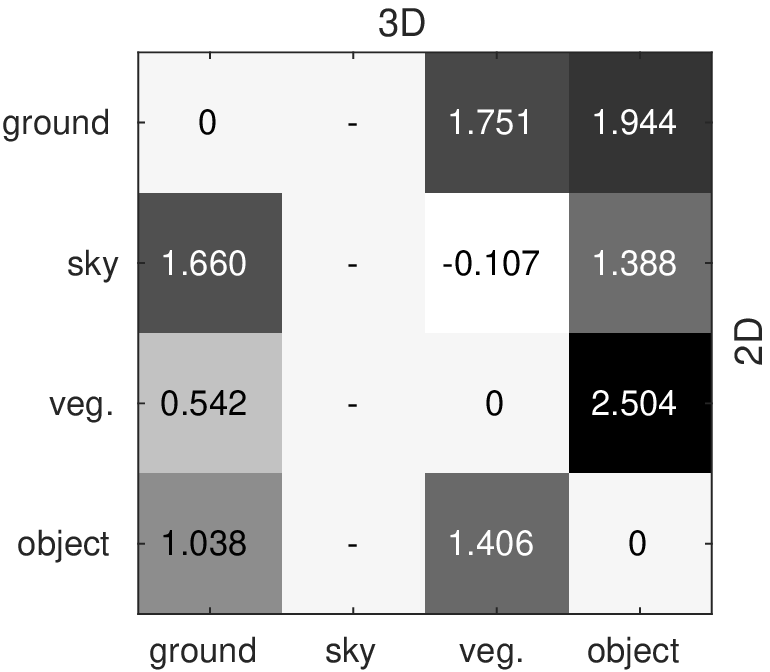} 
\caption{$\mathbf{w}_p^{\text{2D-3D}}$}
\label{crf-weights-2D3D}
\end{subfigure}  
\begin{subfigure}[b]{0.245\textwidth}
\includegraphics[width=\textwidth]{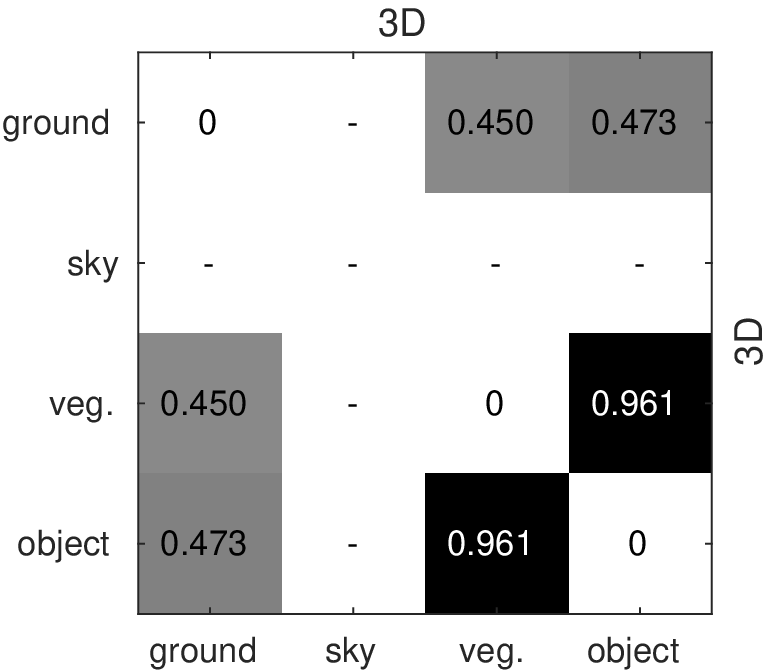} 
\caption{$\mathbf{w}_p^{\text{Time}}$}
\label{crf-weights-rec}
\end{subfigure}  
\caption{Learned CRF weight matrices (described in section~\ref{pairwise_potentials}) averaged over cross-validation folds. 
High weights correspond to rare occurences and vice versa.
The entries in (\subref{crf-weights-2D}) are cost weights for assigning different labels to neighboring 2D segments.		
A cost weight of $0$ is used for assigning two neighboring segments the same label, whereas a high cost weight of $2.037$ is used for assigning two neighbors different labels (\textit{object} and \textit{sky}).
(\subref{crf-weights-3D}), (\subref{crf-weights-2D3D}), and (\subref{crf-weights-rec}) show similar weight matrices for neighboring 3D segments, 2D-3D fusion, and temporal edges.}
\label{fig:crf-weights}
\end{figure}

Figure~\ref{fig:crf-weights} visualizes the learned CRF weights averaged over the 5 cross-validation folds.
As explained in section~\ref{pairwise_potentials}, (a), (b), and (d) are symmetric, whereas (c) is asymmetric.
For visualization purposes, we trained the CRF without bias weights, as these would introduce another matrix for each potential and thus make the interpretation of the weights more difficult.
Figure~\ref{crf-weights-2D} shows the weight matrix for neighboring 2D segments.
The weights depend on the certainty of the initial classifier and how often adjacent superpixels with different labels appeared in the training set.
\textit{ground}-\textit{object} and \textit{vegetation}-\textit{sky} appeared often and thus had low weights, whereas \textit{ground}-\textit{sky} and \textit{object}-\textit{sky} were rare and therefore were penalized with high weights.
Intuitively, this makes sense, as vegetation often separates the ground from the sky in agricultural fields.
Figure~\ref{fig:4class-example} illustrates how \textit{object} superpixels in the middle of the sky in (c) were corrected by the CRF to \textit{sky} in (d).
This was directly caused by a high value of $w_p^{\text{2D}}\left(\textit{ground},\textit{sky}\right)$ and multiple adjacent \textit{sky} neighbors.

Figure~\ref{crf-weights-3D} shows the weight matrix for neighboring 3D segments.
Here, the highest weight was for \textit{object}-\textit{vegetation}.
Structurally, these classes were difficult to distinguish with the initial classifier as seen in Figure~\ref{fig:4class-example} (h). 
However, when introducing spatial links in the CRF, most ambiguities were solved as seen in (i).

Figure~\ref{crf-weights-2D3D} shows the weight matrix for the 2D-3D fusion.
As mentioned in section~\ref{2D-3D_edges}, the matrix is asymmetric, as we allow different interactions between the 2D and 3D domain.
The interpretation of these weights is considerably more complex than $\mathbf{w}_p^{\text{2D}}$ and $\mathbf{w}_p^{\text{3D}}$, since the weights incorporate calibration and synchronization errors between the lidar and the camera, and since overlapping 2D and 3D segments intuitively cannot have different class labels.
However, a notable outlier was the weight for \textit{sky}-\textit{vegetation} which was negative.
The only apparent explanation for this is a calibration error between the two modalities.
Physically, a 2D segment cannot be \textit{sky} if an overlapping 3D segment has observed it.
Therefore, label inconsistencies near border regions of \textit{vegetation} and \textit{sky} will cause the CRF weight to decrease.

Figure~\ref{crf-weights-rec} shows the weight matrix for temporal edges.
The weights were all rather small and thus matched the small increase in classification performance when introducing temporal edges.
As the weights describe the cost of assigning different labels at the approximate same 3D location, we see the same trend as for neighboring 3D segments in Figure~\ref{crf-weights-3D}. 

\subsection{Binary and Multiclass Classification}
Due to the physics of the camera and the lidar, the two modalities perceive significantly different characteristics of the environment.
The lidar is ideal for distinguishing elements that are geometrically unique, whereas the camera is ideal for distinguishing visual uniqueness.
The choice of classes therefore highly affects the resulting improvement with the CRF fusion stage.

In this section, we compare binary and multiclass classification scenarios.
The first scenario maps all annotated labels except \textit{ground} to a common \textit{non-ground} class, such that $x_i = \left\{\textit{ground},\textit{non-ground}\right\}$. 
The second scenario is the same 4-class scenario as presented above.
For convenience, the results from Table~\ref{fig:4class-example} are replicated in this section.

\begin{table}[t]
\caption{Classification results for binary and multiclass scenarios.} \label{multiclass-table}
\begin{center}
\begin{tabular}{lcccc}
\toprule
& \multicolumn{2}{c}{2-class scenario}         & \multicolumn{2}{c}{4-class scenario}     \\
\cmidrule(lr){2-3} \cmidrule(lr){4-5}				& mean IoU			& accuracy			& mean IoU			& accuracy			\\
\midrule
2D, initial   										& 0.914				& 0.956				& 0.685				& 0.900				\\
2D, CRF$_{\text{2D}}$	 							& 0.933				& 0.966				& 0.742				& 0.937				\\
2D, CRF$_{\text{2D-3D}}$ 							& \textbf{0.938}	& \textbf{0.969}	& 0.756				& \textbf{0.943}				\\
2D, CRF$_{\text{2D-3D,Time}}$ 						& \textbf{0.938}	& \textbf{0.969}	& \textbf{0.758}	& \textbf{0.943}	\\
\midrule
3D, initial   										& 0.927				& \textbf{0.963}	& 0.678				& 0.881				\\
3D, CRF$_{\text{3D}}$	 							& \textbf{0.928}	& \textbf{0.963}	& 0.748				& 0.923				\\
3D, CRF$_{\text{2D-3D}}$ 							& 0.901				& 0.949				& 0.827				& 0.943				\\
3D, CRF$_{\text{2D-3D,Time}}$ 						& 0.900				& 0.949				& \textbf{0.842}	& \textbf{0.948}	\\
\bottomrule
\end{tabular}
\end{center}
\end{table}

Table~\ref{multiclass-table} presents the results for the 2D and 3D domains separately. 
For both the $2$- and $4$-class scenarios, \textit{CRF$_{\text{2D}}$} and \textit{CRF$_{\text{3D}}$} improved the initial classification results.
However, for $2$-class classification, the \textit{CRF$_{\text{2D-3D}}$} fusion only improved 2D performance, whereas 3D performance actually declined.
This is because the geometric classifier (lidar) is good at detecting ground points, and thus can single-handedly distinguish \textit{ground} and \textit{non-ground}.
For $4$-class classification, however, the CRF fusion introduced improvements in both 2D and 3D.
This was caused by the geometric classifier being less discriminative for \textit{vegetation} and \textit{object}, since both classes were represented by obstacles protruding from the ground.
Therefore, color and texture cues from the visual classifier could help separate the classes.

To summarize, for binary classification of \textit{ground} vs. \textit{non-ground}, individual sensing modalities and classifiers seemed sufficient, as sensor fusion did not provide significant improvements.
However, for the $4$-class scenario, the two sensors indeed complemented each other, as sensor fusion showed significant classification improvements in both 2D and 3D.

\begin{figure}[p]
\centering
\begin{subfigure}[t]{\textwidth}
\hfill
\includegraphics[width=0.3\textwidth]{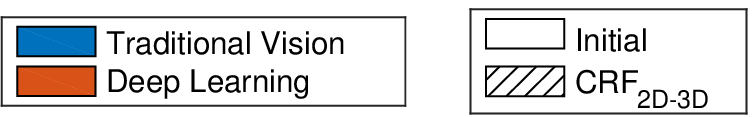} 
\end{subfigure}
\begin{subfigure}[h]{0.467\textwidth}
\includegraphics[width=\textwidth]{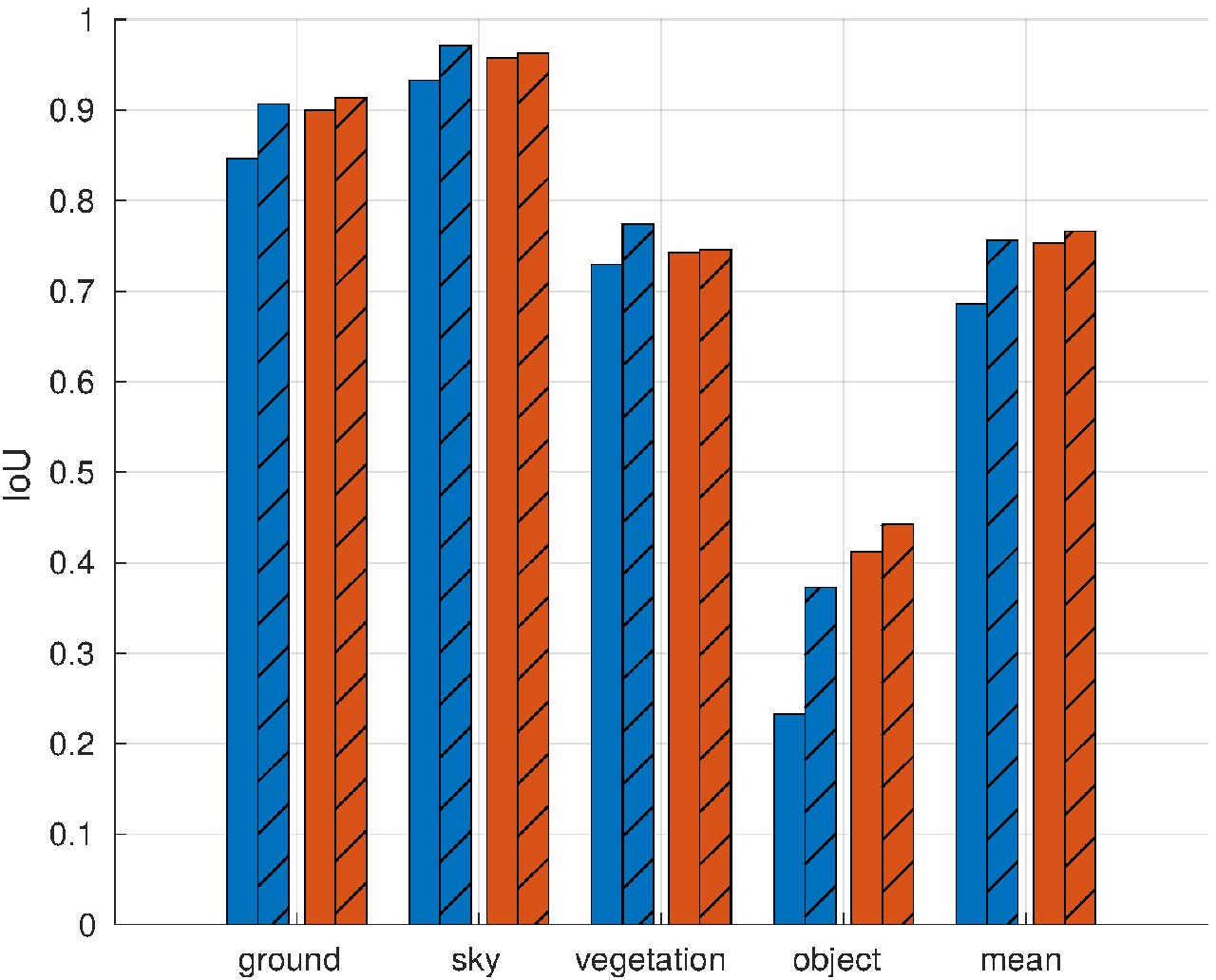} 
\caption{2D results for each class}
\label{2d_classifier_comparison_classes_2D}
\end{subfigure}
\hspace{0.01cm}
\begin{subfigure}[h]{0.467\textwidth}
\includegraphics[width=\textwidth]{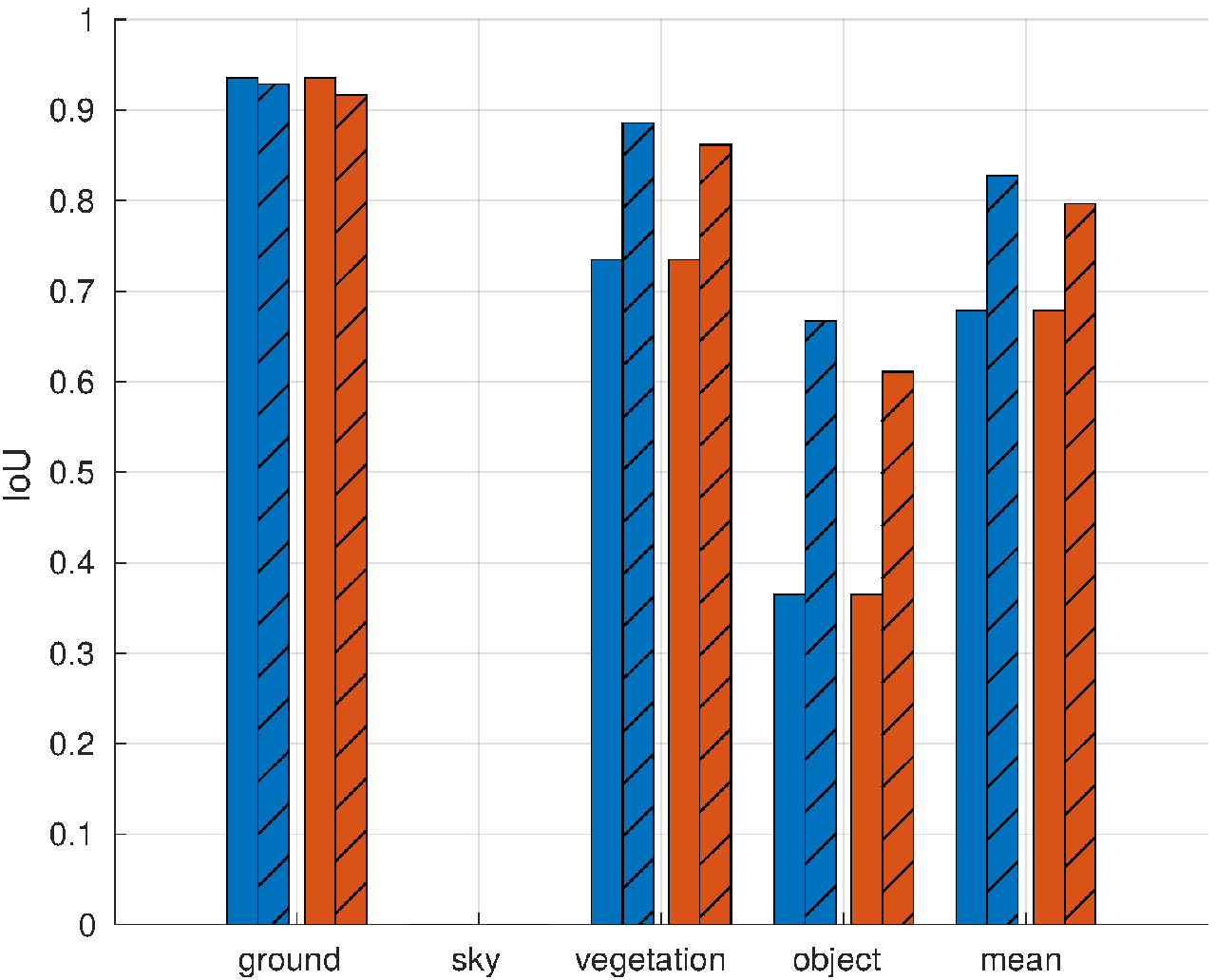} 
\caption{3D results for each class}
\label{2d_classifier_comparison_classes_3D}
\end{subfigure} 
\begin{subfigure}[h]{0.467\textwidth}
\includegraphics[width=\textwidth]{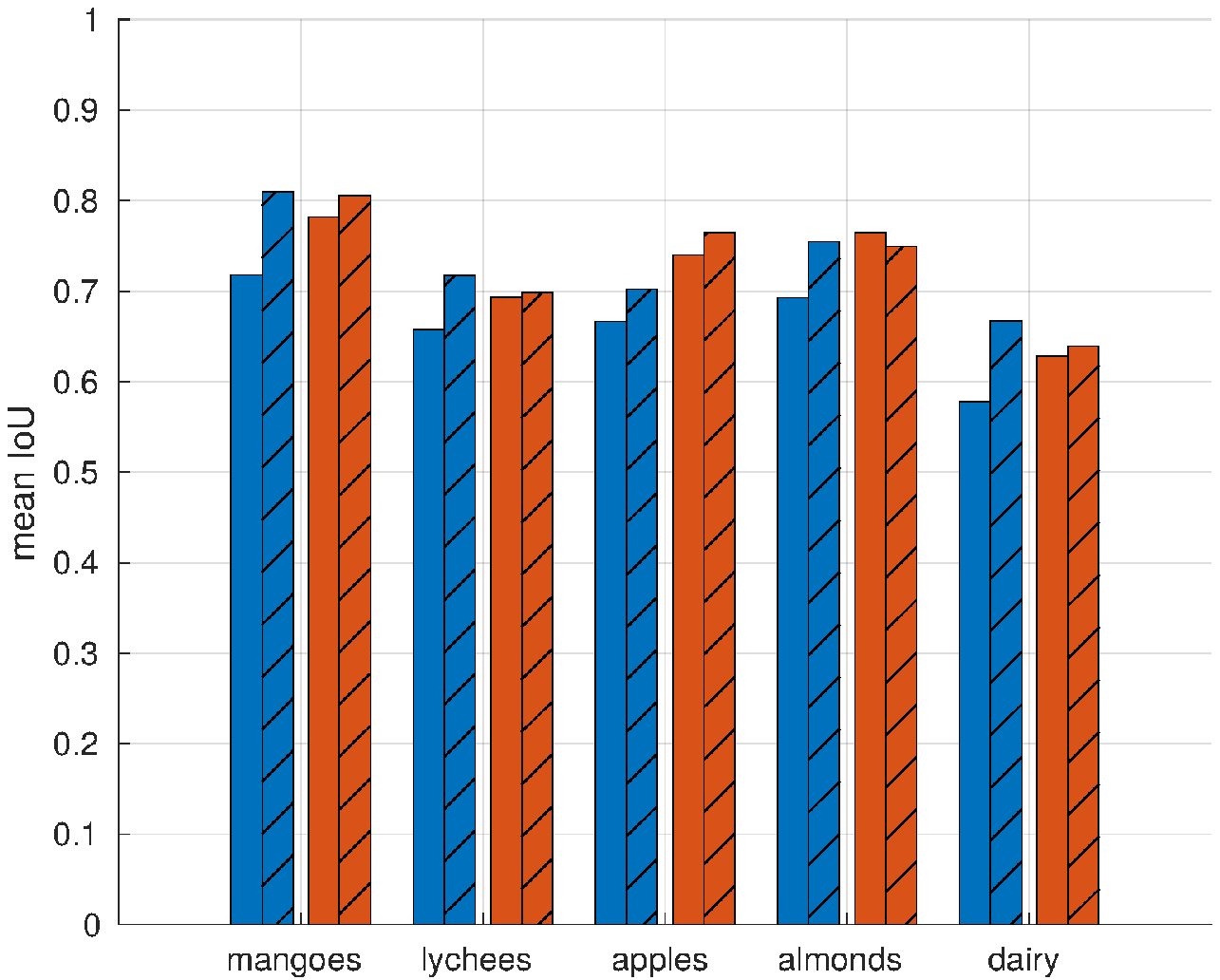} 
\caption{2D results for each dataset}
\label{2d_classifier_comparison_datasets_2D}
\end{subfigure}
\hspace{0.01cm}
\begin{subfigure}[h]{0.467\textwidth}
\includegraphics[width=\textwidth]{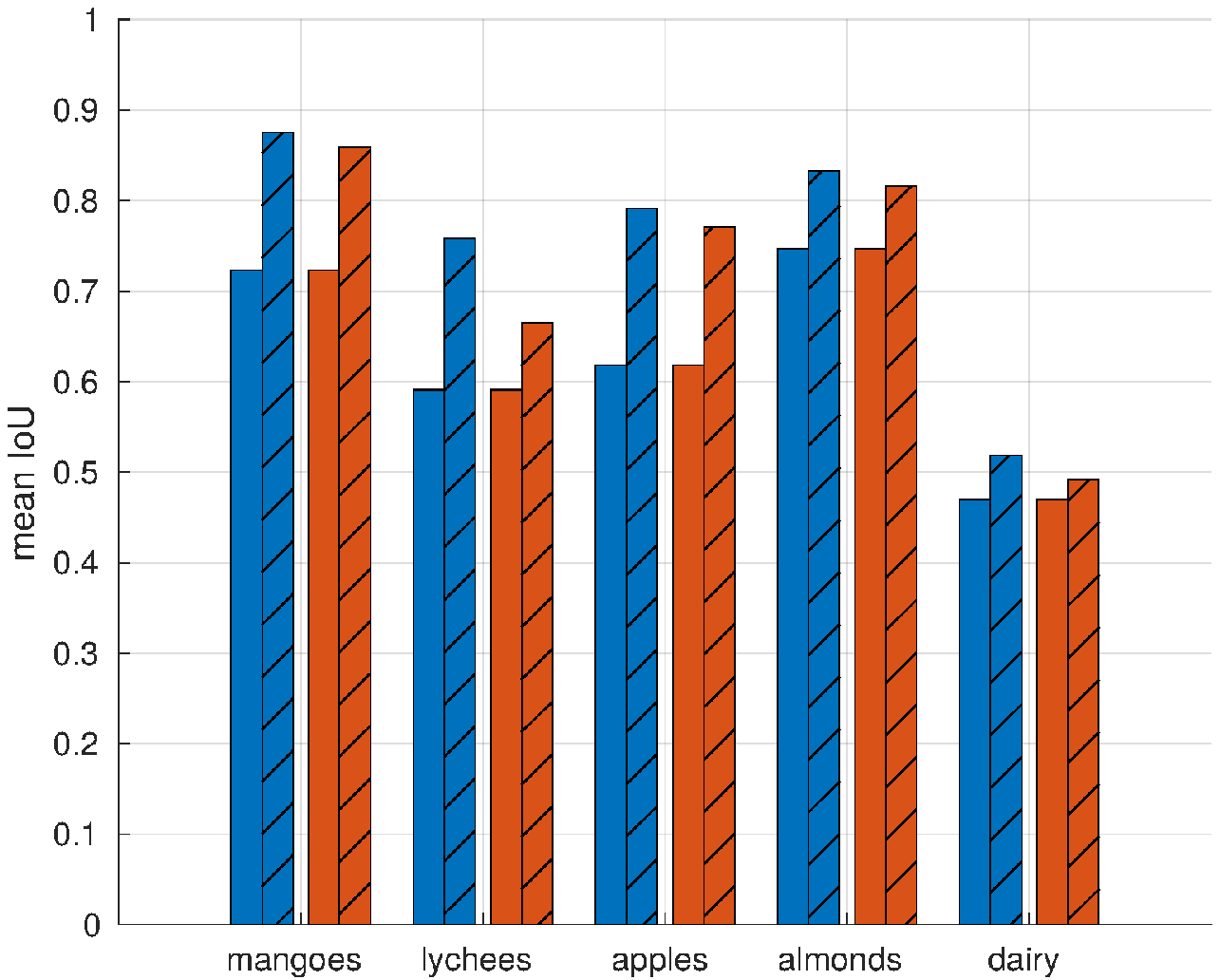} 
\caption{3D results for each dataset}
\label{2d_classifier_comparison_datasets_3D}
\end{subfigure}   
\caption{Evaluation of traditional vision (blue) vs. deep learning (red) before and after sensor fusion. 
Filled bars denote initial classification results, whereas hatched bars show classification results after sensor fusion (CRF$_{\text{2D-3D}}$).
(\subref{2d_classifier_comparison_classes_2D}) and (\subref{2d_classifier_comparison_classes_3D}) show 2D and 3D results for the 4 different classes, whereas (\subref{2d_classifier_comparison_datasets_2D}) and (\subref{2d_classifier_comparison_datasets_3D}) show 2D and 3D results for the 5 different datasets.} 
\label{fig:2d_classifier_comparison}
\end{figure}

\subsection{2D Classifiers}
\label{2D_image_features}
\begin{figure}[p]
\centering
\begin{subfigure}[t]{0.32\textwidth}
\includegraphics[clip,trim={0 7cm 0 9cm},width=\textwidth]{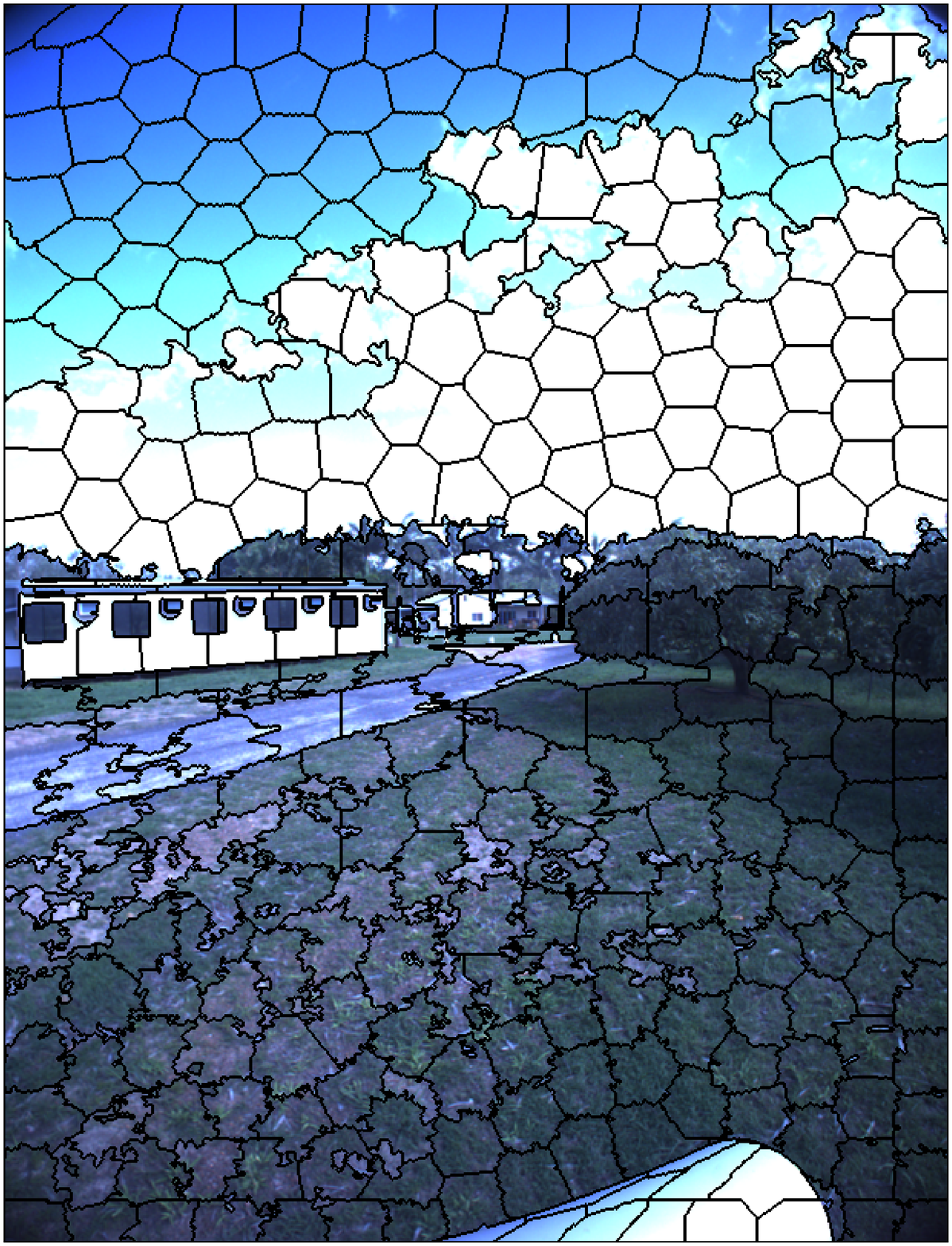} 
\caption{Raw image with 2D superpixels}
\label{fig:2D_DL_boundaries_a}
\vskip 0pt
\end{subfigure}
\begin{subfigure}[t]{0.32\textwidth}
\includegraphics[clip,trim={0 7cm 0 9cm},width=\textwidth]{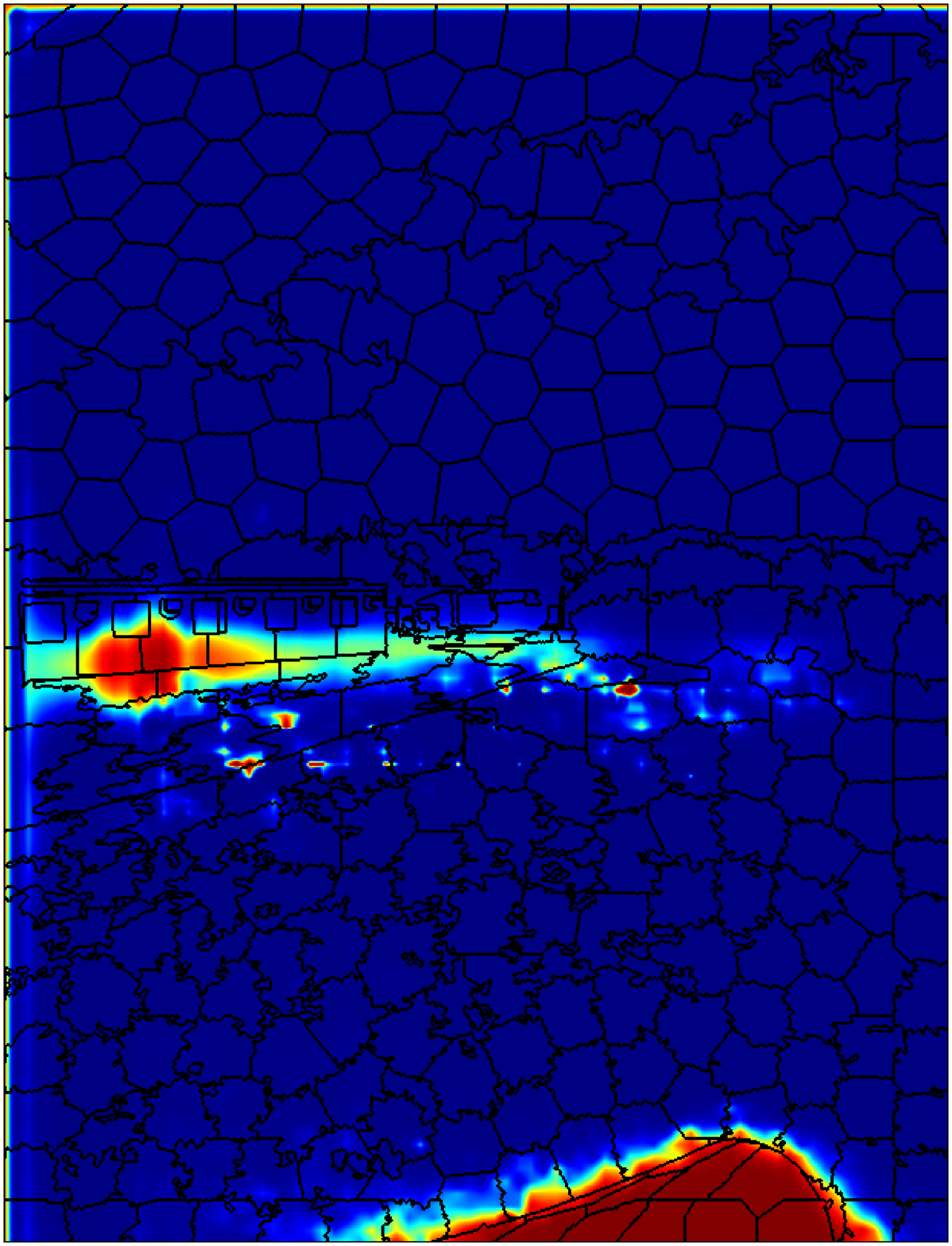} 
\caption{\textit{object} pixel probabilities from deep learning classifier}
\label{fig:2D_DL_boundaries_b}
\vskip 0pt
\end{subfigure}
\begin{subfigure}[t]{0.32\textwidth}
\includegraphics[clip,trim={0 7cm 0 9cm},width=\textwidth]{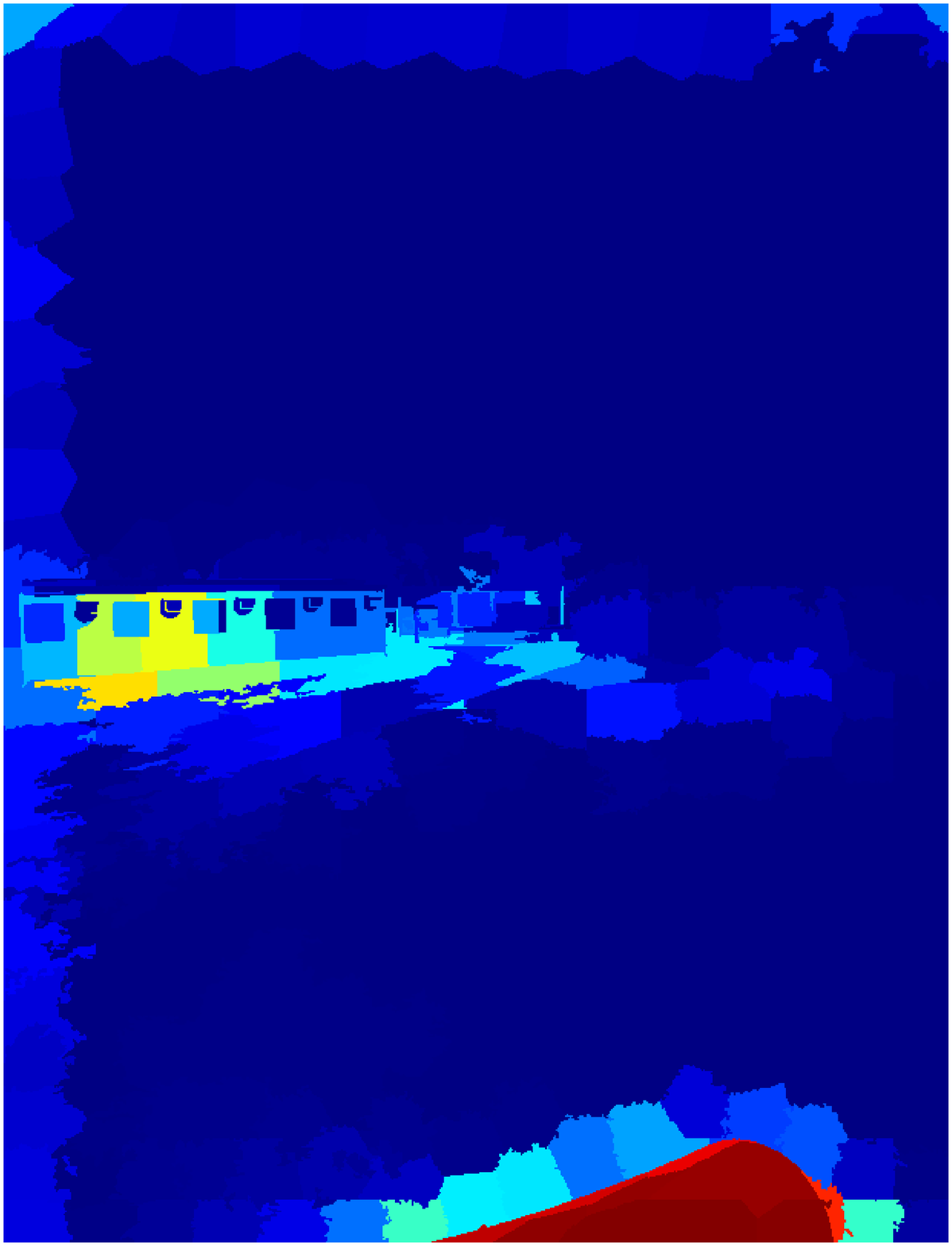} 
\caption{Superpixelized \textit{object} heatmap}
\label{fig:2D_DL_boundaries_c}
\end{subfigure}
\caption{Example of blurred classification boundaries with deep learning classifier.
(\subref{fig:2D_DL_boundaries_a}) shows the raw image, (\subref{fig:2D_DL_boundaries_b}) shows pixel-wise \textit{object} class probabilities from deep learning, and (\subref{fig:2D_DL_boundaries_c}) shows the resulting superpixel probabilities after 2D segmentation.
(\subref{fig:2D_DL_boundaries_b}) and (\subref{fig:2D_DL_boundaries_c}) use pseudo-coloring for visualizing low (dark blue) and high (dark red) probability estimates.}
\label{fig:2D_DL_boundaries}
\end{figure}

As described in section~\ref{2D}, a traditional vision pipeline with hand-crafted features was compared to a deep learning approach with self-learned features.
Figure~\ref{fig:2d_classifier_comparison} compares the two approaches before and after applying the CRF fusion.
(a) and (b) show 2D and 3D results for each class, respectively.
Filled bars denote initial classification results, whereas hatched bars show classification results after sensor fusion (CRF$_{\text{2D-3D}}$).
In Figure~\ref{2d_classifier_comparison_classes_2D}, we see that the initial classification results for deep learning were significantly better than for traditional vision with a mean IoU of $75.3\%$ vs. $68.5\%$.
The most significant difference was for the \textit{object} class.
Here, deep learning had a clear advantage, since the CNN was pre-trained on an extensive dataset with a wide collection of object categories.
When fused with 3D data, however, traditional vision and deep learning reached more similar mean IoUs of $75.6\%$ and $73.6\%$, respectively.
The improvement in classification performance was thus much higher for the traditional vision pipeline than for deep learning.
If we look at 3D classification in Figure~\ref{2d_classifier_comparison_classes_3D}, the best mean IoU was obtained when fusing with the traditional vision pipeline.
Here, a mean IoU of $82.7\%$ was achieved, compared to $79.6\%$ for deep learning.
A possible explanation for this is that deep learning is extremely good at recognition, since it uses a hierarchical feature representation and thus incorporates contextual information around each pixel.
However, in doing this, a large receptive field (spatial neighborhood) is utilized, which along with multiple max-pooling layers reduces the classification accuracy near object boundaries~\citep{Chen2014}. 
Figure~\ref{fig:2D_DL_boundaries} illustrates the phenomenon with blurred classification boundaries in (b) and the resulting \textit{object} heatmap in (c) after applying the superpixel segmentation as described in section~\ref{2D}.
Since the fusion stage of the CRF assumes exact localization in both 2D and 3D, the phenomenon may explain the rather small improvement when fusing 3D predictions with 2D deep learning-based predictions.

Figure~\ref{fig:2d_classifier_comparison} (c) and (d) show 2D and 3D results for each dataset, respectively.
Here, we see the same tendency that deep learning was superior in 2D in its initial classification for all datasets.
However, when fused with 3D data, the two methods basically performed equally well.
Traditional vision was better for \textit{lychees} and \textit{dairy}, deep learning was better for \textit{apples}, and they were almost equal for \textit{mangoes} and \textit{almonds}.

To summarize, when evaluating individual performance, deep learning was better than traditional vision.
However, when applying a CRF and fusing with lidar, the two methods gave similar results.
The CRF was thus able to compensate for the shortcomings in the traditional vision approach.

\begin{figure}[p]
\centering
\begin{subfigure}[t]{\textwidth}
\hfill
\includegraphics[width=0.2\textwidth]{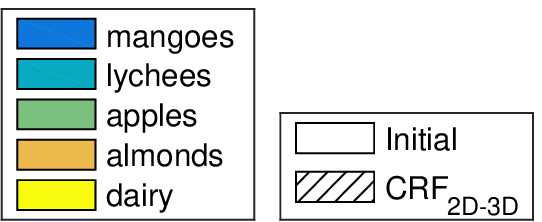} 
\end{subfigure}
\begin{subfigure}[t]{0.493\textwidth}
\includegraphics[width=\textwidth]{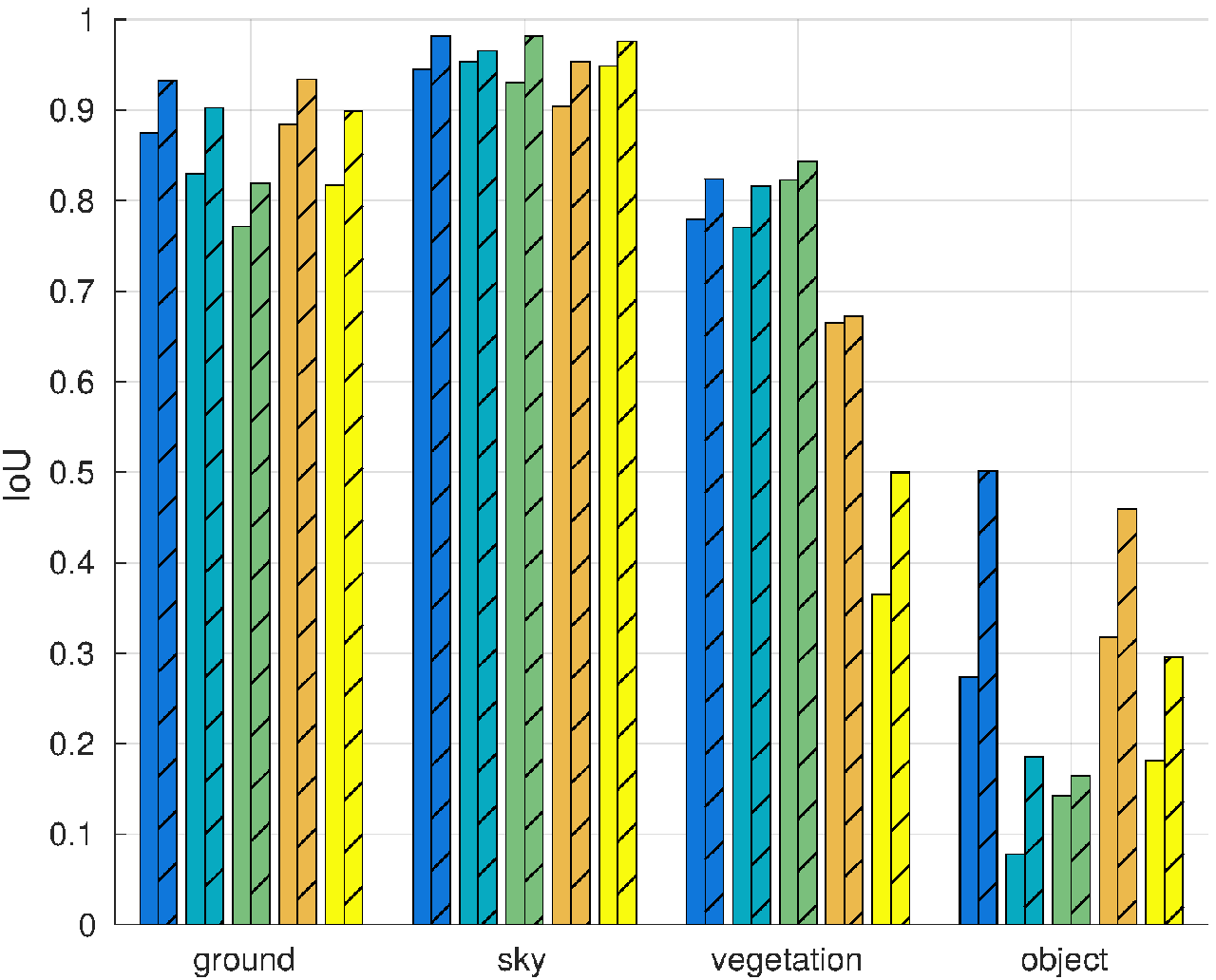} 
\caption{2D}
\label{fig:domain-data_2D}
\end{subfigure}
\hspace{0.01cm}
\begin{subfigure}[t]{0.493\textwidth}
\includegraphics[width=\textwidth]{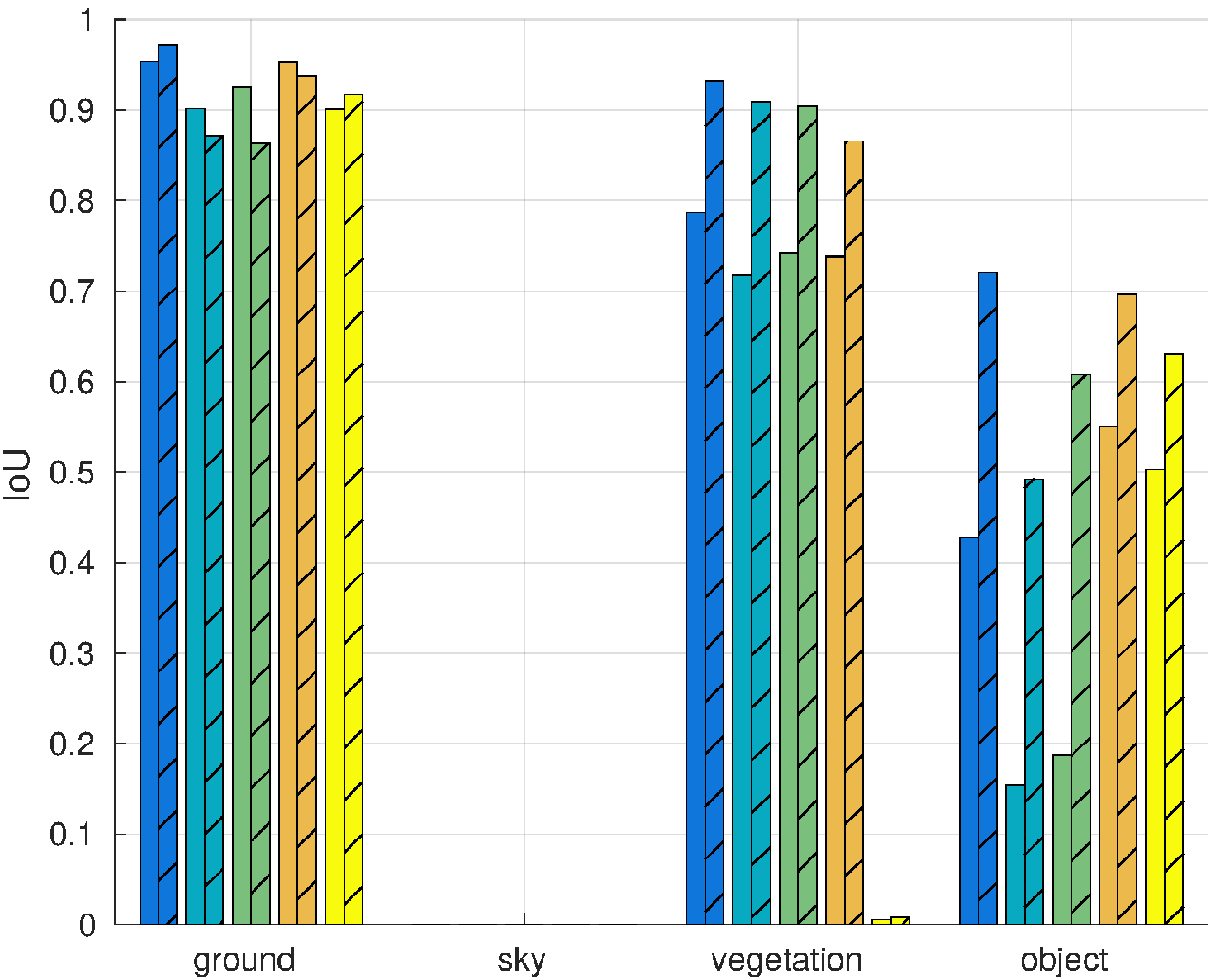} 
\caption{3D}
\label{fig:domain-data_3D}
\end{subfigure}    
\caption{Classification results across the 4 object classes and 5 datasets before and after sensor fusion.
Filled bars denote initial classification results, whereas hatched bars show classification results after sensor fusion (CRF$_{\text{2D-3D}}$).
The 5 different colors denote the 5 different datasets.}
\label{fig:domain-data}
\end{figure}

\subsection{Domain Adaptation}
In section~\ref{results_overview}, we evaluated the combined classification results over all datasets.
In this section, we revisit and break apart these results into separate datasets.
In this way, we can evaluate the transferability of features and classifiers across datasets and across classes.
Within machine learning and transfer learning, this is generally referred to as domain adaptation.
This will allow us to answer a question like:
how well do the features and classifiers trained on the combined imagery from \textit{mangoes}, \textit{lychees}, \textit{apples} and \textit{almonds} generalize to recognize a new scenario, such as \textit{vegetation} in the \textit{dairy} dataset?
Figure~\ref{fig:domain-data} compares the classification performances in 2D and 3D separately across object classes and datasets.
Filled bars denote initial classification results, whereas hatched bars show classification results after sensor fusion (CRF$_{\text{2D-3D}}$).

Figure~\ref{fig:domain-data_2D} shows that for 2D, features and classifiers transferred quite well for \textit{ground} and \textit{sky}, possibly due to a combination of limited variation in visual appearance and an extensive amount of training data.
However, a larger variation was observed across datasets for \textit{vegetation} and \textit{object}.
For the \textit{vegetation} class, the \textit{dairy} dataset had the lowest 2D classification performance.
This might be because the mean distance to the tree line was much higher for the dairy paddock than for the orchards, as seen in Figure~\ref{fig:datasets}.
The visual appearance varies with distance, and especially features describing texture are affected by associated changes in scale and resolution.
For the \textit{object} class, a large variation in 2D performance was seen across all datasets.
This is most likely due to the large variation in \textit{object} appearances, as the class covered humans, vehicles, buildings, and animals.
Also, as listed in Table~\ref{dataset-table}, not all datasets included examples of buildings and animals.
Figure~\ref{fig:domain-data_3D} shows that for 3D, the features and classifiers transferred well for \textit{ground}, but experienced the same tendencies in variation for \textit{vegetation} and \textit{object} as seen in 2D.
For the \textit{vegetation} class, the \textit{dairy} dataset had an IoU close to $0\%$.
This is likely due to the mean distance to the tree line which was outside the range of the lidar.
Only a few 3D points within range were labeled \textit{vegetation}, and since the classification performance decreases with distance, most of these were misclassified.
For the \textit{object} class, a large variation in 3D performance was seen across all datasets, similar to 2D.
However, the initial 3D classifier performed better than 2D, suggesting slightly better transferability for 3D features and classifiers.

Evaluating the transferability of CRF weights, we compared the increase in classification performance across the different datasets (the difference between filled and hatched bars of the same color in Figure~\ref{fig:domain-data}).
Generally, the CRF weights transferred well across all datasets in both 2D and 3D.
However, in 3D, the \textit{ground} class experienced both increases and decreases.
Difference in terrain roughness could possibly explain this phenomenon.

To summarize, with minor exceptions, features and classifiers transferred well across the \textit{ground}, \textit{sky}, and \textit{vegetation} classes for all datasets in both 2D and 3D.
For these classes, the CRF framework is able to deliver performance increases even when training data is supplied from different environments, which is reasonable given that the appearance of these classes to some degree is independent of the specific site.
For the \textit{object} class, however, features and classifiers transferred poorly in both 2D and 3D, resulting in considerable performance variations across datasets.
This was likely caused by limited training data covering the large variation in geometry and appearance within the \textit{object} class, as cows were only present in the \textit{dairy} dataset, tractors in \textit{mangoes}, iron bars in \textit{lychees}, etc.

\subsection{Domain Training}
For all the above evaluations, $5$-fold cross-validation was used corresponding to the $5$ different datasets (domains).
That is, when testing on e.g. \textit{apples}, no data from \textit{apples} were used to train the algorithms.
In this section, we compare this approach with two less challenging scenarios, where training data are available from the same domain.

As the \textit{almonds} dataset consisted of recordings from two separate days, we split it into \textit{almonds-day1} and \textit{almonds-day2} with 16 and 15 annotated frames, respectively.
In the first scenario, we limited the dataset to include \textit{almonds} only.
That is, when testing on \textit{almonds-day1}, we trained on the \textit{almonds-day2} dataset, and vice versa.
This meant that the training data represented the exact same environment, although captured on a different day.
In the second scenario, we combined domain training with domain adaptation.
That is, when testing on \textit{almonds-day1}, we trained on the \textit{almonds-day2} dataset plus all the remaining datasets. 
In this way, a small portion of the training data represented the same environment as the test setup.

\begin{figure}[!p]
\centering
\begin{subfigure}[t]{\textwidth}
\hfill
\includegraphics[width=0.25\textwidth]{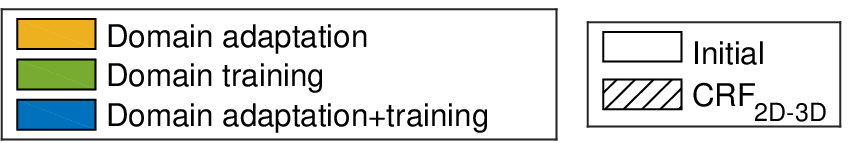} 
\end{subfigure}
\begin{subfigure}[t]{0.493\textwidth}
\includegraphics[width=\textwidth]{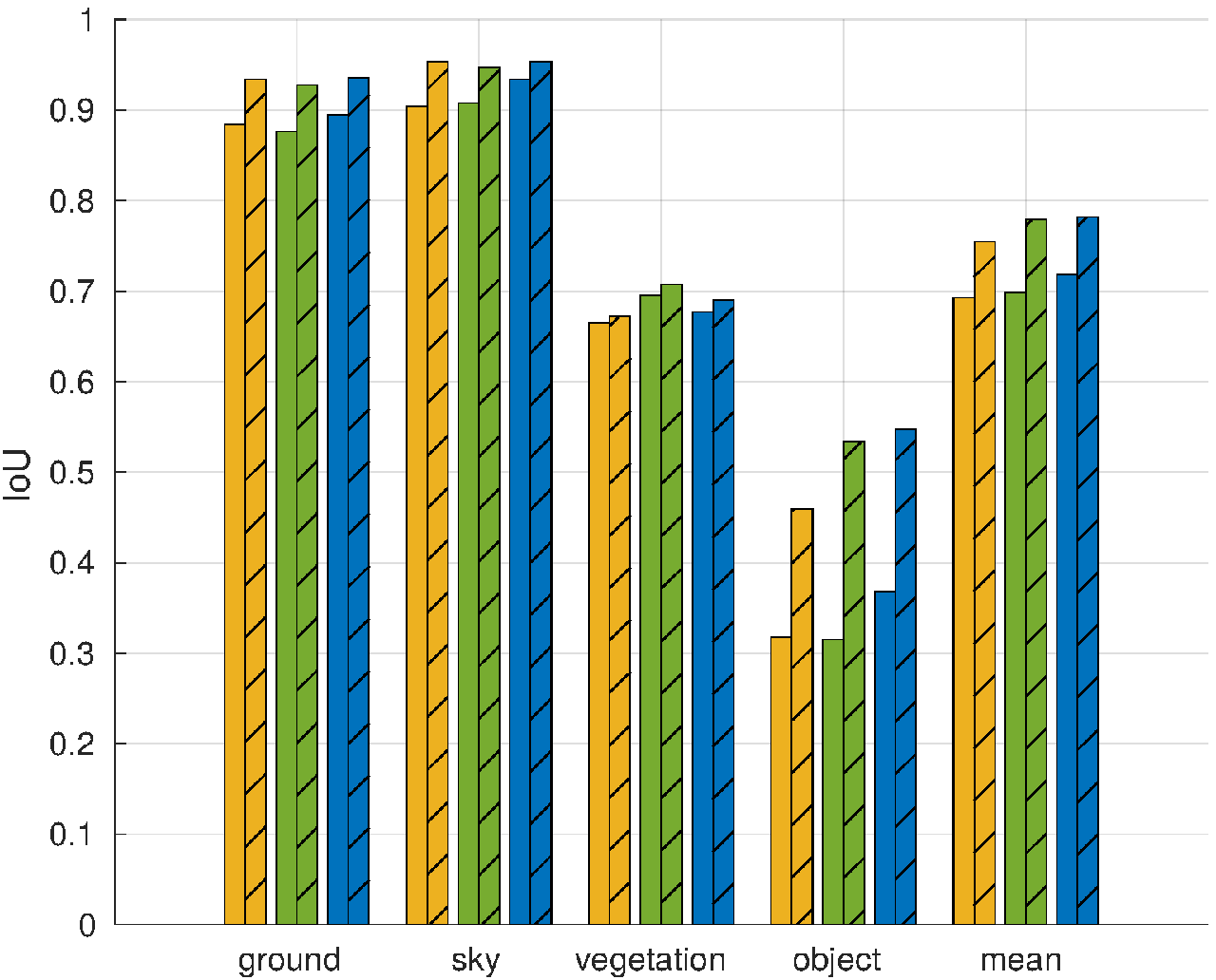} 
\caption{2D}
\label{fig:domain-training_2D}
\end{subfigure}
\hspace{0.01cm}
\begin{subfigure}[t]{0.493\textwidth}
\includegraphics[width=\textwidth]{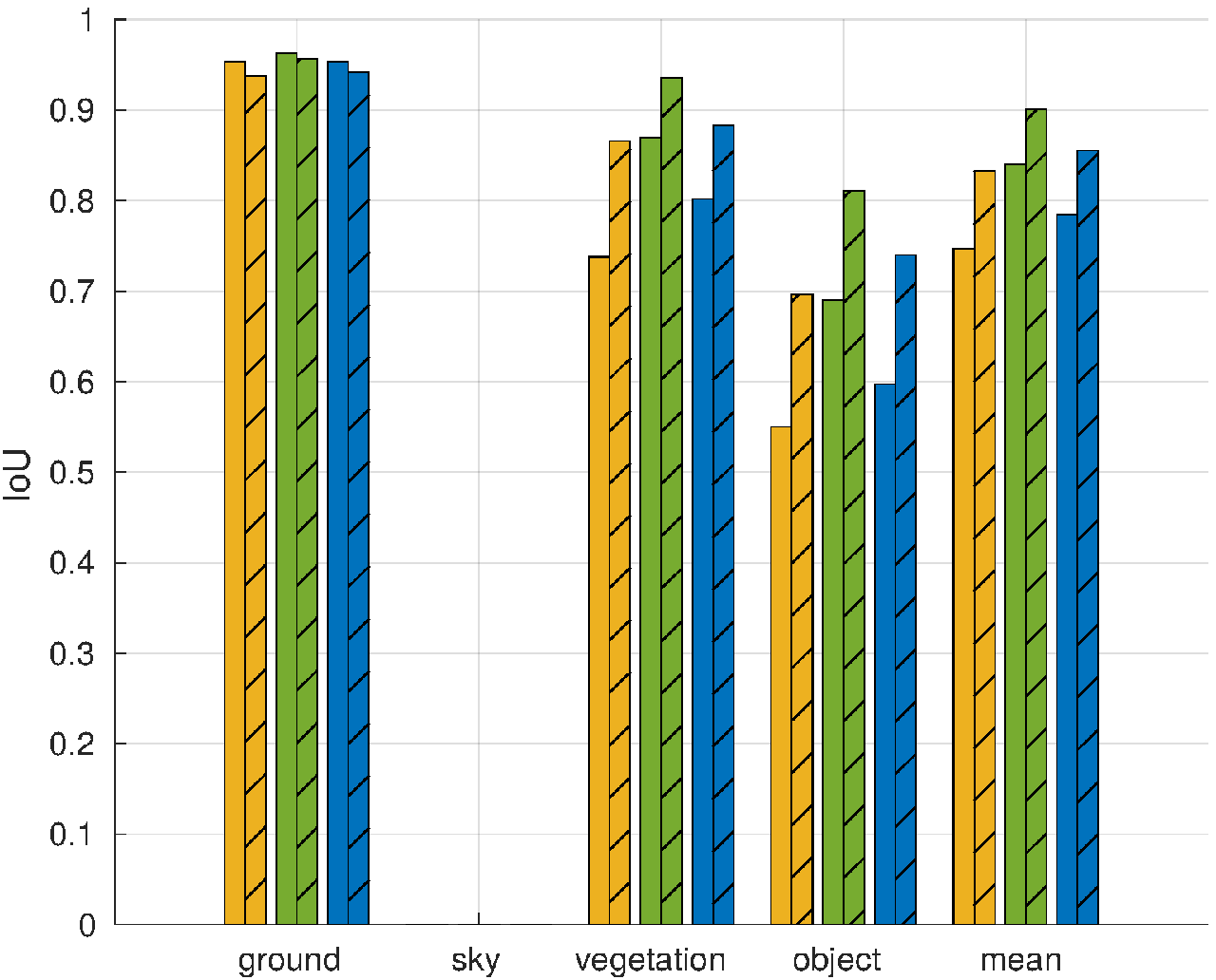} 
\caption{3D}
\label{fig:domain-training_3D}
\end{subfigure}    
\caption{Classification results across the 4 object classes with domain adaptation, domain training, and domain adaptation+training on the \textit{almonds} dataset.
Filled bars denote initial classification results, whereas hatched bars show classification results after sensor fusion (CRF$_{\text{2D-3D}}$).
Domain adaptation (orange) includes training data from all other domains than \textit{almonds}. 
Domain training (green) includes training data from \textit{almonds} only.
And finally, domain adaptation+training (blue) includes training data from all domains including \textit{almonds}.}
\label{fig:domain-training}
\end{figure}

Figure~\ref{fig:domain-training} shows a comparison of 2D and 3D performance between domain adaptation, domain training, and domain adaptation+training on the \textit{almonds} dataset.
Filled bars denote initial classification results, whereas hatched bars show classification results after sensor fusion (CRF$_{\text{2D-3D}}$).
For all methods, we calculated the average performance over the entire \textit{almonds} dataset.
Only the training data varied between the three methods.
Note that the brown bars for domain adaptation were simply copied from \textit{almonds} in Figure~\ref{fig:domain-data} to ease the comparison.

Figure~\ref{fig:domain-training_2D} shows that for 2D, the three methods only resulted in minor performance variations for \textit{ground}, \textit{sky}, and \textit{vegetation}.
This is a surprising result, as the appearance of both \textit{ground} and \textit{vegetation} in the \textit{almonds} dataset differed quite significantly from the remaining datasets as shown in Figure~\ref{fig:datasets-almonds}.
Despite this, the 2D classifiers successfully discriminated the classes even when no training data from the specific environment were available (domain adaptation).
For the \textit{object} class, however, significant improvements were introduced with the two domain training strategies.
For domain training, initial IoU was similar to domain adaptation, while fusion with 3D resulted in an increase of $7.4\%$.
For domain adaptation+training, initial IoU was increased by $5.0\%$, while fusion with 3D resulted in an increase of $8.8\%$.
Again, this underlines that the large variation in \textit{object} appearances required more training data for the initial classifier.
However, fusion with 3D seemed to circumvent this requirement.
Therefore, although initial 2D mean IoU was better for domain adaptation+training, 3D fusion compensated for the differences and made both domain training approaches perform equally well.

Figure~\ref{fig:domain-training_3D} shows that for 3D, the \textit{ground} class was relatively unaffected by domain training.
This is most likely due to the ground geometry of \textit{almonds} being very similar to that of \textit{mangoes}, \textit{lychees}, and \textit{apples}.
The \textit{vegetation} and \textit{object} classes, on the other hand, both experienced large improvements, especially for domain training.
For the \textit{object} class, domain training increased initial IoU by $14.0\%$, while fusion with 3D resulted in an increase of $11.5\%$.
Domain adaptation+training, however, gave smaller increases of $4.8\%$ and $4.4\%$, respectively.
The same trend was seen for \textit{vegetation}, where domain training gave increases of $13.1\%$ and $7.1\%$, whereas domain adaptation+training gave increases of $6.4\%$ and $1.8\%$.
This could be caused by the particular \textit{vegetation} geometry of the \textit{almonds} dataset.
From Figure~\ref{fig:datasets-almonds}, it is clear that the \textit{almonds} dataset was the only dataset captured during flowering, whereas \textit{mangoes}, \textit{lychees}, and \textit{apples} were all captured during fruit-set.
The 3D lidar data therefore varied significantly for \textit{vegetation} due to differences in geometry and 3D point densities.
Domain adaptation, without knowledge of the specific geometry of \textit{vegetation}, therefore gave the lowest 3D performance.
Domain training, on the other hand, gave the best performance, as the 3D classifier was trained specifically on \textit{vegetation} geometry of \textit{almonds} during flowering.
Finally, domain adaptation+training was in between.
Possibly, adding training data from other domains may have made the features of \textit{vegetation} and \textit{object} less seperable.
That is, if the two classes were easily distinguished from a small amount of \textit{almonds} training data, the addition of more (possibly overlapping) feature examples from other domains may have partially contaminated the training set.
This could suggest that including training data from the same season (flowering or fruit-set) may be more important for 3D classification than including it from the same environment (\textit{mangoes}, \textit{lychees}, \textit{apples}, or \textit{almonds}).

To summarize, domain training generally showed better performance than domain adaptation.
Including training data from the same environment thus gave slightly better 2D performance and considerably better 3D performance.
The performance increases were class-dependant, such that classes with large inter-domain variation in appearance and geometry benefited significantly from domain training.
Additionally, combining domain adaptation with domain training introduced more training data and could thus potentially improve performance, as was seen in 2D.
However, as seen in 3D, the performance could also decrease.
This indicates that domain adaptation should only be considered when the feature distributions of the source and target domains are similar.
In this context, the specific season of the dataset may be as important as the specific environment.

\subsection{Timing}
As stated in the introduction, the proposed method is online applicable and thus uses only current and previous information gathered with the perception system of the robot.
This contrasts the fusion algorithm of~\cite{Namin2015} from which it was adapted, since their method uses information acquired over the entire traversal of the scene. 
Their method, therefore, does not distinguish between past, present, and future view points.

Using a combination of libraries from MATLAB and C++, our method has been optimized for research flexibility and not processing speed.
In order to run the proposed method in real-time, further optimization effort would be required, which is outside the scope of this paper.

Table~\ref{timing-table} lists the average computation times for the processing pipeline.
Combining 2D and 3D computations makes the average processing time per frame 8.5 seconds.
This is dominated by segmentation and feature extraction in 2D.
For 2D segmentation, a GPU implementation of SLIC could be used to reduce the processing time down to $\sim$20 ms \citep{gSLICr_2015}.
Similarly, 2D feature extraction and classification could be sped up by applying an inference-optimized semantic segmentation deep neural network such as Enet~\citep{paszke2016enet}.
For 3D, the order of feature extraction, classification, and segmentation could be changed to perform feature extraction and classification on supervoxels instead of each point.
This would significantly speed up feature extraction and classification, although potentially also reduce the accuracy.
Finally, CRF inference, which is currently done in MATLAB, could be sped up by using a C++ toolkit.
8.5 seconds in total is thus plausible to be sped up to realtime, by a combination of replacing MATLAB with C++, plus the use of GPU and parallelization.

\begin{table}[h]
\caption{Average computation times per frame for the processing pipeline.
The timing test was performed on a Fujitsu H730 laptop with a 2.7 GHz Intel Core i7 CPU and 16 GB of memory.} \label{timing-table}
\begin{center}
\begin{tabular}{lll}
\toprule
& 2D & 3D \\
\midrule
Segmentation & 1.4 s & 0.4 s \\
Feature extraction & 4.5 s & 0.9 s \\
Initial classification & 0.3 s & 0.6 s \\
CRF$_{\text{2D-3D,Time}}$ & \multicolumn{2}{c}{0.4 s} \\
\bottomrule
\end{tabular}
\end{center}
\end{table}

\section{Conclusion}
This paper has presented a method for multi-modal obstacle detection by fusing camera and lidar sensing with a conditional random field.
Initial 2D (camera) and 3D (lidar) classifiers have been combined probabilistically, exploiting both spatial, temporal, and multi-modal links between corresponding 2D and 3D regions.
The method has been evaluated on data gathered in various agricultural environments with a moving ground vehicle.

Results have shown that for a two-class classification problem (ground and non-ground), only the camera leveraged from information provided by the lidar.
In this case, the geometric classifier (lidar) could single-handedly distinguish ground and non-ground structures.
For simple traversability assessment, a lidar might therefore be sufficient for distinguishing traversable and non-traversable ground areas.
However, as more classes were introduced (\textit{ground}, \textit{sky}, \textit{vegetation}, and \textit{object}), both modalities complemented each other and improved the mean classification score.

The introduction of spatial, multi-modal, and temporal links in the CRF fusion algorithm showed gradual improvements in the mean intersection over union classification score.
Adding spatial links between neighboring segments in 2D and 3D separately, first improved the initial and individual classification results with $5.7\%$ in 2D and $7.0\%$ in 3D.
Spatial links act as smoothing terms and help reduce local noise and ensure consistent predictions across the entire image and point cloud.
Then, adding multi-modal links between 2D and 3D caused a further improvement of $1.4\%$ in 2D and $7.9\%$ in 3D.
And finally, adding temporal links between successive frames caused an increase of $0.2\%$ in 2D and $1.5\%$ in 3D.
Temporal links act as another smoothing term and help ensure consistent predictions over time, which may ease subsequent motion or path planning.
The method proves that it is possible to reduce uncertainty when probabilistically fusing lidar and camera as opposed to applying each sensor individually. 
Whether the performance gains justify the complexity of the method will depend on the specific agricultural application, including whether binary ground/non-ground classification is sufficient, or whether multiclass classification is required. 

The introduction of temporal links in the CRF caused a smaller improvement than the introduction of spatial and multi-modal links.
We believe, however, that the increase is significant and worth reporting, as it extends and improves an offline method from scene analysis to an online applicable method for robotics.

A traditional computer vision pipeline was compared to a deep learning approach for the 2D classifier.
It was shown that deep learning outperformed traditional vision when evaluating their individual performances.
However, when applying a CRF and fusing with lidar, the two methods gave similar results.

Finally, transferability was evaluated across agricultural domains (\textit{mangoes}, \textit{lychees}, \textit{apples}, \textit{almonds}, and \textit{dairy}) and classes (\textit{ground}, \textit{sky}, \textit{vegetation}, and \textit{object}).
Results showed that features and classifiers transferred well across domains for the \textit{ground} and \textit{sky} classes, whereas \textit{vegetation} and \textit{object} were less transferable due to a larger inter-domain variation in appearance and geometry.
Adding domain-specific training data confirmed this observation, as classification results of particularly \textit{vegetation} and \textit{object} were further increased.

In situations where scene parsing can benefit from input from different sensor modalities, the paper provides a flexible, probabilistically consistent framework for fusing multi-modal spatio-temporal data. 
The approach is flexible and may be extended to include additional heterogeneous data sources in future work, including radar, stereo or thermal vision, all of which are directly applicable within the framework.

\subsubsection*{Funding}
This work is sponsored by the Innovation Fund Denmark as part of the project SAFE - Safer Autonomous Farming Equipment (project no. 16-2014-0) and supported by the Australian Centre for Field Robotics at The University of Sydney and Horticulture Innovation Australia Limited through project AH11009 Autonomous Perception Systems for Horticulture Tree Crops. 
Further information and videos available at: \url{https://sydney.edu.au/acfr/agriculture}.

\bibliographystyle{apalike}
\bibliography{refs}

\section*{Appendix A: Parameter List}
\label{appendix-parameters}
A list of all parameter settings for 2D and 3D classifiers and the CRF fusion framework is available in Table~\ref{parameter-table}.
\begin{table*}[h]
\caption{Algorithm parameters used for initial classifiers (2D and 3D) and CRF fusion.} \label{parameter-table}
\resizebox{\textwidth}{!}{
\begin{tabular}{lr|lr|lr}
\toprule
\multicolumn{2}{c|}{2D classifiers} & \multicolumn{2}{c|}{3D classifier} & \multicolumn{2}{c}{CRF fusion} \\
\midrule
\textbf{Image}								& 			& \textbf{Point cloud} 			& 					& \textbf{Pairwise potentials}	&  		\\
\hspace{2mm} width							& 616 		& \hspace{2mm} beams			& 64 				& \hspace{2mm} $\sigma_{\text{2D}}$	& 0.5	\\
\hspace{2mm} height							& 808 		& \hspace{2mm} $\theta_H$		& 0.08$^{\circ}$ 	& \hspace{2mm} $\sigma_{\text{3D}}$ 	& 0.5		\\
\textbf{SLIC}								& 			& \textbf{Feature extraction}	& 			 		& \hspace{2mm} $\sigma_{\text{Nav}}$ 	& 1 	\\
\hspace{2mm} region size					& 40 		& \hspace{2mm} $M$				& 60 				& \hspace{2mm} $\sigma_{\text{Time}}$	& $1/\sqrt{8}$ 		\\
\hspace{2mm} regularization factor			& 3000 		& \textbf{Supervoxels} 			& 	 				& \hspace{2mm} time between $f_p$ and $f_c$ 									& 2.0 s 		\\
\textbf{SIFT}		 						&  			& \hspace{2mm} seed resolution 	& 0.1 				&  								&  		\\
\hspace{2mm} bin size						& 3 		& \hspace{2mm} voxel resolution	& 0.2 				&  								&  		\\
\hspace{2mm} magnification factor			& 4.8 		& \hspace{2mm} $\lambda$ 		& 1 				&  								&  		\\
\textbf{BoW}			 					&  			& \hspace{2mm} iterations 		& 10 				&  								&  		\\
\hspace{2mm} vocabulary size				& 50 		&  								&  					&  								&  		\\
\hspace{2mm} fraction of strongest features	& 0.5 		&  								&  					&  								&  		\\
\textbf{SVM} 								&  			& \textbf{SVM} 					&  					&  								&  		\\
\hspace{2mm} examples 						& 100000 	& \hspace{2mm} examples 		& 40000				&  								&  		\\
\hspace{2mm} kernel 						& RBF 		& \hspace{2mm} kernel 			& RBF 				&  								&  		\\
\hspace{2mm} $\gamma$ 						& 1/57 		& \hspace{2mm} $\gamma$ 		& 1/9 				&  								&  		\\
\hspace{2mm} C 								& 1 		& \hspace{2mm} C 				& 1 				&  								&  		\\
\textbf{CNN} 								&  			& 			 					&  					&  								&  		\\
\hspace{2mm} optimizer						& SGD		& 						 		& 					&  								&  		\\
\hspace{2mm} learning rate					& $10^{-12}$& 						 		& 					&  								&  		\\
\hspace{2mm} momentum 						& 0.99 		& 					 			& 	 				&  								&  		\\
\hspace{2mm} batch size 					& 1 		& 						 		& 	 				&  								&  		\\
\hspace{2mm} epochs 						& 10 		& 						 		& 	 				&  								&  		\\
\hspace{2mm} data augmentation				& horizontal flip	& 		 				& 	 				&  								&  		\\
\bottomrule
\end{tabular}
}
\end{table*}

\end{document}